\documentclass[]{elsarticle}
\pdfoutput=1

\usepackage[hyphens]{url}
\usepackage[hidelinks]{hyperref}
\usepackage{listings}
\usepackage[usenames,dvipsnames]{color}
\usepackage{csquotes}
\usepackage{dsfont}
\usepackage{caption}
\usepackage{floatrow}
\usepackage{subfig}
\usepackage{bm}
\usepackage{microtype}

\usepackage{booktabs}
\usepackage[svgnames]{xcolor}
\usepackage{colortbl}%

\newcommand{\myrowcolour}{\rowcolor[gray]{0.925}}
\renewcommand{\myrowcolour}{}

\newcommand{\denselist}{\itemsep -1.5pt\partopsep -20pt}

\newif\ifblackandwhite

\ifblackandwhite
  
\else

\fi

\usepackage{pbox}
\newcommand{\myparagraph}[1]{~\\\noindent{}\emph{#1}}

\newcommand{\changed}[1]{#1}
\newcommand{\changedcrc}[1]{#1}

\newcommand{\emcite}[1]{\citeauthor{#1}~\cite{#1}}

\usepackage{amsmath}
\usepackage{amsthm}
\usepackage{amssymb}

\newcommand{\vs}[0]{\emph{vs}}
\newcommand{\ie}[0]{\emph{i.e.}}
\newcommand{\eg}[0]{\emph{e.g.}}

\newcommand{\vTheta}{{\bm{\Theta}}}
\newcommand{\vtheta}{{\bm{\theta}}}

\newcommand{\smac}{\textit{SMAC}} 
\newcommand{\paramils}{\textit{ParamILS}} 

\newcommand{\gga}{\textit{GGA}} 

\newcommand{\minialgofont}[1]{\scriptsize{\textit{#1}}}

\newcommand{\minismacd}{\mbox{\minialgofont{SMAC-d}}}
\newcommand{\minismacc}{\mbox{\minialgofont{SMAC-c}}}
\newcommand{\minipils}{\minialgofont{PILS}}
\newcommand{\miniggad}{\mbox{\minialgofont{GGA-d}}}
\newcommand{\miniggac}{\mbox{\minialgofont{GGA-c}}}

\newcommand{\minisathack}{\mbox{\textit{Minisat-HACK-999ED}}}
\newcommand{\minisathacksmall}{\mbox{\scriptsize\textit{Minisat-HACK-999ED}}}
\newcommand{\riss}{\mbox{\textit{Riss-4.27}}}
\newcommand{\yalsat}{\textit{YalSAT}}
\newcommand{\clasp}{\mbox{\textit{Clasp-3.0.4-p8}}}
\newcommand{\cryptominisat}{\textit{Cryptominisat}}
\newcommand{\lingeling}{\textit{Lingeling}}
\newcommand{\sparrow}{\textit{SparrowToRiss}}
\newcommand{\dccsat}{\mbox{\textit{DCCASat+march-rw}}}
\newcommand{\csccsat}{\textit{CSCCSat2014}}
\newcommand{\probsat}{\textit{ProbSAT}}

\newcommand{\fuzzsat}{\textit{FuzzSAT}} 
\newcommand{\newdame}{\textit{new-Dame}} 
\newcommand{\claspthirteen}{\mbox{\textit{Clasp-2.1.3}}}
\newcommand{\claspbasic}{\textit{Clasp}}
\newcommand{\claspshort}{\textit{Clasp}}
\newcommand{\rissgExt}{\textit{Riss3gExt}} 
\newcommand{\rissg}{\textit{Riss3g}} 
\newcommand{\SolverFourtyThree}{\textit{Solver43}}
\newcommand{\forlnodrup}{\mbox{\textit{Forl-nodrup}}} 
\newcommand{\simpsat}{\textit{Simpsat}}
\newcommand{\satj}{\textit{Sat4j}} 
\newcommand{\gnoveltyGCwa}{\textit{Gnovelty+GCwa}} 
\newcommand{\gnoveltyPCL}{\textit{Gnovelty+PCL}} 
\newcommand{\gnoveltyGCa}{\textit{Gnovelty+GCa}} 

\newcommand{\bmc}{\textit{Bounded Model Checking 2008 (BMC)}}
\newcommand{\bmcshort}{\textit{BMC}}
\newcommand{\circuit}{\textit{Circuit Fuzz}}
\newcommand{\hw}{\textit{Hardware Verification (IBM)}} 
\newcommand{\hwshort}{\textit{IBM}} 
\newcommand{\ibm}{\textit{IBM}} 
\newcommand{\swv}{\textit{SWV}}
\newcommand{\swvshort}{\textit{SWV}}

\newcommand{\gi}{\textit{Graph Isomorphism (GI)}}
\newcommand{\gishort}{\textit{GI}}
\newcommand{\labs}{\textit{Low Autocorrelation Binary Sequence (LABS)}}
\newcommand{\labsshort}{\textit{LABS}}
\newcommand{\queens}{\textit{N-Rooks}}

\newcommand{\threecnf}{\textit{3cnf}}
\newcommand{\kthree}{\textit{K3}}
\newcommand{\unif}{\textit{unif-k5}}

\newcommand{\threesatonek}{\textit{3sat1k}}
\newcommand{\fivesatfiveh}{\textit{5sat500}}
\newcommand{\sevensatninety}{\textit{7sat90}}

\newcommand{\indu}{\textit{Industrial SAT+UNSAT}}
\newcommand{\crafted}{\textit{crafted SAT+UNSAT}}
\newcommand{\random}{\textit{Random SAT+UNSAT}}
\newcommand{\randomsat}{\textit{Random SAT}}

\usepackage{todonotes}

\begin{document}

\title{The Configurable SAT Solver Challenge (CSSC)}

\address[freiburg]{University of Freiburg, Germany}
\address[ulm]{University of Ulm, Germany}
\address[vancouver]{University of British Columbia, Vancouver, Canada}

\author[freiburg]{Frank Hutter} 
\ead{fh@cs.uni-freiburg.de}

\author[freiburg]{Marius Lindauer}
\ead{lindauer@cs.uni-freiburg.de}

\author[ulm]{Adrian Balint} 
\ead{adrian.balint@uni-ulm.de}

\author[vancouver]{Sam Bayless} 
\ead{sbayless@cs.ubc.ca}

\author[vancouver]{Holger Hoos} 
\ead{hoos@cs.ubc.ca}

\author[vancouver]{Kevin Leyton-Brown} 
\ead{kevinlb@cs.ubc.ca}

\begin{abstract}

It is well known that different solution strategies work well for different types of 
instances of hard combinatorial problems. As a consequence, most solvers for the propositional satisfiability 
problem (SAT) expose parameters that allow them to be customized to a particular family of 
instances.
In the international SAT competition series, these parameters are ignored: solvers are run using a single default parameter setting (supplied by the authors) for all benchmark instances in a given track. 
While this competition format rewards solvers with robust default settings, 
it does not reflect the situation faced by a practitioner who only cares about performance on one particular application 
and can invest some time into tuning solver parameters for this application.
The new Configurable SAT Solver Competition (CSSC) compares solvers in this latter setting,
scoring each solver by the performance it achieved after a fully automated 
configuration step. This article describes the CSSC in more detail, and reports the results 
obtained in its two instantiations so far, CSSC 2013 and 2014.
\end{abstract}

\begin{keyword}
Propositional satisfiability \sep algorithm configuration \sep empirical evaluation \sep competition
\end{keyword}

\maketitle


\section{Introduction}

The propositional satisfiability problem (SAT) is one of the most prominent problems in AI.
It is relevant both for theory (having been the first problem proven to be NP-hard~\cite{cook-stoc71})
and for practice (having important applications in many fields, such as hardware
and software verification~\cite{biere-dac99,prasad-jsttt05a,clarke2004tool}, 
test-case generation~\cite{stephan-ieee96,cadar2008klee}, AI
planning~\cite{kautz-aaai96a,kautz-ijcai99}, 
scheduling~\cite{crawford-aaai94}, and graph colouring~\cite{gelder-color02}).
%
The SAT community has a long history of regularly assessing the state of
the art via competitions~\cite{Jarvisalo-aij12}. The first SAT competition dates back to the year 2002~\cite{simon-amai05a},
and the event has been growing over time: in 2014, it had a record participation of 58 solvers by 79 authors in 11 tracks~\cite{satchallenge14}.


In practical applications of SAT, solvers can typically be adjusted to perform well for the specific type of instances at hand, such as software verification instances generated by a particular static checker on a particular software system~\cite{babic-cav07}, or a particular family of bounded model checking instances~\cite{zarpas-sat05a}.
To support this type of customization, most SAT solvers already expose a range of command line
\emph{parameters} whose settings substantially affect most parts of the solver. 
Solvers typically come with robust default parameter settings meant to provide good all-round performance, but it is widely known that adjusting parameter settings to particular target instance classes can yield orders-of-magnitude speedups \cite{hutter-fmcad07a,khudabukhsh-ijcai09a,tompkins-sat11}.
Current SAT competitions do not take this possibility of customizing solvers into account, and rather evaluate solver performance with default parameters.

Unlike the SAT competition, the \emph{Configurable SAT Solver Challenge (CSSC)} evaluates SAT solver performance \emph{after} application-specific 
customization, thereby taking into account the fact that
effective algorithm configuration procedures can automatically customize
solvers for a given distribution of benchmark instances.
Specifically, for each type of instances $T$ and each SAT solver $S$, an automated fixed-time 
offline configuration phase determines parameter settings of $S$ optimized for high performance
on $T$. Then, the performance of $S$ on $T$ is evaluated with these settings, and the solver
with the best performance wins.

To avoid a potential misunderstanding, we note that for winning the competition, only solver performance after configuration counts, and that it does \emph{not} matter how much performance was improved by configuration. As a consequence, in principle, even a parameterless solver could win the CSSC if it was very strong: it would not benefit from configuration, but if it nevertheless outperformed all solvers that were specially configured for the instance families in a given track, it would still win that track. (In practice, we have not observed this, since the improvements resulting from configuration tend to be large.)

The competition conceptually most closely related to the CSSC is the learning track of the international planning competition (IPC, see, e.g., the description by \emcite{fern-ml11}\footnote{\url{http://www.cs.colostate.edu/~ipc2014/}},
which also features an offline time-limited learning phase on training instances from
a given planning domain and an online testing phase on a disjoint set of instances from the same domain.
The main difference between this IPC learning track and the CSSC (other than their focus on 
different problems) is that in the IPC learning track every planner uses its own learning method, and the learning methods thus vary between entries.
In contrast, in the CSSC, the corresponding customization process is part of the competition setup and
uses the same algorithm configuration procedure for each submitted solver.
Our approach to evaluating solver performance after configuration could of course be transferred to any other competition.
(In fact, the 2014 IPC learning track for non-portfolio solvers was won by FastDownward-\smac{}~\cite{seipp-ipc14a}, a system that employs
a similar combination of general algorithm configuration and a highly parameterized solver framework as we do in the CSSC.)

In the following, we first describe the criteria we used for the design of the CSSC
(Section~\ref{sec:design_criteria}).
Next, we provide some background on the automated algorithm configuration methods we used 
when running the competition (Section \ref{sec:algoconf}).
Then, we discuss the two CSSCs we have held so far (in 2013 and 2014); we discuss each of these
competitions in turn (Sections \ref{sec:cssc2013} and \ref{sec:cssc2014}), including
the specific benchmarks used, the participating solvers, and the results.
We describe two main insights that we obtained from these results: 
\begin{enumerate}
	\item In many cases, automated algorithm configuration found parameter settings that performed much better than the solver defaults, in several cases yielding average speedups of several orders of magnitude. 
	\item Some solvers benefited more from automated configuration than others; as a result, the ranking of algorithms after configuration was often substantially different from the ranking based on the algorithm defaults (as, e.g., measured in the SAT competition). 
\end{enumerate}
Finally, we analyze various aspects of these results (Section \ref{sec:discussion-of-results}) and discuss the implications we see for future algorithm development (Section \ref{sec:conclusion}).

\section{Design Criteria for the CSSC} \label{sec:design_criteria}

We organized the CSSC 2013 and 2014 in coordination with the international SAT competition and presented them in the competition slots at the 2013 and 2014 SAT conferences (as well as in the 2014 FLoC Olympic Games, in which all SAT-related competitions took part). We coordinated solver submission deadlines with the SAT competition to minimize overhead for participants, who could submit their solver to 
the SAT competition using default parameters and then open up their parameter spaces for the CSSC. 

We designed the CSSC to remain close to the international SAT competition's established format;
in particular, 
we used the same general categories: \emph{industrial}, \emph{crafted}, and \emph{random}, and, in 2014 also \emph{random satisfiable}. 
Furthermore, we used the same input and output formats, 
the SAT competition's mature code for verifying correctness of solver outputs (only for checking models of satisfiable instances; we did not have a certified UNSAT track), and the same scoring function (number of instances solved, breaking ties by average runtime).

The main way our setup differed from that of the SAT competition was that we used a relatively small budget of five minutes per solver run.
We based this choice partly on the fact that many solvers have runtime distributions with rather long tails (or even heavy tails~\cite{gomes-jar00a}),
and that practitioners often use \emph{many instances} and relatively \emph{short runtimes} to benchmark solvers for a new application domain.
There is also evidence that SAT competition results would remain quite similar if
based on shorter runtimes, but not if based on fewer instances~\cite{HutHooLey10}.
Therefore, in order to achieve more robust performance within a fixed computational budget, we chose to use 
many test instances (at least 250 for each benchmark) but relatively low runtime cutoffs per solver run (five minutes).
(We also note that a short time limit of five minutes has already been used in the agile track of the 2014 International Planning Competition.)
Due to constraints imposed by our computational infrastructure, we used a memory limit of 3GB for each solver run.

%
To simulate the situation faced by practitioners with limited computational resources, 
we limited the computational budget for configuring a solver on a benchmark with a 
given configuration procedure to two days on 4 or 5 cores (in 2014 and 2013, respectively). 
Our results are therefore indicative of what could be obtained by performing configuration runs over the weekend on a modern desktop machine.

\subsection{Controlled Execution of Solver Runs} \label{sec:controlled_execution}

Since all configuration procedures ran in an entirely automated fashion, they had to be robust against any kind of 
solver failure (segmentation faults, unsupported combinations of parameters, wrong results, infinite loops, etc.).
We handled all such conditions in a generic wrapper script that used Olivier Roussel's runsolver tool~\cite{olivier-jsat11a} to limit runtime and memory,
and counted any errors or limit violations as timeouts at the maximum runtime of 300 seconds.
We also kept track of the rich solver runtime data we gathered in our configuration runs and made it publicly available on the competition website.

\subsection{Choice of Configuration Pipeline} 
To avoid bias arising from our choice of algorithm configuration method, we independently used 
all
three 
state-of-the-art methods applicable for runtime optimization (\paramils{}~\cite{hutter-jair09a}, \gga{}~\cite{ansotegui-cp09a}, and \smac{}~\cite{hutter-lion11a}, as described in detail in Section \ref{sec:algoconf}).
We evaluated the configurations resulting from all configuration runs on the entire training data set and
selected the configuration with the best training performance. We then executed only this configuration 
on the test set to determine the performance of the configured solver.
Except where specifically noted otherwise, all performance data we report in this article is for this optimized 
configuration on previously unseen test instances from the respective benchmark set.

\changed{
\subsection{Pre-submission Bug Fixing} 
As part of the submission package, we provided solver authors with our configuration pipeline, so that they could run it themselves to identify bugs in their solver before submission (e.g., problems due to the choice of non-default parameters). We also provided some trivial benchmark sets for this pre-submission preparation, which were not part of the competition. 

We did not offer a bug fixing phase after solver submission, except that we ran a very simple configuration experiment (10 minutes on trivial instances) to verify that the setup of all participants was correct.
}

\subsection{Choice of Benchmarks} 

We chose the benchmark families for the CSSC to be relatively homogeneous in terms of the origin and/or construction process of instances in the same family.
Typically, we selected benchmark families that are neither too easy (since speedups are less interesting for easy instances), 
nor too hard (so that solvers could solve a large fraction of instances within the available computational budgets). 
%
We aimed for benchmark sets of which at least 20-40\% could be solved within the maximum runtime on a recent machine by the default configuration of a SAT solver that would perform reasonably well in the SAT competition.
We also aimed for benchmark sets with a sufficient number of instances to safeguard against over-tuning; in practice, the smallest datasets we used had 500 instances: 250 for training and 250 for testing.
%

\changed{We did not disclose which benchmark sets we used until the competition results were announced. While we encouraged competition entrants to also contribute benchmarks, we made sure to not substantially favor any solver by using such contributed benchmarks.}

\section{Automated Algorithm Configuration Procedures} \label{sec:algoconf}

The problem of finding performance-optimizing algorithm parameter settings arises 
for many computational problems. In recent years, the AI community has developed several
dedicated systems for this general \emph{algorithm configuration} problem \cite{hutter-jair09a,ansotegui-cp09a,lopez-ibanez-tech11a,hutter-lion11a}.

We now describe this problem more formally.
Let $A$ be an algorithm having $n$ parameters with domains $\Theta_1, \dots, \Theta_n$.
Parameters can be \emph{real-valued} (with domains $[a,b]$, where $a,b \in \mathds{R}$ and $a < b$), \emph{integer-valued} (with domains $[i,j]$, where $i,j \in \mathds{Z}$ and $i < j$), or \emph{categorical} (with finite unordered domains, such as \{red, blue, green\}).
Parameters can also be \emph{conditional} on an instantiation of other (so-called \emph{parent}) parameters; as an example, consider the parameters of a heuristic mechanism $h$,
	which are completely ignored unless $h$ is chosen to be used by means of another, categorical parameter.
Finally, some combinations of parameter instantiations can be labelled as \emph{forbidden}.

Algorithm $A$'s \emph{configuration space} $\vTheta$ then consists of all possible combinations of parameter values:
$\vTheta = \Theta_1 \times \cdots \times \Theta_n$. 
We refer to elements $\vtheta = \langle{}\theta_1, \dots, \theta_n\rangle{}$ of this configuration 
space as \emph{parameter configurations}, or simply \emph{configurations}. 
Given a benchmark set $\Pi$ and a performance metric $m(\vtheta,\pi)$ capturing
the performance of configuration $\vtheta\in\vTheta$ on problem instance
$\pi\in\Pi$, the algorithm configuration problem then aims to 
find a configuration $\vtheta\in\vTheta$ that minimizes
$m$ over $\Pi$, \ie{}, that minimizes\footnote{An alternative 
definition considers the optimization of expected performance across a \emph{distribution} of instances 
rather than average performance across a \emph{set} of instances~\cite{hutter-jair09a}.
%
%
What we consider here can be seen as a special case where the distribution is uniform over a given set of 
training instances. It is also possible
to optimize performance metrics other than mean performance across instances, but mean
performance is by far the most widely used option.}
\begin{equation*}
\label{eqn:f}f(\vtheta) = \frac{1}{|\Pi|} \cdot \sum_{\pi\in\Pi} m(\vtheta,\pi).
\end{equation*}

\noindent{}In the CSSC, the specific metric $m$ we optimized was \emph{penalized average runtime (PAR-10)}, which counts runs that exceed a maximal cutoff time $\kappa_{max}$ without solving the given instance as $10 \cdot \kappa_{max}$. We terminated individual solver runs as unsuccessful after $\kappa_{max} = 300$ seconds.

We refer to an instance of the algorithm configuration problem as a \emph{configuration scenario} and to a method for solving the algorithm configuration problem as a \emph{configuration procedure} (or a \emph{configurator}), in order to avoid confusion with the solver to be optimized (which we refer to as the \emph{target algorithm}) and the problem instances the solver is being optimized for.

Algorithm configuration has been demonstrated to be very effective for optimizing various SAT solvers in the literature.
For example, \citet{hutter-fmcad07a} configured the algorithm Spear~\cite{BabHut07} on formal verification
instances, achieving a 500-fold speedup on software verification instances generated with the
static checker Calysto~\cite{babic-cav07} and a 4.5-fold speedup on IBM bounded model checking instances by \citet{zarpas-sat05a}.
Algorithm configuration has also enabled the development of general frameworks for stochastic local search SAT solvers
that can be automatically instantiated to yield state-of-the-art performance on new types of instances; examples for such 
frameworks are SATenstein~\cite{khudabukhsh-ijcai09a} and Captain Jack~\cite{tompkins-sat11}.

While all of these applications used the local-search based algorithm configuration
method \paramils{}~\cite{hutter-jair09a}, in the CSSC we wanted to avoid bias that could arise from commitment to one particular algorithm configuration method and thus used all three existing general algorithm configuration methods for runtime optimization: 
\paramils{}, \gga{}~\cite{ansotegui-cp09a}, and \smac{}~\cite{hutter-lion11a}.\footnote{We did not use the iterated racing method I/F-Race~\cite{lopez-ibanez-tech11a}, since it does not effectively support runtime optimization and its authors thus discourage its use for this purpose (personal communication with Manuel L{\'o}pez-Ib{\'a}{\~n}ez and Thomas St\"{u}tzle).} 
We refer the interested reader to \ref{app:conf-proc-details} for details on each of these configurators. Here, we only mention some details that were important for the setup of the CSSC:
\begin{itemize}
	\item \paramils{} does not natively support parameters specified only as real-
		or integer-valued intervals, but requires all parameter values to be listed explicitly;
		for simplicity, we refer to the transformation used to satisfy this requirement as 
		\emph{discretization}.
		 When multiple parameter spaces were available for a solver, we only ran \paramils{} on the discretized version, whereas we ran \gga{} and \smac{} on both the discretized and the non-discretized versions.
	\item 
	\paramils{} and \smac{} have been shown to benefit substantially from multiple independent runs, \changed{since they are randomized algorithms}. Given $k$ cores, the usual approach is simply to execute $k$ independent configurator runs and pick the configuration from the one with best performance on the training set. \gga{}, on the other hand, can use multiple cores on a single machine, and in fact requires these to run effectively. Therefore, given $k$ available cores per configuration approach, we used $k$ independent runs of each \paramils{} and \smac{}, and one run using all $k$ cores for \gga{}.
	\item \gga{} could not handle the complex parameter conditionalities found in some solvers; for those solvers, we only ran \paramils{} and \smac{}.
\end{itemize}

%
%

\section{The Configurable SAT Solver Challenge 2013} \label{sec:cssc2013}

\begin{table}[tb]
\sffamily\small\centering%
\begin{tabular}{@{}l l l l l r}
\toprule[1.0pt]

Benchmark & \#Train & \#Test & \#Variables & \#Clauses & Reference\\
\midrule

\swvshort & 302 & 302 & $68.9\text{k} \pm 57.0\text{k}$ & $182\text{k} \pm 206\text{k}$ & \cite{babic-hvc07}\\
\myrowcolour{}\hwshort & 383 & 302 & $96.4\text{k} \pm 170\text{k}$ & $413\text{k} \pm 717\text{k}$ & \cite{zarpas-sat05a}\\
\circuit & 299 & 302 &$5.53\text{k} \pm 7.45\text{k}$ & $18.8\text{k} \pm 25.3\text{k}$ & \cite{brummayer-sat10a}\\
\myrowcolour{}\bmcshort{} & 807 & 302 & $446\text{k} \pm 992\text{k}$ & $1.09\text{m} \pm 2.70\text{m}$ & \cite{biere-misc08a}\\

\midrule

\gishort & 1032 & 351 & $11.2\text{k} \pm 17.8\text{k}$ & $2.98\text{m} \pm 8.03\text{m}$ & \cite{mugrauer2013,toran2013}\\
\myrowcolour{}\labsshort & 350 & 351 & $75.9\text{k} \pm 75.7\text{k}$ & $154\text{k} \pm 153\text{k}$ & \cite{mugrauer2013-2}\\

\midrule

\kthree & 300 & 250 & $262 \pm 43$ & $1116 \pm 182$ & \cite{bayless-lion14a}\\
\myrowcolour{}\unif & 300 & 250 & $50 \pm 0$ & $1056 \pm 0$ & --\\
\fivesatfiveh{} & 250 & 250 & $500 \pm 0$ & $10000 \pm 0$ & \cite{tompkins-sat11}\\

\bottomrule[1.0pt]

\end{tabular}
\caption{Overview of benchmark sets used in the CSSC 2013 tracks \indu{}, \crafted{}, and \random{} (from top to bottom); k and m stand for factors of one thousand and one million, respectively.\label{tab:benchmarks-2013}}

\end{table}


The first CSSC\footnote{\url{http://www.cs.ubc.ca/labs/beta/Projects/CSSC2013/}} was held in 2013. 
It featured three tracks mirroring those of the SAT competition: \indu{}, \crafted{}, and \random{}.
Table \ref{tab:benchmarks-2013} lists the benchmark families we used in each of these tracks, 
%
all of which are described in detail in \ref{app:benchmarks}.
Within each track, we used the same number of test instances 
for each benchmark family, thereby weighting each equally in our analysis.

%

\subsection{Participating Solvers and Their Parameters} \label{sec:cssc2013-solvers}

\begin{table}[tb]
\sffamily\small\centering%
\setlength\tabcolsep{0.5em}
\begin{tabular}{@{}l rrrr rrr r}
\toprule[1.0pt]

Solver & \multicolumn{4}{c}{\# Parameters} & \multicolumn{3}{c}{\# Configurations} & Reference \\
		& c & i & r & cond. & original & discretized & disc. subset & 	 \\
\midrule
\gnoveltyGCa & $2$ & $0$ & $0$ & $0$ & $110$ & -- & -- & \cite{gnovelty}\\
\myrowcolour{}\gnoveltyGCwa & $2$ & $0$ & $0$ & $0$ & $110$ & -- & -- & \cite{gnovelty}\\
\gnoveltyPCL & $5$ & $0$ & $0$ & $0$ & $20\,000$ & -- & -- & \cite{gnovelty}\\
\myrowcolour{}\simpsat & $5$ & $0$ & $0$ & $0$ & $2\,400$ & -- & -- & \cite{han2012simpsat}\\
\satj & $10$ & $0$ & $0$ & $4$ & $2 \times 10^{7}$ & -- & -- & \cite{leberre-sat4j}\\
\myrowcolour{}\SolverFourtyThree & $12$ & $0$ & $0$ & $0$ & $5 \times 10^{6}$ & -- & -- & \cite{solver43}\\
\forlnodrup & $44$ & $0$ & $0$ & $0$ & $3 \times 10^{18}$ & -- & -- & \cite{forlnodrup}\\
\myrowcolour{}\claspthirteen & $42$ & $34$ & $7$ & $60$ & $\infty$ & $10^{45}$ & -- & \cite{clasp}\\
\rissg & $125$ & $0$ & $0$ & $107$ & $2 \times 10^{53}$ & -- & -- & \cite{riss3g}\\
\myrowcolour{}\rissgExt & $193$ & $0$ & $0$ & $168$ & $2 \times 10^{82}$ & -- & -- & \cite{riss3g}\\
\lingeling & $102$ & $139$ & $0$ & $0$ & $1 \times 10^{974}$ & $1 \times 10^{136}$ & $2 \times 10^{39}$ & \cite{biere-tech13a}\\
\bottomrule[1.0pt]

\end{tabular}
\caption{Overview of solvers in the Configurable SAT Solver Challenge (CSSC) 2013 and their parameters of various types (`c' for categorical, `i' for integer, `r' for real-valued'); `cond' identifies how many of these parameters are conditional. We also list the sizes of the configuration spaces provided by the solver developers (original, discretized, and the subset of the discrete parameters). Solvers are ordered by the total number of parameters they expose ($c+i+r$).\label{tab:solvers2013}}

\end{table}

Table \ref{tab:solvers2013} summarizes the solvers that participated in the CSSC 2013, along with 
information on their configuration spaces. The eleven submitted solvers ranged from complete solvers based on conflict-directed clause learning (CDCL; \cite{cdcl}) to stochastic local search (SLS; \cite{SLS}) solvers. The degree of parameterization varied substantially across these submitted solvers, from 2 to 241 parameters.
We briefly discuss the main features of the solvers' parameter configuration spaces, ordering solvers by their number of parameters.

\myparagraph{\gnoveltyGCa{} \emph{and} \gnoveltyGCwa{}}~\cite{gnovelty} are closely related SLS solvers. Both have two numerical parameters: the probability of selecting false clauses randomly and the probability of smoothing clause weights.
The parameters were pre-discretized by the solver developer to 11 and 10 values, yielding 110 possible combinations.

\myparagraph{\gnoveltyPCL{}}~\cite{gnovelty}  is an SLS solver with five parameters: one binary parameter (determining whether the stagnation path is dynamic or static)
and four numerical parameters: the length of the stagnation path, the size of the time window storing stagnation paths, the probability of smoothing stagnation weights, and the probability of smoothing clause weights. All numerical parameters were pre-discretized to ten values each by the solver developer, yielding 20\,000 possible combinations. 

\myparagraph{\simpsat{}}~\cite{han2012simpsat} is a CDCL solver based on \cryptominisat{}~\cite{cryptominisat}, which adds additional strategies for explicitly handling XOR constraints~\cite{han2012boolean}. 
It has five numerical parameters that govern both these XOR constraint strategies and the frequency of random decisions. All parameters were pre-discretized by the solver developer, yielding 2\,400 possible combinations. 

\myparagraph{\satj{}}~\cite{leberre-sat4j} is full-featured library of solvers for Boolean satisfiability and optimization problems. For the contest, it applied its default CDCL SAT solver with ten exposed parameters: four categorical parameters deciding between different
restart strategies, phase selection strategies, simplifications, and cleaning; and six numerical parameters pre-discretized by its developer.

\myparagraph{\SolverFourtyThree{}}~\cite{solver43} is a CDCL solver with 12 parameters: three categorical parameters concerning sorting heuristics used in bounded variable eliminiation, in definitions and in adding blocked clauses; and nine numerical parameters concerning various frequencies, factors, and limits.
All parameters were pre-discretized by the solver developer.

\myparagraph{\forlnodrup{}}~\cite{forlnodrup} is a CDCL solver with 44 parameters. Most notably, these control variable selection, Boolean propagation, restarts, and learned clause removal.
About a third of the parameters are numerical (particularly most of those concerning restarts and learned clause removal); all parameters were pre-discretized by the solver developer.

\myparagraph{\claspthirteen{}}~\cite{clasp} is a solver for the more general answer set programming (ASP) problem, but it can also solve SAT, MAXSAT and PB problems.
As a SAT solver, \claspthirteen{} is a CDCL solver with $83$ parameters: $7$ for pre-processing,
$14$ for the variable selection heuristic, $18$ for the restart policy, 
$34$ for the deletion policy, and 10 for a variety of other uses.
The configuration space is highly conditional,
with several top-level parameters enabling or disabling certain strategies.
\claspthirteen{} exposes both a mixed continuous/discrete parameter configuration space and a manually-discretized one.

\myparagraph{\rissg{}}~\cite{riss3g} is a CDCL solver with 125 parameters. These include 6 numerical parameters from MiniSAT~\cite{minisat}, 10 numerical parameters from Glucose~\cite{glucose}, 17 mostly numerical Riss3G parameters, and 92 parameters controlling preprocessing/inprocessing performed by the integrated Coprocessor~\cite{manthey2012coprocessor}.
The inprocessor parameters resemble those in \lingeling{}~\cite{biere-tech13a}, emphasizing blocked clause elimination~\cite{BCE}, bounded variable addition~\cite{Mbva}, and probing~\cite{probing}.
About 50 of the parameters are Boolean, and most others are numerical parameters pre-discretized by the solver developer.
The parameter space is highly conditional, with inprocessor parameters dependent on a switch turning them on alongside various other dependencies.
Indeed, there are only 18 unconditional parameters. Finally, there are also seven forbidden parameter combinations that ascertain various switches are turned on if inprocessing is used.

\myparagraph{\rissgExt}~\cite{riss3g} is an experimental extension of \rissg{}. It exposes all of the parameters previously discussed for \rissg{}, along with an additional 11 Riss3G parameters and 57 inprocessing parameters. Its developer 
implemented all of these extensions in one week and did not have time for extensive testing before the CSSC; therefore, he submitted \rissgExt{} as closed source, making it ineligible for medals. We discuss the results of this closed-source solver separately, in~\ref{sec:riss3ext-discussion}.

\myparagraph{\lingeling{}}\cite{biere-tech13a} is a CDCL solver with 241 parameters (making it the solver with the largest configuration space in the CSSC 2013).
102 of these parameters are categorical, and the remaining 139 are integer-valued (76 of them with the trivial upper bound of max-integer, $2^{31}-1$).
\lingeling{} parameterizes many details of the solution process, including 
probing and look-ahead (about 25 mostly numerical parameters),
blocked clause elimination and bounded variable elimination (about 20 mostly categorical parameters each),
glue clauses (about 15 mostly numerical parameters), and a host of other mechanisms parameterized by about 5--10 parameters each.
%
\lingeling{} exposes its full parameter space, a discretized version of all parameters, and a subspace of only the categorical parameters (102 of them).

\subsection{Configuration Pipeline} \label{sec:cssc2013-pipeline}

We executed this competition on the QDR partition of the Compute Canada Westgrid cluster Orcinus. Each node in this cluster was provisioned with 24 GB memory and two 6-core, 2.66 GHz Intel Xeon X5650 CPUs with 12 MB L2 cache each, and ran Red Hat Enterprise Linux Server 5.5 (kernel 2.6.18, glibc 2.5).

In this first edition of the CSSC, we were unfortunately unable to run \gga{}. This was because it requires multiple cores for effective runtime minimization, and the respective multiple-core jobs we submitted on the Orcinus cluster were stuck in the queue for months without getting started. (Single-core runs, on the other hand, were often scheduled within minutes.)

\changed{We thus limited ourselves to using \paramils{} for the discretized parameter space of each of the 11 solvers and \smac{} for each of the parameter spaces that solver authors submitted (as discussed above, 9 submissions with one parameter space, 1 submission with two, and 1 submission with three, i.e., 14 in total). For each of the nine benchmark families, this gave rise to 11 configuration scenarios for \paramils{} and 14 for \smac{}, for a total of 225 configuration scenarios.
Since our budget for each configuration procedure was two CPU days on five cores (five independent runs of \paramils{} and \smac{}, respectively),
the competition's configuration phase required a total of 2250 CPU days (just over 6 CPU years). Thanks to a special allocation on the Orcinus cluster, we were able to complete this phase within a week.
} 

Following standard practice, we then evaluated the configurations resulting from all configuration runs on the entire training data set and selected the configuration with the best training performance. We then executed only this configuration on the test set to assess the performance of the configured solver. \changed{This evaluation phase required much less time than the configuration phase.}

\changed{We note that all scripts we used for performing the configuration and analysis experiments were written in Ruby and are available for download on the competition website.}

\subsection{Results} \label{sec:cssc2013-results}

\begin{table}[tb]
		\setlength\tabcolsep{1.00em}
		\sffamily\footnotesize\centering
		\begin{tabular}{l | lll}
		\toprule[1.0pt]
		Rank & \indu{} & \crafted & \random \\
		\midrule
		1$^{st}$ & \myrowcolour{}\lingeling & \clasp & \clasp\\
		2$^{nd}$ & \rissg & \forlnodrup & \lingeling\\
		3$^{rd}$ & \myrowcolour{}\SolverFourtyThree & \lingeling & \rissg\\
		\bottomrule[1.0pt]
		\end{tabular}
		\caption{Winners of the three tracks of CSSC 2013.\label{tab:results13}}
\end{table}

For each the three tracks of CSSC 2013, we configured each of the eleven submitted solvers for each of the benchmark families within the track and aggregated results across the respective test instances.
We show the winners in Table \ref{tab:results13} and discuss the results for each track
in the following sections.
Additional details, tables, and figures are provided in an accompanying technical report~\cite{hutter-tech14a}.

\changedcrc{We remind the reader that the CSSC score only depends on how well the configured solver did and \emph{not} on the difference between default and configured performance. We nevertheless still cover default performance prominently in the following results, in order to emphasize the impact configuration had and the difference between the CSSC and standard solver competitions (e.g., the SAT competition).}

%

\subsubsection{Results of the \indu{} Track} \label{sec:cssc2013-results-indu}

\begin{figure}[p]
\fbox{
	\noindent\begin{minipage}[t]{1.01\linewidth}
		\begin{center}\bf{}\underline{Results for CSSC 2013 \indu{} track}\end{center}\vspace*{-0.7cm}
		\begin{figure}[H]
		\centering
		\captionsetup{justification=centering,captionskip=0cm}
		\subfloat[\bmcshort{}\newline PAR-10: $302 \to
		282$]{\includegraphics[width=0.4\textwidth]{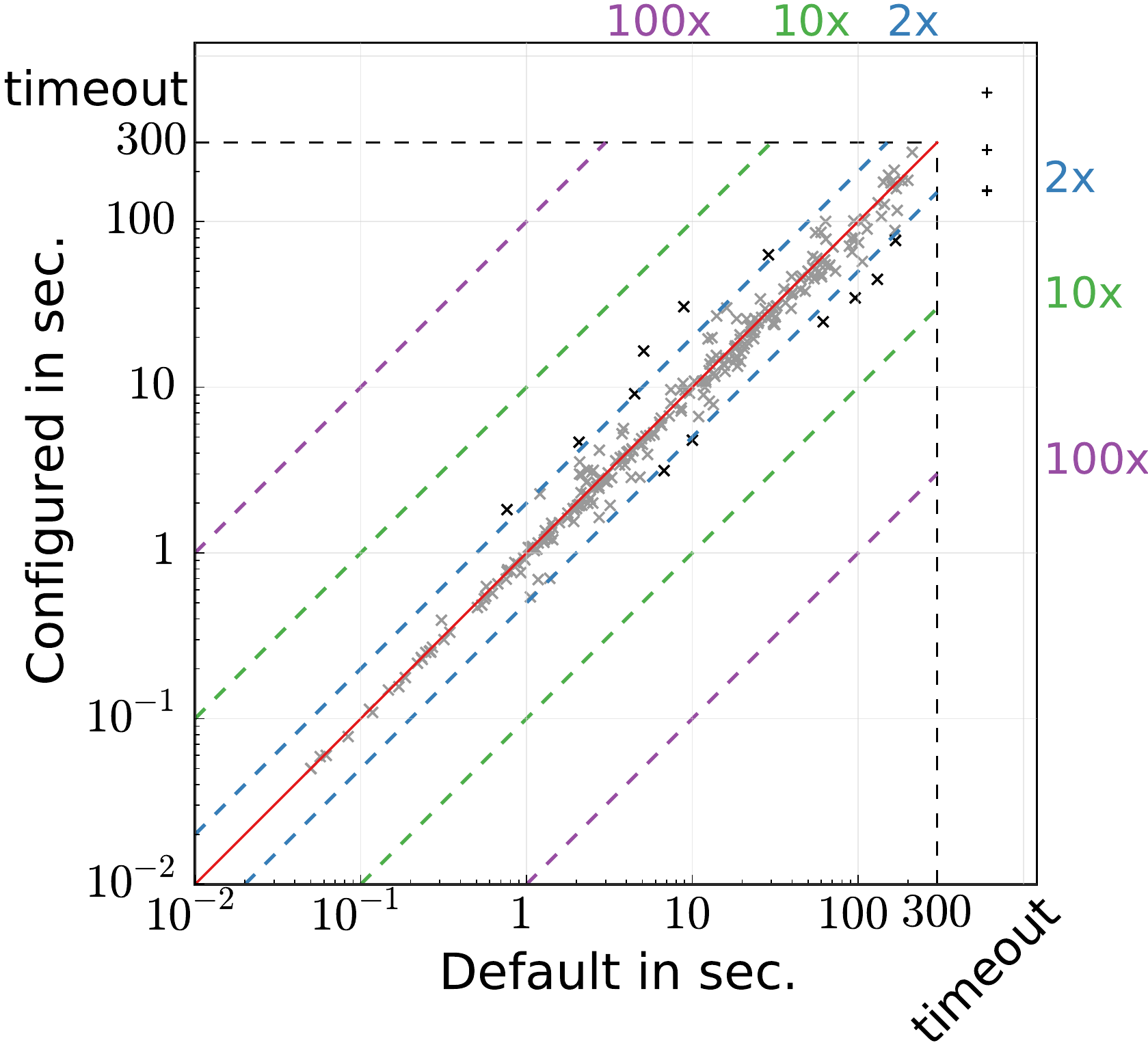}}
		\subfloat[\circuit{}\newline PAR-10: $409 \to
		241$]{\includegraphics[width=0.4\textwidth]{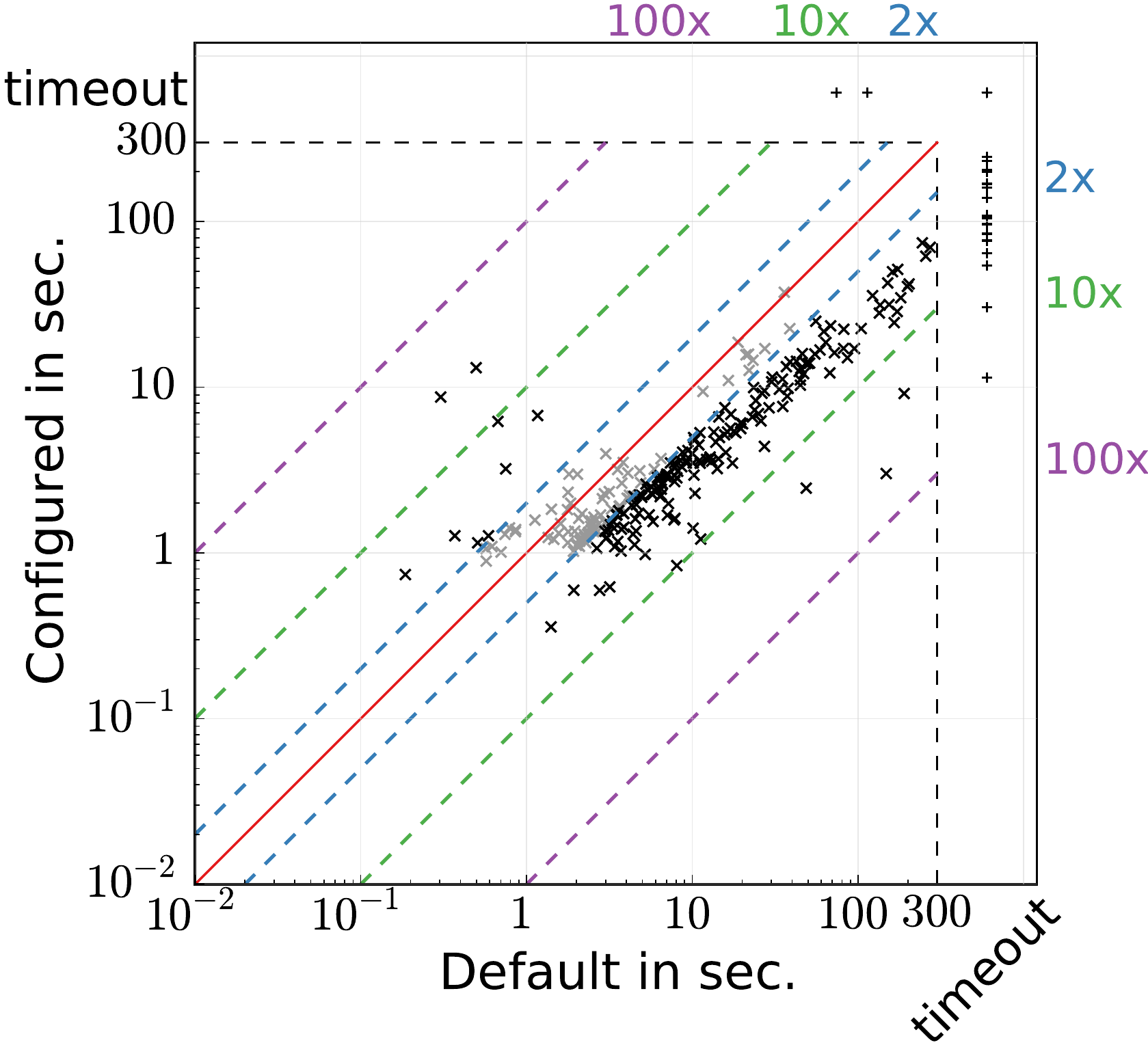}}\\
		\subfloat[\ibm{}\newline PAR-10: $694 \to
		692$]{\includegraphics[width=0.4\textwidth]{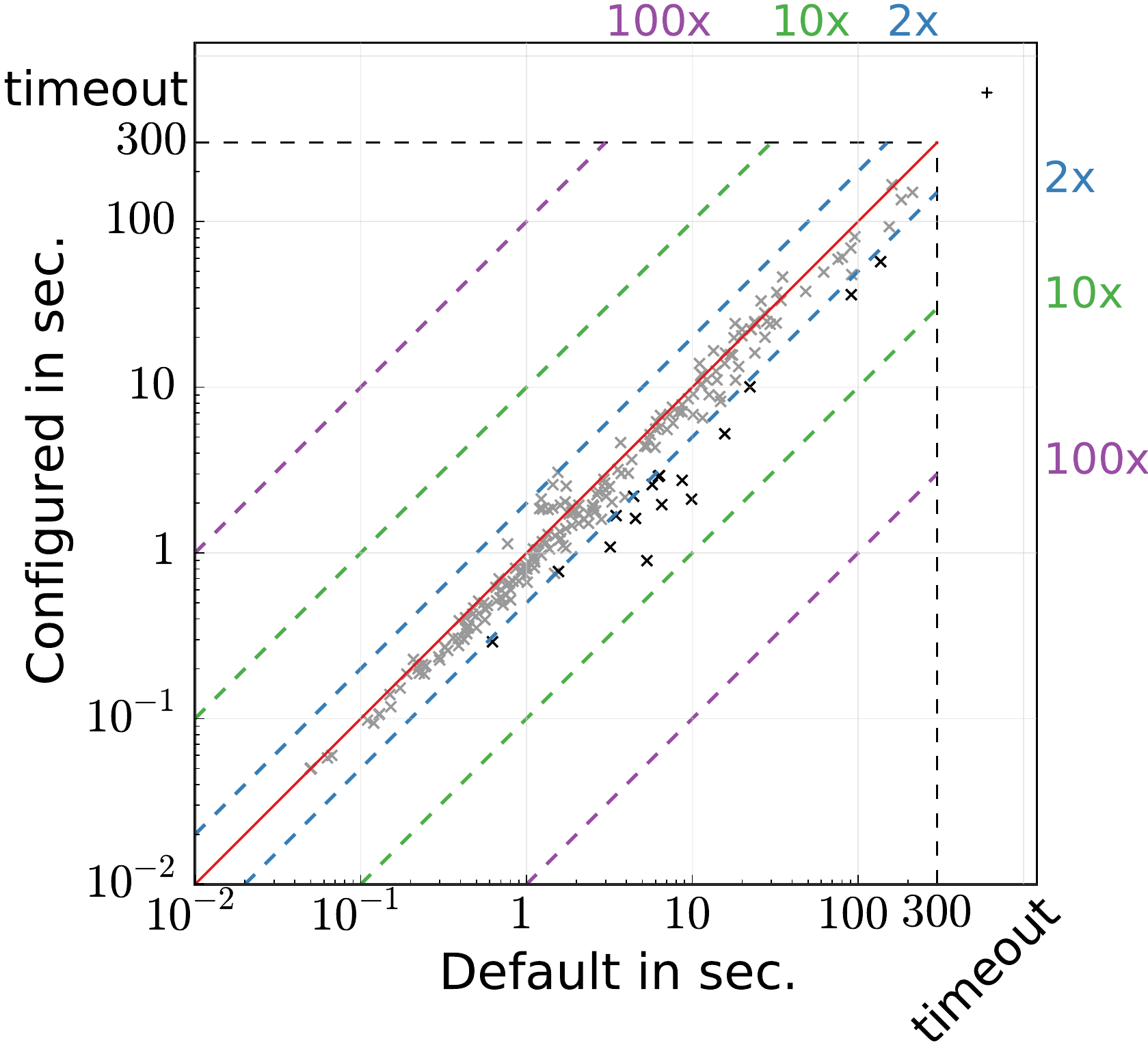}} \subfloat[\swvshort{}\newline PAR-10: $3.32 \to
		0.16$]{\includegraphics[width=0.4\textwidth]{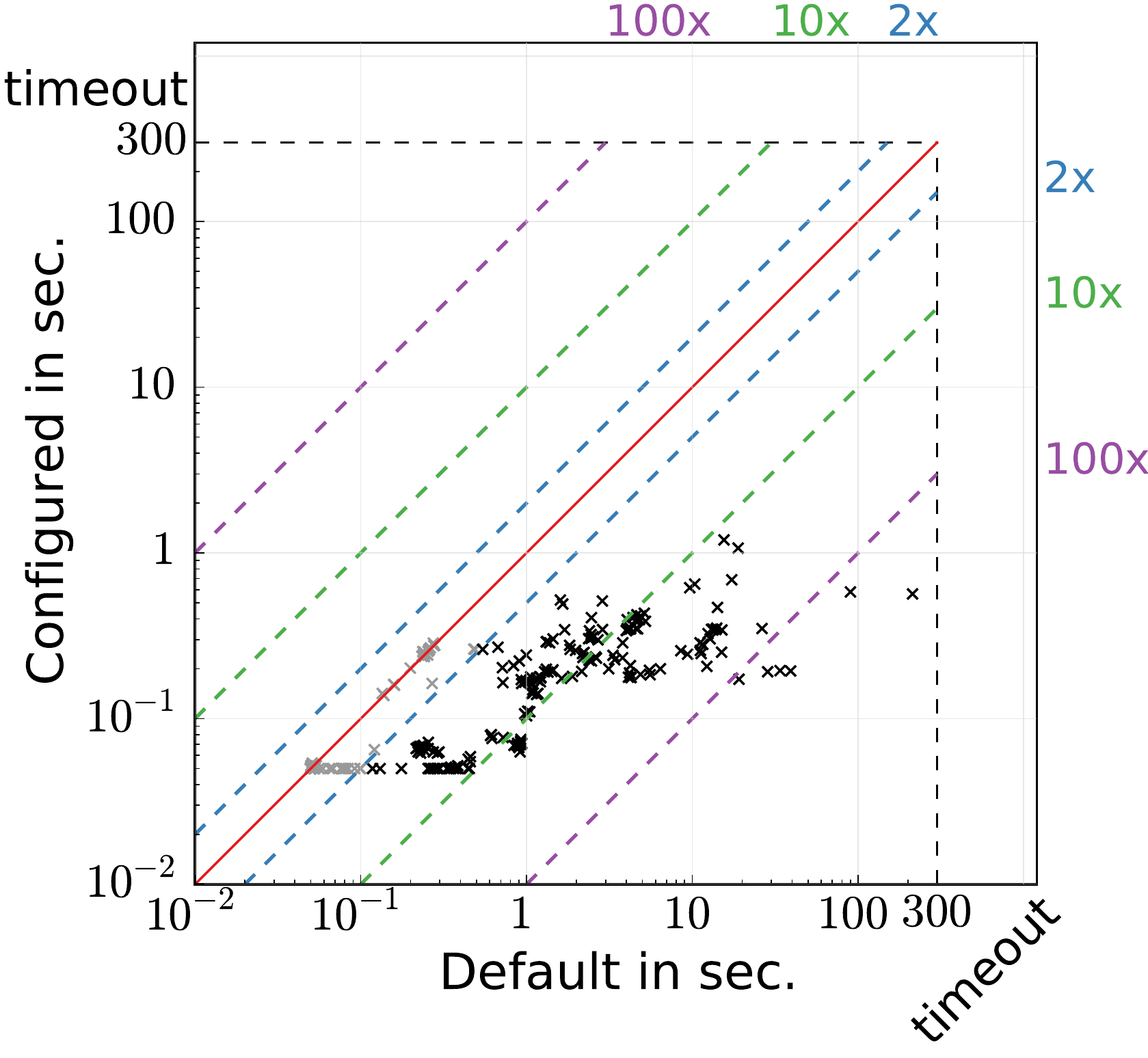}}
		\captionsetup{justification=justified}
		\caption{Speedups achieved by configuration of \lingeling. For each benchmark, we show scatter plots of solver defaults vs.\ configured parameter settings.\label{fig:indu13:lingeling}}
		\end{figure}
		\vspace*{-0.5cm}
		\begin{table}[H]
		\setlength\tabcolsep{0.35em}
		\sffamily\footnotesize\centering
		\begin{tabular}{l | cccc | c | cc}
		\toprule[1.0pt]
		 & \multicolumn{5}{c|}{$\#$timeouts default $\to$ $\#$ timeouts configured (on test set)} & \multicolumn{2}{c}{Rank} \\
		 & \bmcshort & \circuit{} & \ibm & \swvshort & \textit{Overall} & def & \hspace*{-0.2cm}\scriptsize{CSSC}\hspace*{-0.2cm}\\
		\midrule
		\myrowcolour{}\lingeling & $28 \to \mathbf{26}$ & $39 \to \mathbf{20}$ & $69 \to 69$ & $0 \to 0$ & $136 \to \mathbf{115}$ & 4 & 1\\
		\rissg & $32 \to \mathbf{30}$ & $20 \to \mathbf{18}$ & $70 \to \mathbf{69}$ & $0 \to 0$ & $122 \to \mathbf{117}$ & 1 & 2\\
		\myrowcolour{}\SolverFourtyThree & $30 \to 30$ & $20 \to 20$ & $77 \to 77$ & $0 \to 0$ & $127 \to 127$ & 2 & 3\\
		\forlnodrup & $50 \to \mathbf{36}$ & $33 \to \mathbf{23}$ & $69 \to 69$ & $0 \to 0$ & $152 \to \mathbf{128}$ & 5 & 4\\
		\myrowcolour{}\simpsat & $38 \to \mathbf{35}$ & $26 \to \mathbf{24}$ & $70 \to \mathbf{69}$ & $0 \to 0$ & $134 \to \mathbf{128}$ & 3 & 5\\
		\clasp & $66 \to \mathbf{42}$ & $26 \to \mathbf{17}$ & $71 \to 71$ & $0 \to 0$ & $163 \to \mathbf{130}$ & 6 & 6\\
		\myrowcolour{}\satj & $70 \to 70$ & $36 \to \mathbf{30}$ & $77 \to \mathbf{76}$ & $1 \to \mathbf{0}$ & $184 \to \mathbf{176}$ & 7 & 7\\
		\gnoveltyGCwa & $291 \to \mathbf{285}$ & $301 \to \mathbf{295}$ & $295 \to 295$ & $244 \to \mathbf{215}$ & $1131 \to \mathbf{1090}$ & 10 & 8\\
		\myrowcolour{}\gnoveltyPCL & $289 \to \mathbf{288}$ & $302 \to 302$ & $295 \to \mathbf{294}$ & $215 \to 215$ & $1101 \to \mathbf{1099}$ & 8 & 9\\
		\gnoveltyGCa & $291 \to \mathbf{290}$ & $\mathbf{300} \to 302$ & $295 \to 295$ & $243 \to \mathbf{217}$ & $1129 \to \mathbf{1104}$ & 9 & 10\\
		\bottomrule[1.0pt]
		\end{tabular}
		\caption{Results for CSSC 2013 competition track \indu. For each solver and benchmark, we show the number of test set timeouts achieved with the default and the configured parameter setting, bold-facing the better one; we broke ties by the solver's average runtime (not shown for brevity). We aggregated results across all benchmarks to compute the final ranking.\label{tab:indu13}}
		\end{table}
	\end{minipage}	
}
\end{figure}

Our \indu{} track consisted of the four industrial benchmarks detailed in \ref{sec:benchmarks-indu}: \bmc~\cite{biere-misc07a}, \circuit~\cite{brummayer-sat10a}, \hw~\cite{zarpas-sat05a}, and \swv~\cite{babic-hvc07}.
%

Figure \ref{fig:indu13:lingeling} visualizes the results of the configuration process for the winning solver \lingeling{} on these four benchmark sets.
It demonstrates that even \lingeling{}, a highly competitive solver in terms of default performance, can be configured for improved performance on a wide range of benchmarks. We note that for the easy benchmark \swvshort, configuration sped up \lingeling{} by a factor of $20$ (average runtime 3.3s \vs{} 0.16s), and that for the harder \circuit{} instances, it nearly halved the number of timeouts (39 \vs{} 20). The improvements were smaller for more traditional hardware verification instances (\ibm{} and \bmcshort) similar to those used to determine \lingeling{}'s default parameter settings.

Table \ref{tab:indu13} summarizes the results of the ten solvers that were eligible for medals.
%
From this table, we note that, like \lingeling{}, many other solvers benefited from configuration.
Indeed, some solvers (in particular \forlnodrup{} and \clasp{}) benefited much more from configuration on the \bmcshort{} instances, largely because their default performance was worse on this benchmark. 
On the other hand, \rissg{} featured stronger default performance than \lingeling{} but did not benefit as much from configuration.

Table \ref{tab:indu13} also aggregates results across the four benchmark families to yield the overall results for the \indu{} track.
These results show that many solvers benefited substantially from configuration, and that some benefited more than others, causing the CSSC ranking to differ substantially from the ranking according to default
solver performance; for instance, based on default performance, the overall winning solver, \lingeling{}, would have only ranked fourth.  


\subsubsection{Results of the \crafted{} Track} \label{sec:cssc2013-results-crafted}

\begin{figure}[tb]
\fbox{
	\noindent\begin{minipage}[t]{0.96\linewidth}
		\begin{center}\bf{}\underline{Results for CSSC 2013 \crafted{} track}\end{center}\vspace*{-0.7cm}
		\begin{figure}[H]
		\centering
		\captionsetup{justification=centering,captionskip=0cm}
		\subfloat[\gi{}{}\newline PAR-10: $362 \to
		65$]{\includegraphics[width=0.4\textwidth]{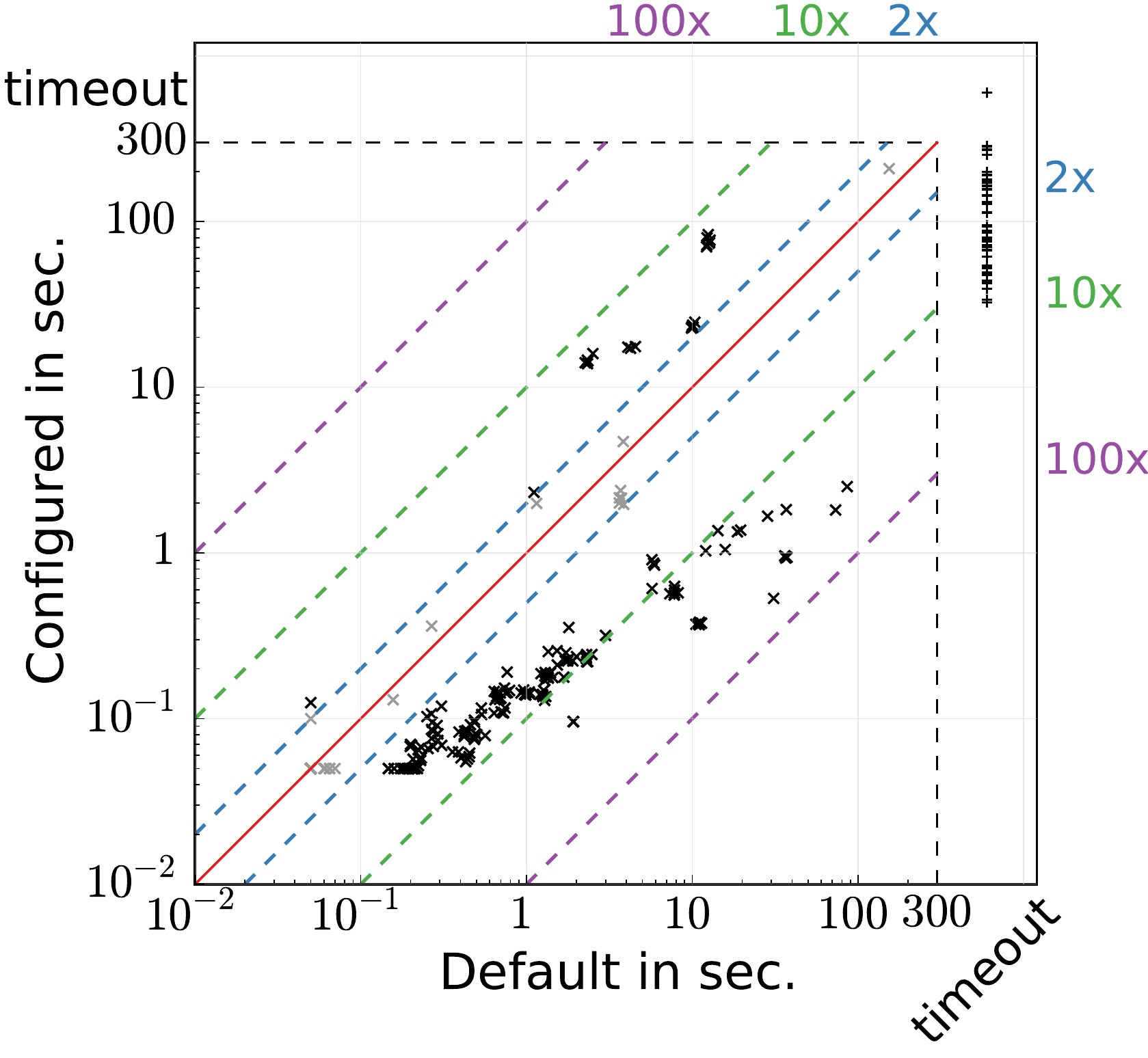}} \subfloat[\labs{}{}.\newline PAR-10: $837 \to
		779$]{\includegraphics[width=0.4\textwidth]{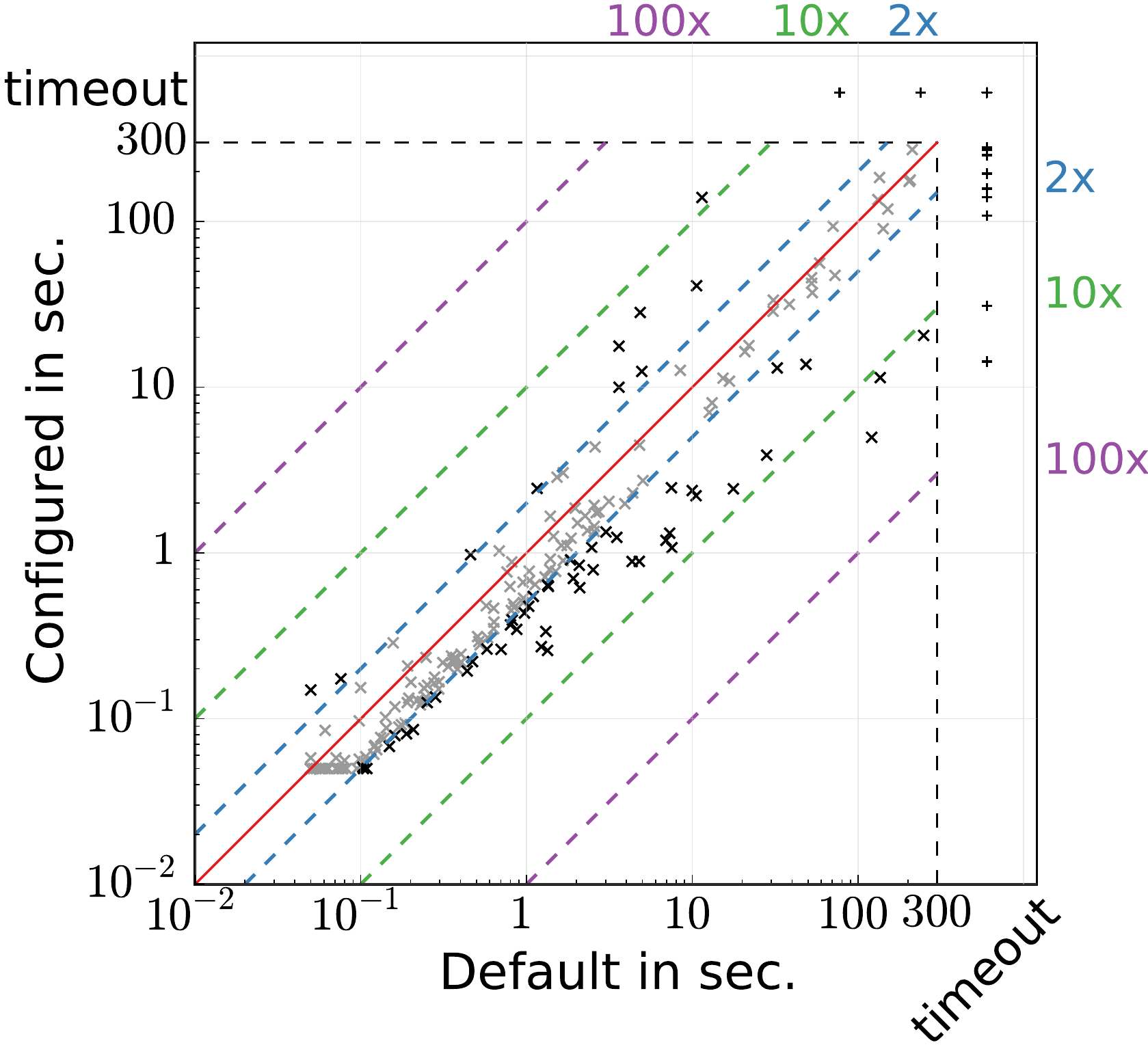}}
		\captionsetup{justification=justified}
		\caption{Speedups achieved by configuration of \clasp{} on the CSSC 2013 \crafted{} track. We show scatter plots of default vs.\ configured versions of \clasp{}.\label{fig:clasp-crafted-2013}}
		\end{figure}
		\vspace*{-0.5cm}
		\begin{table}[H]
		\setlength\tabcolsep{0.4em}
		\sffamily\small\centering
		\begin{tabular}{l | cc | c | cc}
		\toprule[1.0pt]
		 & \multicolumn{3}{c|}{$\#$TOs default $\to$ $\#$TOs configured (on test set)} & \multicolumn{2}{c}{Rank} \\
		 & \gishort & \labsshort & \textit{Overall} & def & {\scriptsize{CSSC}}\\
		\midrule
		\myrowcolour{}\clasp & $42 \to \mathbf{6}$ & $97 \to \mathbf{90}$ & $139 \to \mathbf{96}$ & 2 & 1\\
		\forlnodrup & $40 \to \mathbf{7}$ & $95 \to \mathbf{91}$ & $135 \to \mathbf{98}$ & 1 & 2\\
		\myrowcolour{}\lingeling & $43 \to \mathbf{10}$ & $105 \to \mathbf{97}$ & $148 \to \mathbf{107}$ & 3 & 3\\
		\rissg & $51 \to \mathbf{42}$ & $97 \to \mathbf{89}$ & $148 \to \mathbf{131}$ & 4 & 4\\
		\myrowcolour{}\simpsat & $42 \to 42$ & $107 \to 107$ & $149 \to 149$ & 5 & 5\\
		\SolverFourtyThree & $66 \to \mathbf{65}$ & $90 \to \mathbf{87}$ & $156 \to \mathbf{152}$ & 6 & 6\\
		\myrowcolour{}\satj & $62 \to \mathbf{57}$ & $110 \to \mathbf{104}$ & $172 \to \mathbf{161}$ & 7 & 7\\
		\gnoveltyGCwa & $180 \to 180$ & $195 \to \mathbf{154}$ & $375 \to \mathbf{334}$ & 8 & 8\\
		\myrowcolour{}\gnoveltyGCa & $183 \to \mathbf{180}$ & $240 \to \mathbf{173}$ & $423 \to \mathbf{353}$ & 10 & 9\\
		\gnoveltyPCL & $179 \to \mathbf{178}$ & $199 \to \mathbf{183}$ & $378 \to \mathbf{361}$ & 9 & 10\\
		\bottomrule[1.0pt]
		\end{tabular}
		\caption{Results for CSSC 2013 competition track \crafted. For each solver and benchmark, we show the number of test set timeouts achieved with the default and the configured parameter setting, bold-facing the better one. We aggregated results across all benchmarks to compute the final ranking.	\label{tab:crafted2013}}
		\end{table}
	\end{minipage}
}		
\end{figure}

The \crafted{} track consisted of the two crafted benchmarks detailed in \ref{sec:benchmarks-crafted}: \gi{} and \labs{}.

Figure \ref{fig:clasp-crafted-2013} visualizes the improvements algorithm configuration yielded for the best-performing solver \clasp{} on these benchmarks.
Improvements were particularly large on the \gishort{} instances, where algorithm configuration decreased the number of timeouts from 42 to 6.
Table \ref{tab:crafted2013} summarizes the results we obtained for all solvers on these benchmarks, showing that configuration also substantially improved the performance of many other solvers. The table also aggregates results across both benchmark families to yield overall results for the \crafted{} track. 
While  \forlnodrup{} showed the best default performance and benefited substantially from configuration (\#timeouts reduced from 135 to 98), \clasp{} improved even more (\#timeouts reduced from 139 to 96).

\begin{figure}[tb]
\fbox{
	\noindent\begin{minipage}[t]{0.96\linewidth}
		\begin{center}\bf{}\underline{Results for CSSC 2013 \random{} track}\end{center}\vspace*{-0.7cm}
		\begin{figure}[H]
		\centering
		\captionsetup{justification=centering,captionskip=0cm}
		\subfloat[\gnoveltyGCa{} on \fivesatfiveh{}\newline PAR-10: $1997 \to
		77$]{\includegraphics[width=0.33\textwidth]{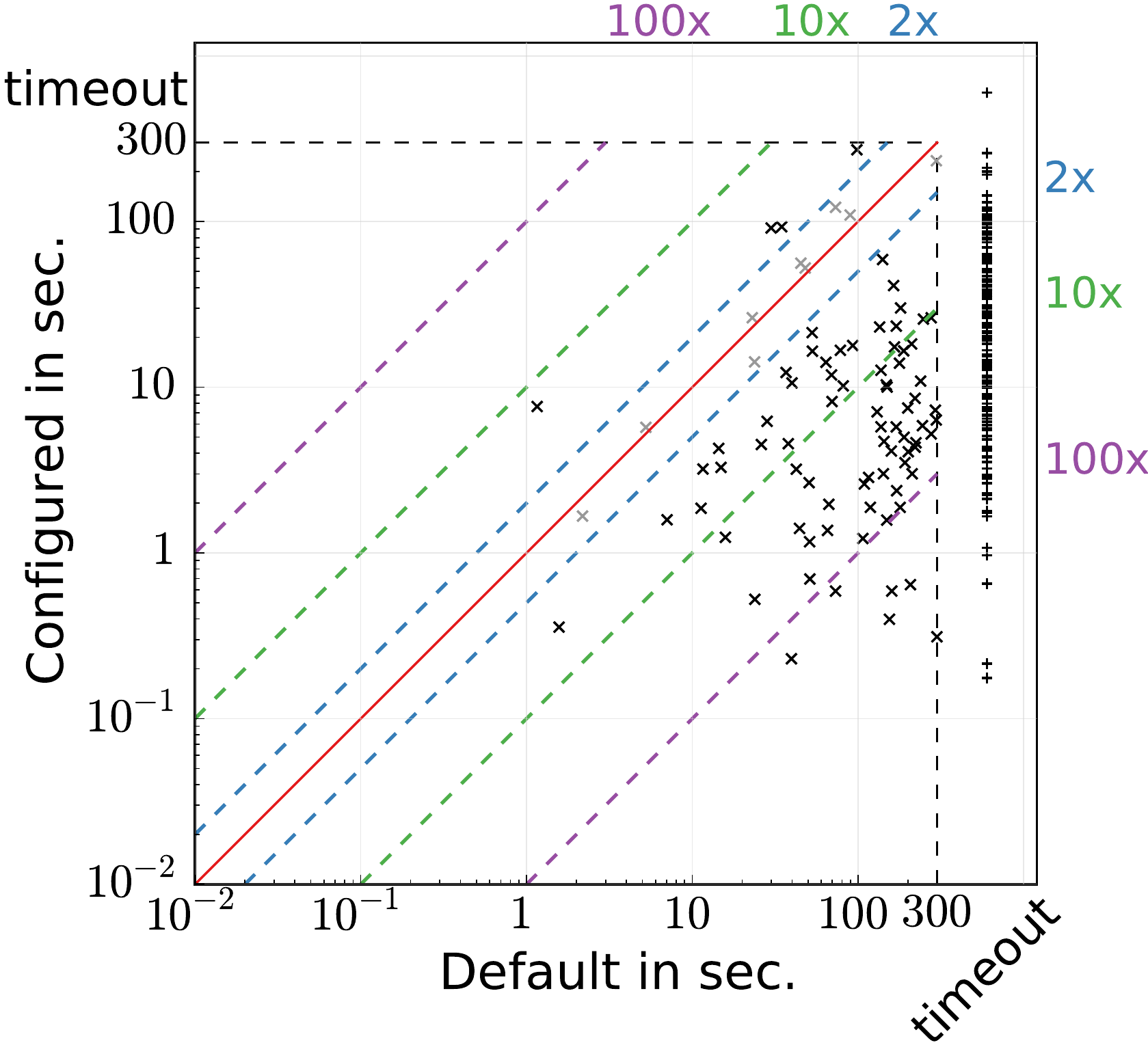}}
		\subfloat[\clasp{} on \kthree{}\newline PAR-10: $158 \to
		2.79$]{\includegraphics[width=0.33\textwidth]{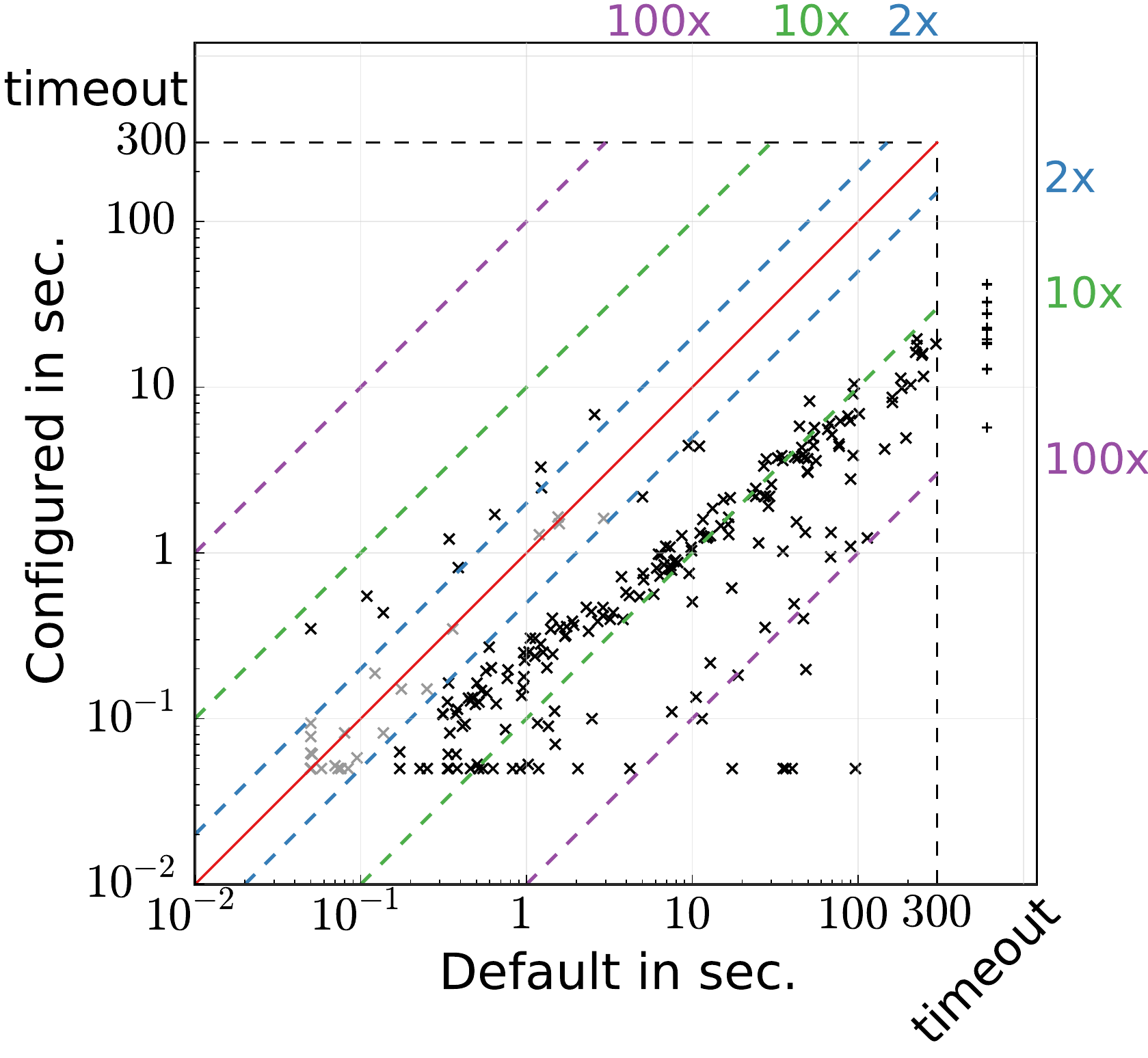}}
		\subfloat[\clasp{} on \unif{}\newline PAR-10: $1.44 \to
		0.37$]{\includegraphics[width=0.33\textwidth]{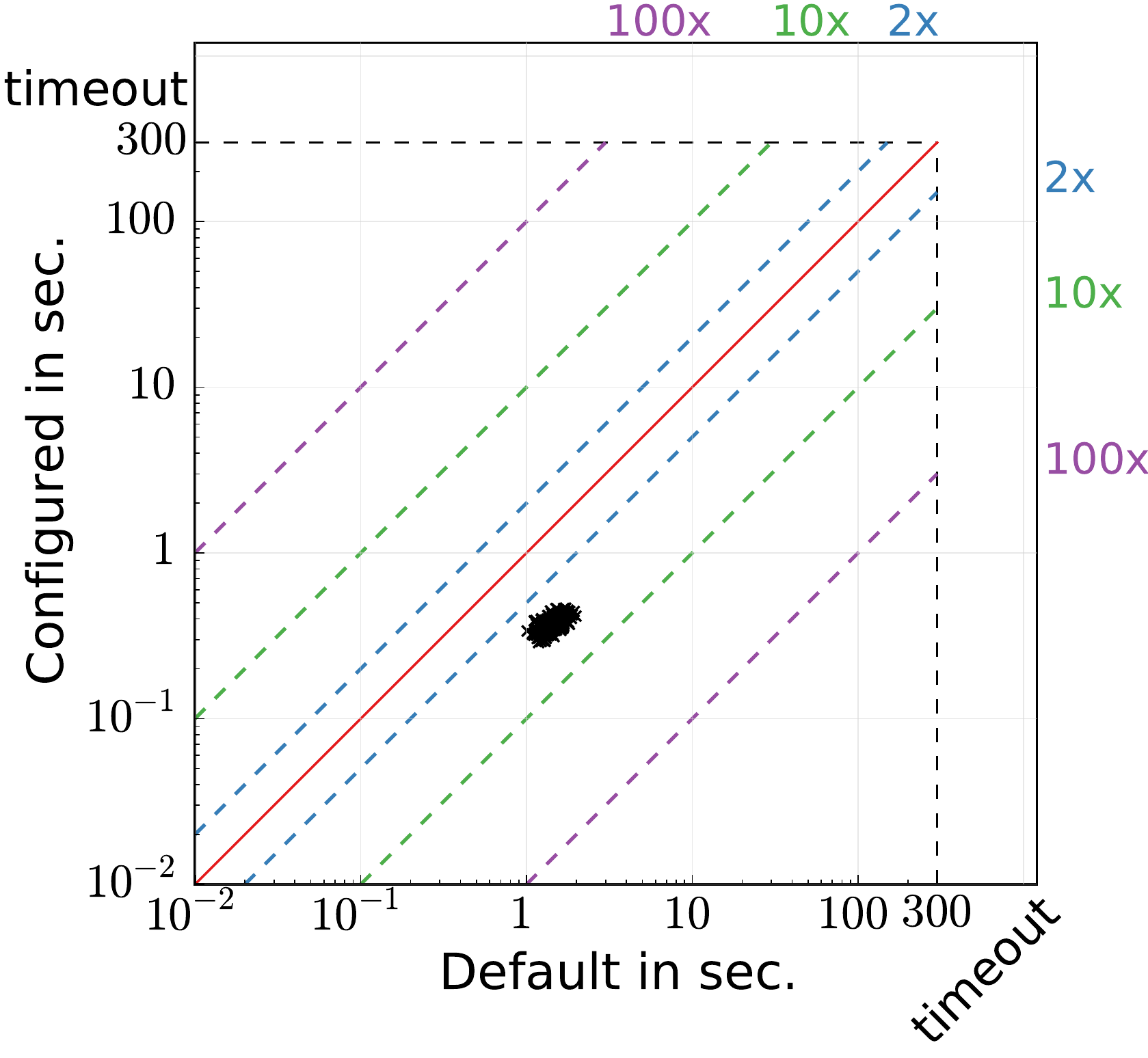}}
		\captionsetup{justification=justified}
		\caption{Speedups achieved by configuration on the CSSC 2013 \random{} track. We show scatter plots of default vs. configured solvers.\label{fig:random-2013}.}
		\end{figure}
		\vspace*{-0.5cm}
		\begin{table}[H]
		\setlength\tabcolsep{0.4em}
		\sffamily\footnotesize\centering
		\begin{tabular}{l | ccc | c | cc}
		\toprule[1.0pt]
		 & \multicolumn{4}{c|}{$\#$TOs default $\to$ $\#$TOs configured (on test set)} & \multicolumn{2}{c}{Rank} \\
		 & \fivesatfiveh & \kthree & \unif & \textit{Overall} & def & {\scriptsize{CSSC}}\\
		\midrule
		\clasp & $250 \to 250$ & $11 \to \mathbf{0}$ & $0 \to 0$ & $261 \to \mathbf{250}$ & 6 & 1\\
		\myrowcolour{}\lingeling & $250 \to 250$ & $8 \to \mathbf{0}$ & $0 \to 0$ & $258 \to \mathbf{250}$ & 4 & 2\\
		\myrowcolour{}\rissg & $250 \to 250$ & $10 \to \mathbf{0}$ & $0 \to 0$ & $260 \to \mathbf{250}$ & 5 & 3\\
		\SolverFourtyThree & $250 \to 250$ & $6 \to \mathbf{3}$ & $0 \to 0$ & $256 \to \mathbf{253}$ & 2 & 4\\
		\myrowcolour{}\simpsat & $250 \to 250$ & $4 \to 4$ & $0 \to 0$ & $254 \to 254$ & 1 & 5\\
		\satj & $250 \to 250$ & $7 \to \mathbf{5}$ & $0 \to 0$ & $257 \to \mathbf{255}$ & 3 & 6\\
		\myrowcolour{}\forlnodrup & $250 \to 250$ & $39 \to \mathbf{8}$ & $0 \to 0$ & $289 \to \mathbf{258}$ & 7 & 7\\
		\gnoveltyGCwa & $8 \to \mathbf{1}$ & $124 \to 124$ & $250 \to 250$ & $382 \to \mathbf{375}$ & 8 & 8\\
		\myrowcolour{}\gnoveltyGCa & $163 \to \mathbf{4}$ & $124 \to 124$ & $250 \to 250$ & $537 \to \mathbf{378}$ & 9 & 9\\
		\gnoveltyPCL & $250 \to \mathbf{11}$ & $124 \to 124$ & $250 \to 250$ & $624 \to \mathbf{385}$ & 10 & 10\\
		\bottomrule[1.0pt]
		\end{tabular}
		\caption{Results for CSSC 2013 competition track \random. For each solver and benchmark, we show the number of test set timeouts achieved with the default and the configured parameter setting, bold-facing the better one. Results were aggregated across all benchmarks to compute the final ranking. We broke ties by the solver's average runtime. While we do not show runtimes for brevity, the runtimes important for the ranking were the average runtimes of the top 3 solvers on the union of \kthree{} and \unif{}: $1.58$s (\clasp), $4.20$s (\lingeling), and $7.68$s (\rissg).}
		\label{tab:random2013}
		\end{table}
	\end{minipage}}	
\end{figure}

\subsubsection{Results of the \random{} Track} \label{sec:cssc2013-results-random}

The \random{} track consisted of three random benchmarks detailed in \ref{sec:benchmarks-random}: \fivesatfiveh{}, \kthree{}, and \unif{}.
The instances in \fivesatfiveh{} were all satisfiable, those in \unif{} all unsatisfiable, and those in \kthree{} were mixed.

Table \ref{tab:random2013} summarizes the results for these benchmarks. It shows that the \unif{} benchmark set was very easy for complete solvers (although configuration still yielded up to 4-fold speedups), that the \kthree{} benchmark was also quite easy for the best solvers, and that only the SLS solvers could tackle 
benchmark \fivesatfiveh{}, with configuration making a big difference to performance.

Here again, our aggregate results demonstrate that rankings were substantially different between the default and configured versions of the solvers: the three solvers with top default performance were ranked 4th to 6th in the CSSC, and vice versa. Figure \ref{fig:random-2013} visualizes the very substantial speedups achieved by configuration for the winning solver \clasp{} on \kthree{} and \unif{}, and for the SLS solver \gnoveltyGCa{} on \fivesatfiveh{}.

\section{The Configurable SAT Solver Challenge 2014} \label{sec:cssc2014}

\begin{table}
\sffamily\small\centering%
\begin{tabular}{@{}l l l l l r}
\toprule[1.0pt]

Benchmark & \#Train & \#Test & \#Variables & \#Clauses & Reference\\
\midrule

\hwshort & 383 & 302 & $96.4\text{k} \pm 170\text{k}$ & $413\text{k} \pm 717\text{k}$ & \cite{zarpas-sat05a}\\
\myrowcolour{}\circuit & 299 & 302 &$5.53\text{k} \pm 7.45\text{k}$ & $18.8\text{k} \pm 25.3\text{k}$ & \cite{brummayer-sat10a}\\
\bmcshort{} & 604 & 302 & $424\text{k} \pm 843\text{k}$ & $1.03\text{m} \pm 2.30\text{m}$ & \cite{biere-misc08a}\\

\midrule

\myrowcolour{}\gishort & 1032 & 351 & $11.2\text{k} \pm 17.8\text{k}$ & $2.98\text{m} \pm 8.03\text{m}$ & \cite{mugrauer2013,toran2013}\\
\labsshort & 350 & 351 & $75.9\text{k} \pm 75.7\text{k}$ & $154\text{k} \pm 153\text{k}$ & \cite{mugrauer2013-2}\\
\myrowcolour{}\queens & 484 & 351 & $38.2\text{k} \pm 37.4\text{k}$ & $125\text{k} \pm 126\text{k}$ & \cite{manthey-sat14r}\\

\midrule

\myrowcolour{}\kthree & 300 & 250 & $262 \pm 43$ & $1116 \pm 182$ & \cite{bayless-lion14a}\\
\threecnf{} & 500 & 250 & $350 \pm 0$ & $1493 \pm 0$ & \cite{bebel-sat13a}\\
\myrowcolour{}\unif & 300 & 250 & $50 \pm 0$ & $1056 \pm 0$ & --\\

\midrule

\threesatonek & 250 & 250 & $500 \pm 0$ & $10000 \pm 0$ & \cite{tompkins-sat11}\\
\myrowcolour{}\fivesatfiveh & 250 & 250 & $1000 \pm 0$ & $4260 \pm 0$ & \cite{tompkins-sat11}\\
\sevensatninety & 250 & 250 & $90 \pm 0$ & $7650 \pm 0$ & \cite{tompkins-sat11}\\

\bottomrule[1.0pt]

\end{tabular}
\caption{Overview of benchmark sets used in the CSSC 2014 tracks \indu{}, \crafted{}, \random{}, and \randomsat{} (from top to bottom); k and m stand for factors of one thousand and one million, respectively.
\label{tab:benchmarks-2014}}
\end{table}

The second CSSC\footnote{\url{http://aclib.net/cssc2014/}} was held in 2014.
Compared to the inaugural CSSC in 2013, we improved the competition design in several ways:
\begin{itemize}
\denselist
	\item We used a different computer cluster,\footnote{We executed this competition on the META cluster at the University of Freiburg, whose compute 
nodes contained 64GB of RAM and two 2.60GHz Intel Xeon E5-2650v2 8-core CPUs with 20 MB L2 cache each, running Ubuntu 14.04 LTS, 64bit.} enabling us to run \gga{} as one of the configuration procedures.
	\item We added a \randomsat{} track to facilitate comparisons of stochastic local search solvers.
	\item We dropped the (too easy) SWV benchmark family and introduced four new benchmark families, yielding a total of three benchmark families in each of the four tracks, summarized in Table \ref{tab:benchmarks-2014} and described in detail in \ref{app:benchmarks}.
  \item We let solver authors decide which tracks their solver should run in.
	\item For fairness, for each solver, we performed the same number of configuration experiments. (This is in contrast to 2013, where we performed the same number of configuration runs \emph{for every configuration space} of every solver, which lead to a larger combined configuration budget for solvers submitted with multiple configuration spaces).
	\item We kept track of all of the (millions of) solver runs performed during the configuration process and made all information about errors available to solver developers \changed{after the competition.} 
\end{itemize}

\subsection{Participating Solvers} \label{sec:cssc2014-solvers}

\begin{table}
\sffamily\footnotesize\centering%
\setlength\tabcolsep{0.3em}
\begin{tabular}{@{}l rrrr r r r r}
\toprule[1.0pt]

Solver & \multicolumn{4}{c}{\# Parameters} & \multicolumn{2}{c}{\# Configurations} & Categories & Ref.\\
       & c & i & r & cond. 				& discretized & original  				   & 			& 		\\
\midrule

\dccsat & $1$ & $0$ & $0$ & $0$ & $9$ & $9$ & Random & \cite{dccsat}\\
\myrowcolour{}\csccsat & $3$ & $0$ & $0$ & $0$ & $567$ & $567$ & Random SAT & \cite{csccat1,dccsat}\\
\probsat & $5$ & $1$ & $3$ & $4$ & $1 \times 10^{5}$ & $\infty$ & Random SAT & \cite{probSAT}\\
\myrowcolour{}\minisathack & $10$ & $0$ & $0$ & $3$ & $8 \times 10^{5}$ & $8 \times 10^{5}$ & All categories & \cite{minisat_hack}\\
\yalsat & $16$ & $10$ & $0$ & $0$ & $5 \times 10^{6}$ & $2 \times 10^{72}$ & Crafted \!\!\&\!\! Random SAT & \cite{lingelingyalsat}\\
\myrowcolour{}\cryptominisat & $14$ & $15$ & $7$ & $2$ & $3 \times 10^{24}$ & $\infty$ & Industrial \& Crafted & \cite{cryptominisat}\\
\clasp & $38$ & $30$ & $7$ & $55$ & $1 \times 10^{49}$ & $\infty$ & All categories & \cite{clasp}\\
\myrowcolour{}\riss & $214$ & $0$ & $0$ & $160$ & $5 \times 10^{86}$ & $5 \times 10^{86}$ & All but Random SAT & \cite{riss}\\
\sparrow & $170$ & $36$ & $16$ & $176$ & $1 \times 10^{112}$ & $\infty$ & All categories & \cite{sparrowtoriss}\\
\myrowcolour{}\lingeling & $137$ & $186$ & $0$ & $0$ & $1 \times 10^{53}$ & $2\! \cdot\! 10^{1341}$ & All categories & \cite{lingelingyalsat}\\

\bottomrule[1.0pt]

\end{tabular}
\caption{Overview of solvers in the CSSC 2014 and their parameters of various types (`c' for categorical, `i' for integer, `r' for real-valued'); `cond' identifies how many of these parameters are conditional. For each solver, we also list the sizes of the original configuration space submitted by solver developers and of a discretized version, as well as the categories in which the solver participated. Solvers are ordered by the number of parameters they expose ($c+i+r$).\label{tab:solvers2014}}

\end{table}

The ten solvers that participated in the CSSC 2014 are summarized in Table~\ref{tab:solvers2014};
they included CDCL, SLS and hybrid solvers.
These solvers differed substantially in their degree of parameterization, with the number of parameters 
ranging from 1 to 323.
We briefly discuss the main features of each solver's parameter configuration space, ordering solvers by their number of parameters.

\myparagraph{\dccsat{}}~\cite{dccsat} combines the SLS solver DCCASat with the CDCL solver march-rw. It was submitted to the \random{} track. Its only
(continuous) parameter is the time ratio of the SLS solver. This parameter was pre-discretized to nine values. 

\myparagraph{\csccsat{}}~\cite{csccat1,dccsat} is an SLS solver based on configuration checking and dynamic local search methods. It was submitted to the \randomsat{} track. 
It features 3 continuous parameters that were pre-discretized to 7, 9, and 9 values each, giving rise
to a total configuration space of 567 possible parameter configurations.
The parameters control the weighting of the dynamic local search part and the probabilities for the linear make functions used in the random walk steps. 

\myparagraph{\probsat}~\cite{probSAT} is a simple SLS solver based on 
probability distributions that are built from simple features, such as the make and break of variables~\cite{probSAT}.
\probsat{}'s 9 parameters control the type 
and the parameters of the probability distribution, as well as the type of restart. 
\probsat{} was submitted to the \randomsat{} track.

\myparagraph{\minisathack{}}~\cite{minisat_hack} is a CDCL solver; it was submitted to all tracks.
It has one categorical parameter (whether or not to use the Luby restarting strategy) and 9 
numerical parameters fine-tuning the Luby and geometric restart strategies, as well as controlling clause removal and the treatment of glue clauses. 
3 of these 9 numerical parameters are conditional on the choice of the Luby restart strategy, and all numerical parameters were pre-discretized by the solver developer. 
There are also 3 forbidden parameter combinations derived from a weak inequality constraint between two parameter values.


\myparagraph{\yalsat{}}~\cite{lingelingyalsat} is an SLS solver; it was submitted to the tracks \crafted{} and \randomsat{}.
It has 27 parameters that parameterize the solver's restart component (7 parameters) amongst many other components.
11 of the 27 parameters are numerical, with 6 of them having a trivial upper bound of max-integer ($2^{31}-1$).

\myparagraph{\cryptominisat{}}~\cite{cryptominisat} is a CDCL solver; it was submitted to the tracks \indu{} and \crafted{}.
It has 29 parameters that control restarts (6 mostly numerical parameters), clause removal (7 mostly numerical parameters),
variable branching and polarity (3 parameters each), simplification (5 parameters), and several other mechanisms.
2 of the numerical parameters further parameterize the blocking restart mechanism and are thus conditional on that mechanism being selected.

\myparagraph{\clasp{}}~\cite{clasp} is a solver for the more general answer set programming (ASP) problem, but it can also solve SAT, MAXSAT and PB problems.
It is fundamentally similar to the solver submitted in 2013; changes in the new version focused on the ASP solving part rather than the SAT solving part.
As a SAT solver, \clasp{} has $75$ parameters, of which $7$ control preprocessing,
$14$ variable selection, $19$ the restart policy, 
$28$ the deletion policy
and $7$ miscellaneous other mechanisms. 
The configuration space is highly conditional,
with several top-level parameters enabling or disabling certain strategies.
Finally, there are also $2$ forbidden parameter combinations that prevent certain combinations of deletion strategies.
\clasp{} exposes both a mixed continuous/discrete parameter configuration space and a manually-discretized one.
It was submitted to all tracks.

\myparagraph{\riss{}}~\cite{riss} is a CDCL solver submitted to all tracks except \randomsat{}.
Compared to the 2013 version \rissg{}, it almost doubled its number of parameters, yielding 214 parameters organized into 121 simplification and 93 search parameters.
In particular, it added many new preprocessing and inprocessing techniques, including XOR handling (via Gaussian elimination~\cite{han2012boolean}), and extracting cardinality constraints~\cite{MlazyCard}.
Roughly half of the simplification parameters and a third of the search parameters are categorical (in both cases most of the categoricals are binary).
The simplification parameters comprise about 20 Boolean switches for preprocessing techniques 
and about 100 in-processor parameters, prominently including blocked clause elimination, bounded variable addition, equivalance elimination~\cite{equivalence-reasoning}, numerical limits, 
probing, symmetry breaking, unhiding~\cite{unhiding},  Gaussian elimination, covered
literal elimination~\cite{covered-literal-elimination}, and even some stochastic local search.
The search parameters parameterize a wide range of mechanisms including variable selection, clause learning and removal, restarts, clause minimization,
restricted extended resolution,
and interleaved clause strengthening.

\myparagraph{\sparrow{}}~\cite{sparrowtoriss} combines the SLS solver \emph{Sparrow} with the CDCL solver \riss{} by first running \emph{Sparrow}, followed by \riss{}. It was submitted to all tracks.
\sparrow{}'s configuration space is that of \riss{} plus 6 \emph{Sparrow} parameters and 2 parameters controlling when to switch from \emph{Sparrow} to \riss{}: the maximal number of flips for \emph{Sparrow} (by default 500 million) and the CPU time for \emph{Sparrow} (by default 150 seconds).
Also, in contrast to \riss{}, \sparrow{} does not pre-discretize its numerical parameters, but expresses them as 36 integer and 16 continuous parameters.

\myparagraph{\lingeling{}}~\cite{lingelingyalsat} is a successor to the 2013 version; it was submitted to the tracks \indu{} and \crafted{}.
Compared to 2013, \lingeling{}'s parameter space grew by roughly a third, to a total of 323 parameters (meaning that again, \lingeling{} was the solver with the most parameters). As in 2013, roughly 40\% of these parameters were categorical and the rest integer-valued (many with a trivial upper bound of max-integer, $2^{31}-1$). Notable groups of parameters that were introduced in the 2014 version include additional preprocessing/inprocessing options and new restart strategies. 

\subsection{Configuration Pipeline} \label{sec:cssc2014-pipeline}

In the CSSC 2014, we used the configurators \paramils{}, \gga{}, and \smac{}.
For each benchmark and solver, we ran \gga{} and \smac{} on the solver's full configuration space, which could contain an arbitrary combination of numerical and categorical parameters.
We also ran all configurators on a discretized version of the configuration space (automatically constructed unless provided by the solver authors),
yielding a total of five configuration approaches: \paramils{}-discretized, \gga{}, \gga{}-discretized, \smac{}, and \smac{}-discretized. 
\gga{} could not handle the complex conditionals of some solvers;
therefore, for these solvers we only ran \paramils{} and the two \smac{} variants.

Due to the cost of running a third configurator on nearly every configuration scenario, we reduced the budget for each configuration approach from two CPU days on five cores in CSCC 2013 to 
two CPU days on four cores in CSSC 2014. 
In the case of \paramils{} and \smac{}, as in 2013, we used these four cores to perform four independent 2-day configurator runs.
In the case of \gga{}, we performed one 2-day run using all four cores.
We evaluated the configurations resulting from each of the 14 configuration runs (4 \paramils{}-discretized, 4 \smac{}-discretized, 4 \smac{}, 1 \gga{}-discretized, and 1 \gga{}) on the entire training data set of the benchmark at hand and selected the configuration with the best performance. We then executed only this configuration on the benchmark's test set to determine the performance of the configured solver.

\changed{In the four tracks of the CSSC (\indu{}, \crafted{}, \random{}, \randomsat{}) we had 6, 6, 5, and 6 participating solvers, respectively, and since there were three benchmark families per track, we ended up with $(6+6+5+6)\times 3 = 72$ pairs of solvers and benchmarks to configure them on. For each of these configuration scenarios, each of the 5 configuration approaches above required four cores for 2 days, yielding a total computational expense of $72 \times 5\times 4 \times 2 = 2880$ CPU days (close to 8 CPU years). Thanks to a special allocation on the META cluster at the University of Freiburg, we were able to finish this process within 2 weeks.}

\changed{We note that all scripts we used for performing the configuration and analysis experiments were written in Python (updated from Ruby in 2013) and are available for download on the competition website.}

\subsection{Results} \label{sec:cssc2014-results}

\begin{table}[tb]
		\setlength\tabcolsep{0.3em}
		\sffamily\footnotesize\centering
		\begin{tabular}{l | llll}
		\toprule[1.0pt]
		Rank & \indu{} & \crafted & \random & \randomsat\\
		\midrule
		1$^{st}$ & \myrowcolour{}\lingeling & \clasp & \clasp & \probsat\\
		2$^{nd}$ & \minisathack & \lingeling & \dccsat & \sparrow\\
		3$^{rd}$ & \myrowcolour{}\clasp & \cryptominisat & \minisathack & \csccsat\\
		\bottomrule[1.0pt]
		\end{tabular}
		\caption{Winners of the four tracks of CSSC 2014.\label{tab:results14}}
\end{table}
		
For each of the four tracks of CSSC 2014, we configured the solvers submitted to
the track on each of the three benchmark families from that track and aggregated results across the
respective test instances.
We show the winners for each track in Table \ref{tab:results14} and discuss	the results in the following sections.
Additional details, tables, and figures are provided in an accompanying technical
report~\cite{hutter-tech14b}.

\subsubsection{Results of the \indu{} Track} \label{sec:cssc2014-results-indu}

\begin{figure}[t]
\fbox{
	\noindent\begin{minipage}[t]{0.96\linewidth}
		\begin{center}\bf{}\underline{Results for CSSC 2014 \indu{} track}\end{center}\vspace*{-0.7cm}
		\begin{figure}[H]
		\centering
		\captionsetup{justification=centering,captionskip=0cm}
		\subfloat[\bmcshort{}, PAR-10: $222 \to 221$][\bmcshort{}\\ PAR-10: $222 \to
		221$]{\includegraphics[width=0.333\textwidth]{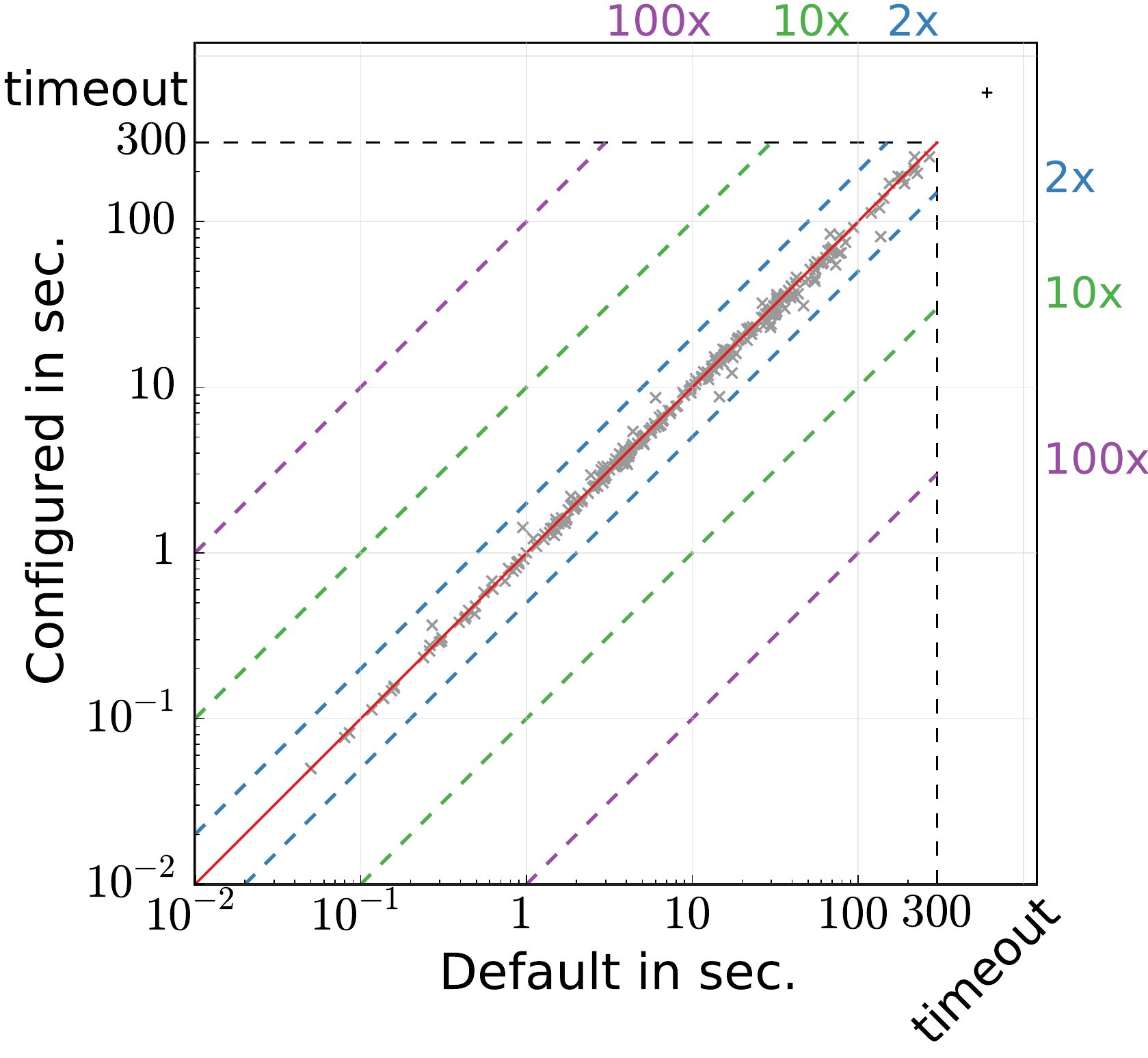}}
		\subfloat[\circuit{}, PAR-10: $316 \to 193$][\circuit{}\\ PAR-10: $316 \to
		193$]{\includegraphics[width=0.333\textwidth]{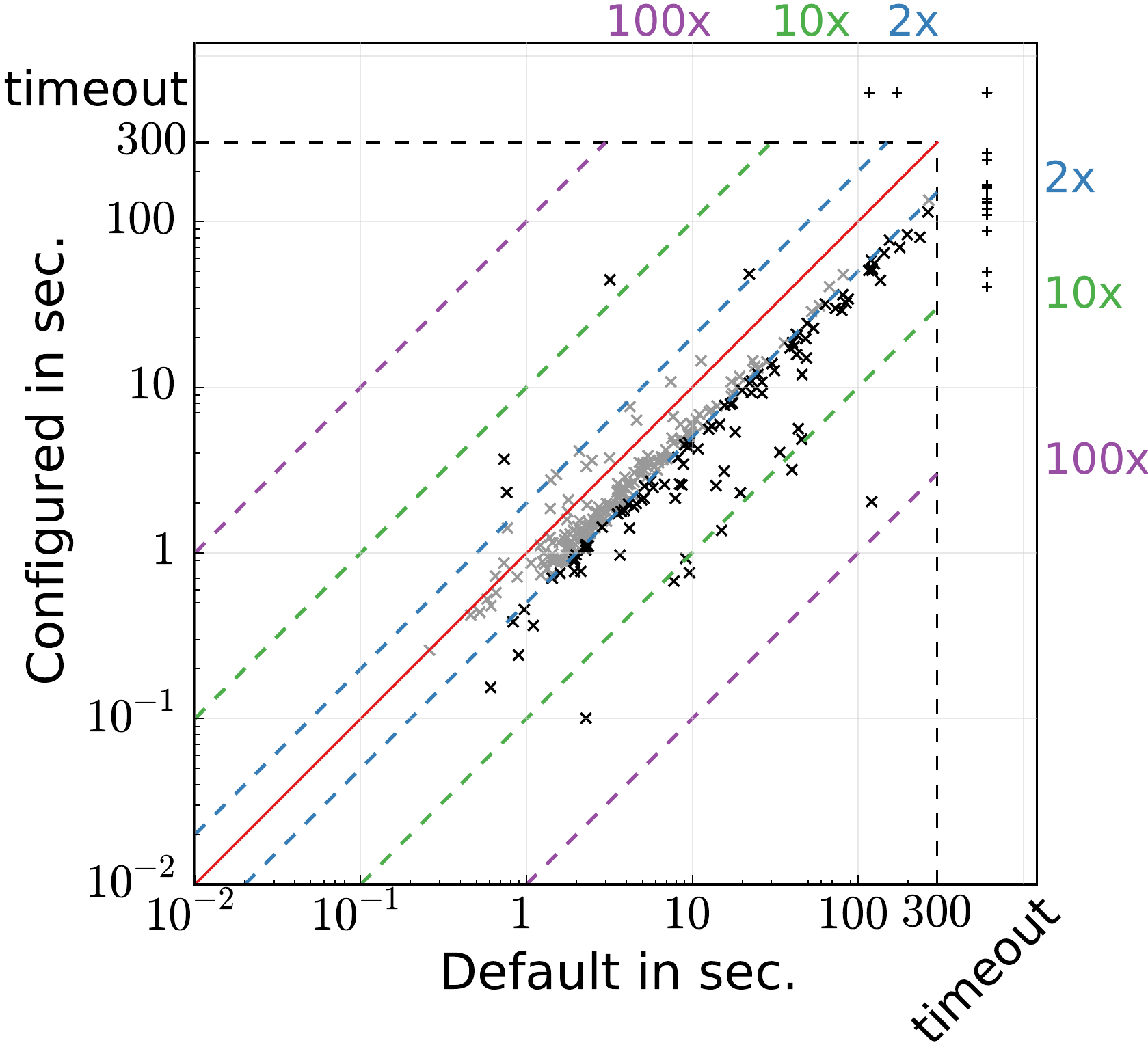}}
		\subfloat[\ibm{}, PAR-10: $697 \to 694$][\ibm{}\\ PAR-10: $697 \to
		694$]{\includegraphics[width=0.333\textwidth]{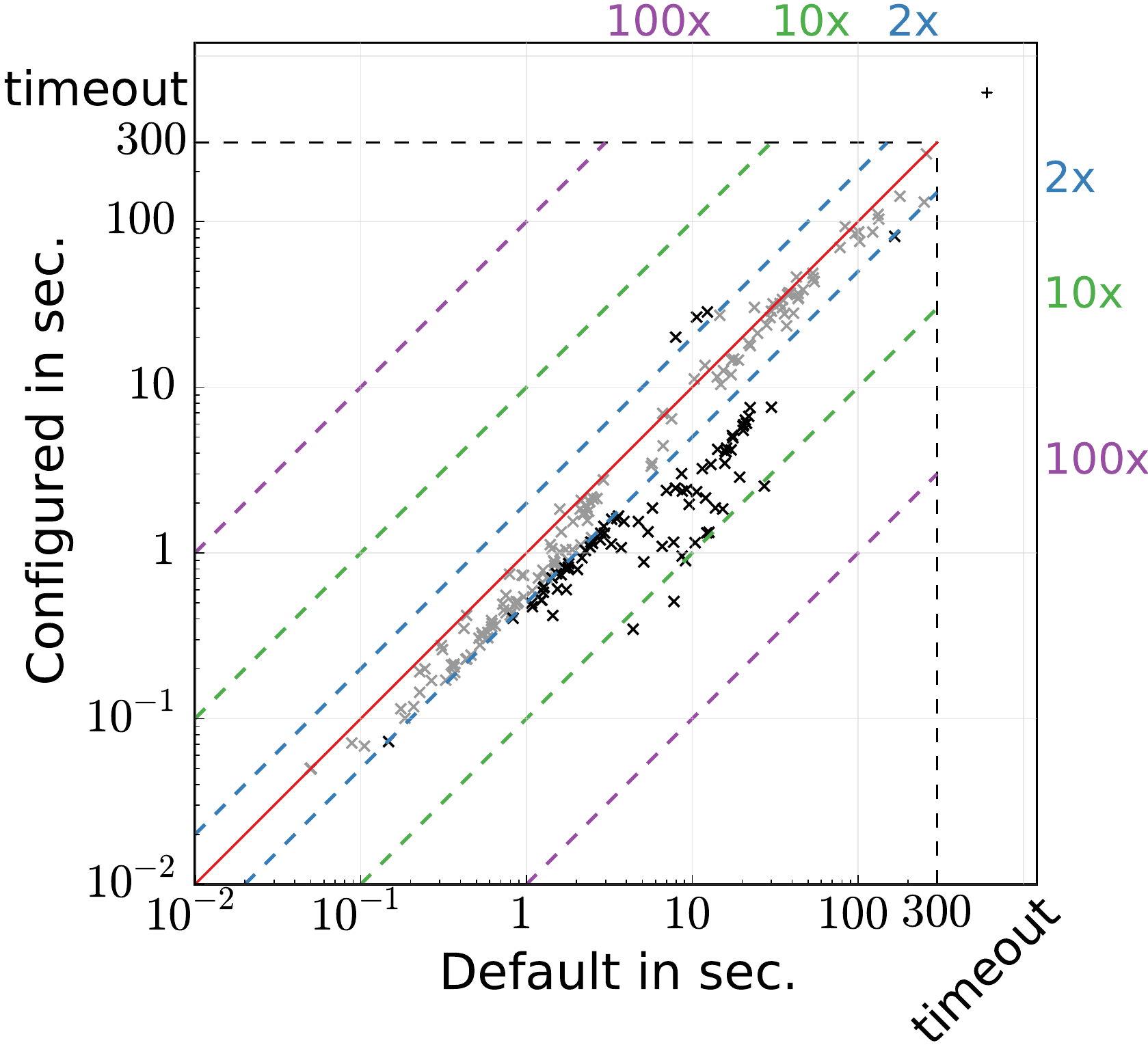}}
		\captionsetup{justification=justified}
		\caption{Scatter plots of default vs. configured \lingeling{}, the gold-medal winner of the \indu{} track of CSSC 2014.\label{fig:cssc14-lingeling}}
							\end{figure}
\vspace*{-0.5cm}
		\begin{table}[H]
		\setlength\tabcolsep{0.3em}
		\sffamily\footnotesize\centering
		\begin{tabular}{l | ccc | c | cc}
		\toprule[1.0pt]
		 & \multicolumn{4}{c|}{$\#$timeouts default $\to$ $\#$ timeouts configured (on test set)} & \multicolumn{2}{c}{Rank} \\
		 & \bmcshort & \circuit & \ibm & \textit{Overall} & def & {\scriptsize{CSSC}}\\
		\midrule
		\lingeling & $20 \to 20$ & $30 \to \mathbf{18}$ & $69 \to 69$ & $119 \to \mathbf{107}$ & 2 & 1\\
		\myrowcolour{}\minisathacksmall & $22 \to 22$ & $21 \to \mathbf{19}$ & $70 \to 70$ & $113 \to \mathbf{111}$ & 1 & 2\\
		\clasp & $44 \to \mathbf{30}$ & $18 \to \mathbf{12}$ & $71 \to 71$ & $133 \to \mathbf{113}$ & 4 & 3\\
		\myrowcolour{}\riss & $39 \to \mathbf{26}$ & $\mathbf{20} \to 22$ & $72 \to 72$ & $131 \to \mathbf{120}$ & 3 & 4\\
		\cryptominisat & $40 \to \mathbf{37}$ & $31 \to \mathbf{20}$ & $70 \to \mathbf{69}$ & $141 \to \mathbf{126}$ & 5 & 5\\
		\myrowcolour{}\sparrow & $62 \to \mathbf{36}$ & $29 \to \mathbf{21}$ & $72 \to 72$ & $163 \to \mathbf{129}$ & 6 & 6\\
		\bottomrule[1.0pt]
		\end{tabular}
		\caption{Results for CSSC 2014 competition track \indu. For each solver and benchmark, we show the number of test set timeouts achieved with the default and the configured parameter setting, bold-facing the better one. We aggregated results across all benchmarks to compute the final ranking.\label{tab:indu2014}}
		\end{table}
	\end{minipage}
}
\end{figure}

The \indu{} track consisted of three industrial benchmarks detailed in \ref{sec:benchmarks-indu}: \bmcshort~\cite{biere-misc07a}, \circuit~\cite{brummayer-sat10a}, and \ibm~\cite{zarpas-sat05a}. 
%
Figure \ref{fig:cssc14-lingeling} visualizes the results of applying algorithm configuration to the winning solver \lingeling{} on these three benchmark sets.
It shows similar results as in the \indu{} track of CSSC 2013: \lingeling{}'s strong default
performance on `typical' hardware verification benchmarks (IBM and BMC) could
only be improved slightly by configuration, but much larger improvements were possible on less standard benchmarks, such as \circuit{}.

Table \ref{tab:indu2014} summarizes the results for all six solvers that participated in the \indu{} track.
These results demonstrate that, in contrast to \lingeling{}, several solvers (in particular, \clasp{}, \riss{}, and \sparrow{})
benefited largely from configuration on the BMC benchmark, but did not reach \lingeling{}'s performance even after configuration. 
\minisathack{} performed even better than \lingeling{} with its default parameters, but did not benefit from configuration as much as \lingeling{}
(particularly on the \circuit{} benchmark family).

\subsubsection{Results of the \crafted{} Track} \label{sec:cssc2014-results-crafted}

\begin{figure}[tb]
\fbox{
	\noindent\begin{minipage}[t]{0.96\linewidth}
		\begin{center}\bf{}\underline{Results for CSSC 2014 \crafted{} track}\end{center}\vspace*{-0.7cm}
		\begin{figure}[H]
		\centering
		\captionsetup{justification=centering,captionskip=0cm}
		\subfloat[\gishort{}, PAR-10: $370 \to 90$][\gishort{}\\ PAR-10: $370 \to
		90$]{\includegraphics[width=0.333\textwidth]{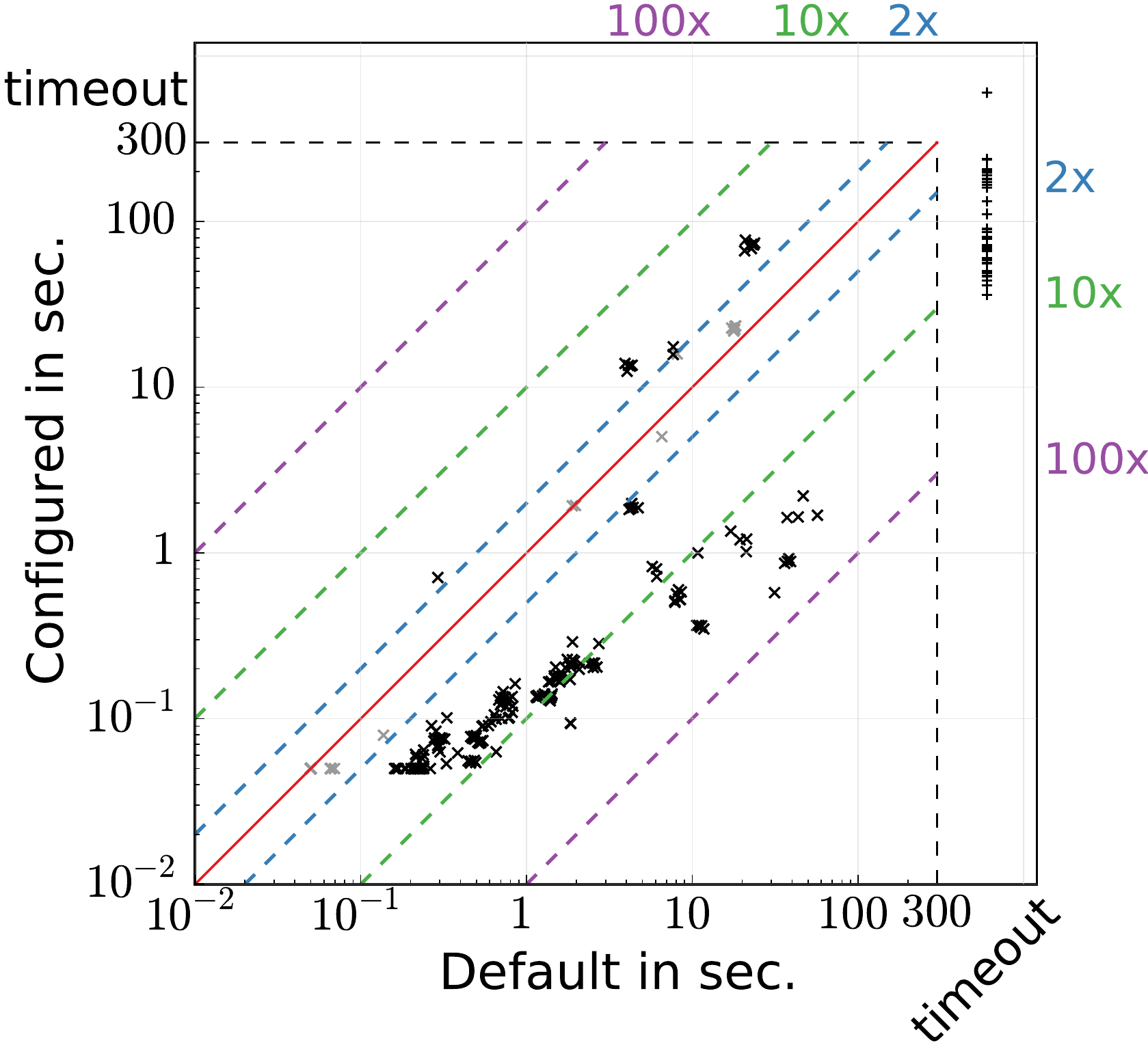}} \subfloat[\labsshort{}, PAR-10: $755 \to 804$][\labsshort{}\\ PAR-10: $755 \to
		804$\label{fig:cssc14-clasp-labs}]{\includegraphics[width=0.333\textwidth]{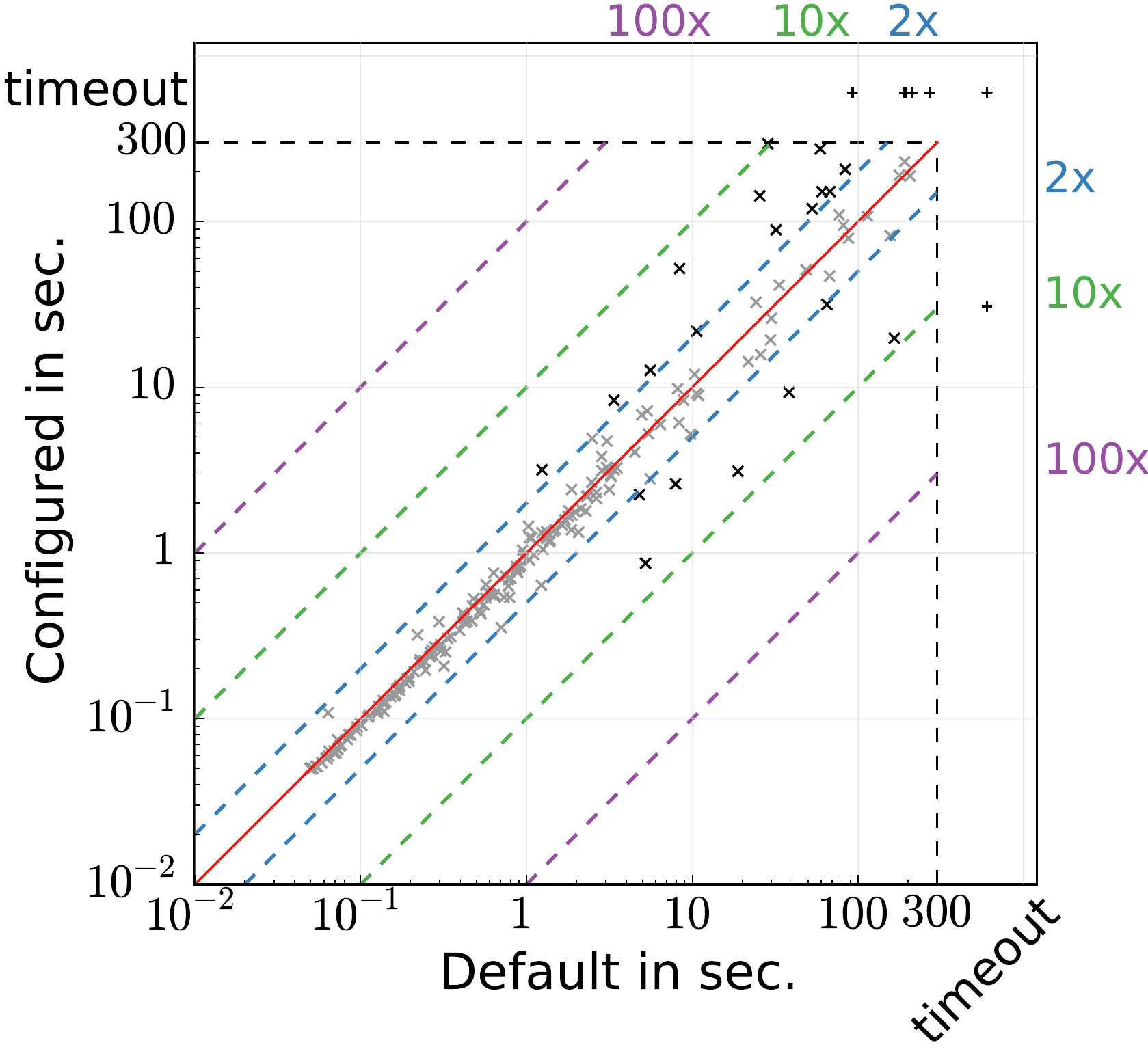}} \subfloat[\queens{}, PAR-10: $705 \to 4.68$][\queens{}\\ PAR-10: $705
		\to
		5$]{\includegraphics[width=0.333\textwidth]{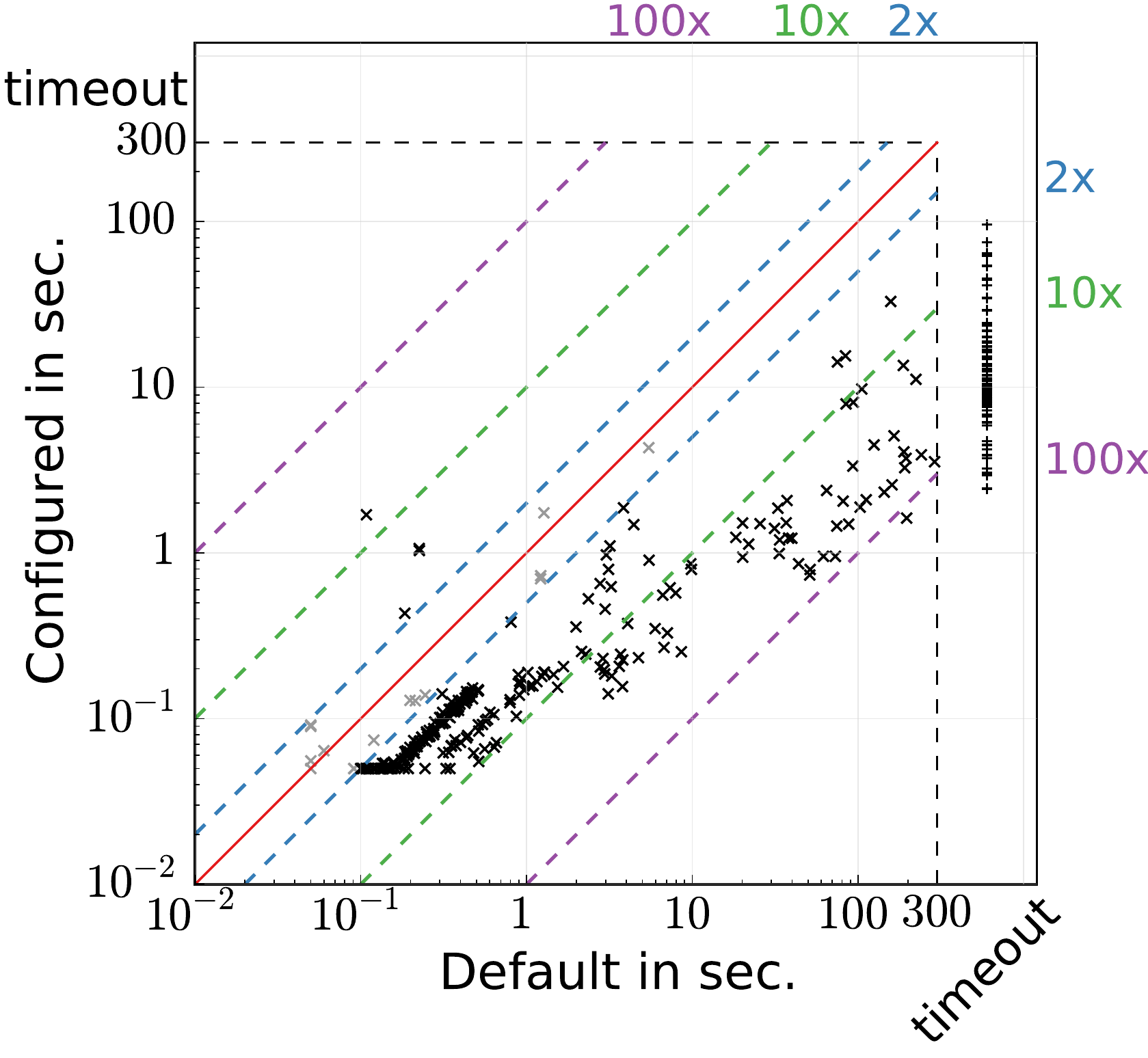}}
		\captionsetup{justification=justified}
		\caption{Scatter plots of default vs.\ configured \clasp{}, the gold medal winner of the \crafted{} track of CSSC 2014.\label{fig:cssc14-clasp-crafted}}
		\end{figure}
\vspace*{-0.5cm}
		\begin{table}[H]
		\setlength\tabcolsep{0.3em}
		\sffamily\footnotesize\centering
		\begin{tabular}{l | ccc | c | cc}
		\toprule[1.0pt]
		 & \multicolumn{4}{c|}{$\#$timeouts default $\to$ $\#$ timeouts configured (on test set)} & \multicolumn{2}{c}{Rank} \\
		 & \gishort & \labsshort & \queens & \textit{Overall} & def & {\scriptsize{CSSC}}\\
		\midrule
		\clasp & $43 \to \mathbf{9}$ & $\mathbf{87} \to 93$ & $81 \to \mathbf{0}$ & $211 \to \mathbf{102}$ & 5 & 1\\
		\myrowcolour{}\lingeling & $11 \to \mathbf{5}$ & $\mathbf{101} \to 104$ & $3 \to \mathbf{0}$ & $115 \to \mathbf{109}$ & 1 & 2\\
		\cryptominisat & $43 \to \mathbf{24}$ & $95 \to \mathbf{89}$ & $2 \to \mathbf{1}$ & $140 \to \mathbf{114}$ & 3 & 3\\
		\myrowcolour{}\riss & $43 \to \mathbf{30}$ & $91 \to \mathbf{88}$ & $2 \to \mathbf{0}$ & $136 \to \mathbf{118}$ & 2 & 4\\
		\minisathacksmall & $50 \to 50$ & $91 \to 91$ & $0 \to 0$ & $141 \to 141$ & 4 & 5\\
		\myrowcolour{}\yalsat & $186 \to 186$ & $218 \to \mathbf{207}$ & $351 \to 351$ & $755 \to \mathbf{744}$ & 6 & 6\\
		\midrule
		\sparrow (disq.) & $55 \to \mathbf{42}$ & $98 \to \mathbf{94}$ & $3 \to \mathbf{0}$ & $156 \to \mathbf{136}$ & - & -\\
		\bottomrule[1.0pt]
		\end{tabular}
		\caption{Results for CSSC 2014 competition track \crafted. For each solver and benchmark, we show the number of test set timeouts achieved with the default and the configured parameter settings, bold-facing the better one. We aggregated results across all benchmarks to compute the final ranking. \sparrow{} was disqualified from this track, since it returned `satisfiable' for one instance without producing a model.
		\label{tab:crafted2014}}
		\end{table}
	\end{minipage}
}	
\end{figure}

The \crafted{} track consisted of the three crafted benchmarks detailed in \ref{sec:benchmarks-crafted}: \gi{}, \labs{}, and \queens{}.
Figure \ref{fig:cssc14-clasp-crafted} visualizes the improvements configuration yielded on these benchmarks for the best-performing solver, \clasp{}.
The effect of configuration was particularly large on the \queens{} instances, where it reduced the number of timeouts from 81 to 0.
Similar to the results from CSSC 2013, configuration also substantially improved performance on the \gishort{} instances, decreasing the number of timeouts from 43 to 9.
In contrast to 2013, an unusual effect occurred for \clasp{} on the \labsshort{} instances, where the number of timeouts on the test set \emph{increased} from 87 to 93 by configuration; we study the reasons for this in more detail in Section \ref{sec:why-does-it-work}.

Table \ref{tab:crafted2014} summarizes the results of all solvers on the \crafted{} track, showing that the performance of many other solvers also substantially improved on the benchmarks \gishort{} and \queens{}, and only mildly (if at all) on the \labsshort{} benchmark. 
The aggregate results across these 3 benchmark families show that \lingeling{} had the best default performance, but only benefited mildly from configuration (\#timeouts reduced from 115 to 109), whereas \clasp{} benefited much more from configuration and thus outperformed \lingeling{} after configuration (\#timeouts reduced from 211 to 102). 
Once again, we note that the winning solver only showed mediocre performance based on its default: \clasp{} would have ranked 5th in a comparison based on default performance.

\subsubsection{Results of the \random{} Track} \label{sec:cssc2014-results-random}

\begin{figure}[t]
\fbox{
	\noindent\begin{minipage}[t]{0.96\linewidth}
		\begin{center}\bf{}\underline{Results for CSSC 2014 \random{} track}\end{center}\vspace*{-0.7cm}
		\begin{figure}[H]
		\centering
		\captionsetup{justification=centering,captionskip=0cm}
		\subfloat[\threecnf{}, PAR-10: $309 \to 35$][\threecnf{}\\ PAR-10: $309 \to
		35$]{\includegraphics[width=0.333\textwidth]{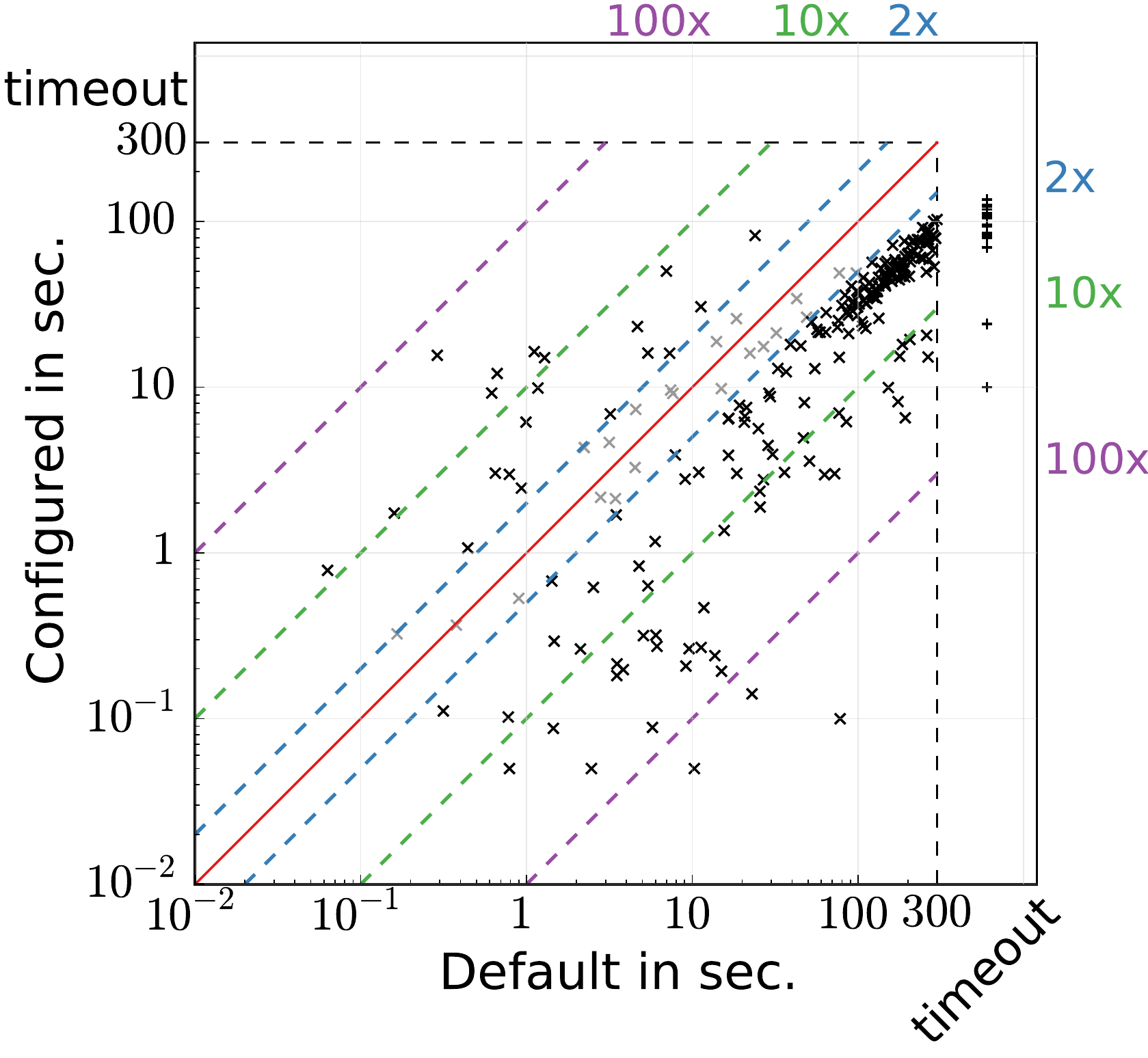}} \subfloat[\kthree{}, PAR-10: $8 \to 3$][\kthree{}\\ PAR-10: $7.91 \to
		2.66$]{\includegraphics[width=0.333\textwidth]{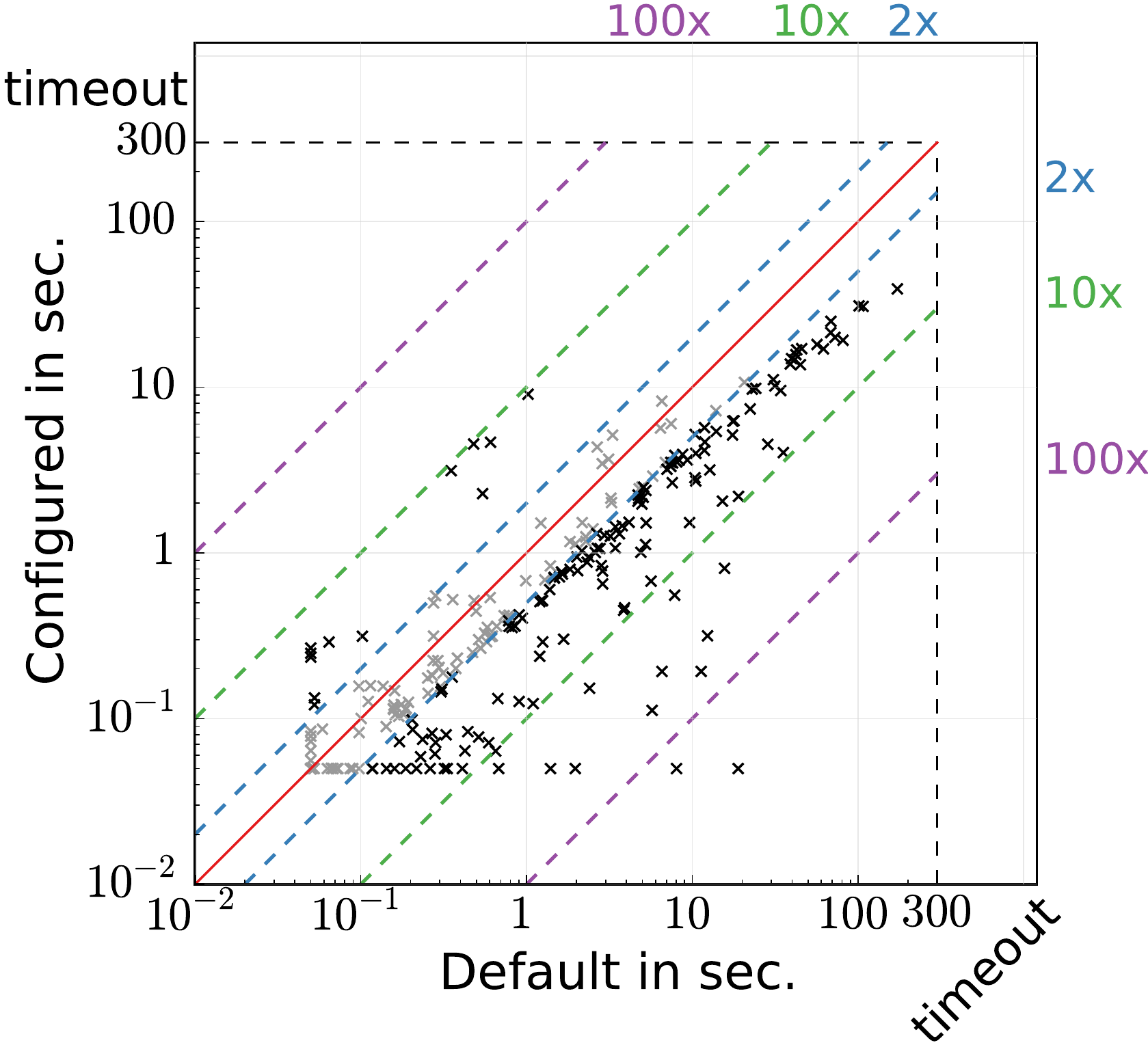}}
		\subfloat[\unif{}, PAR-10: $1 \to < 0.5$][\unif{}\\ PAR-10: $0.74 \to
		0.30$]{\includegraphics[width=0.333\textwidth]{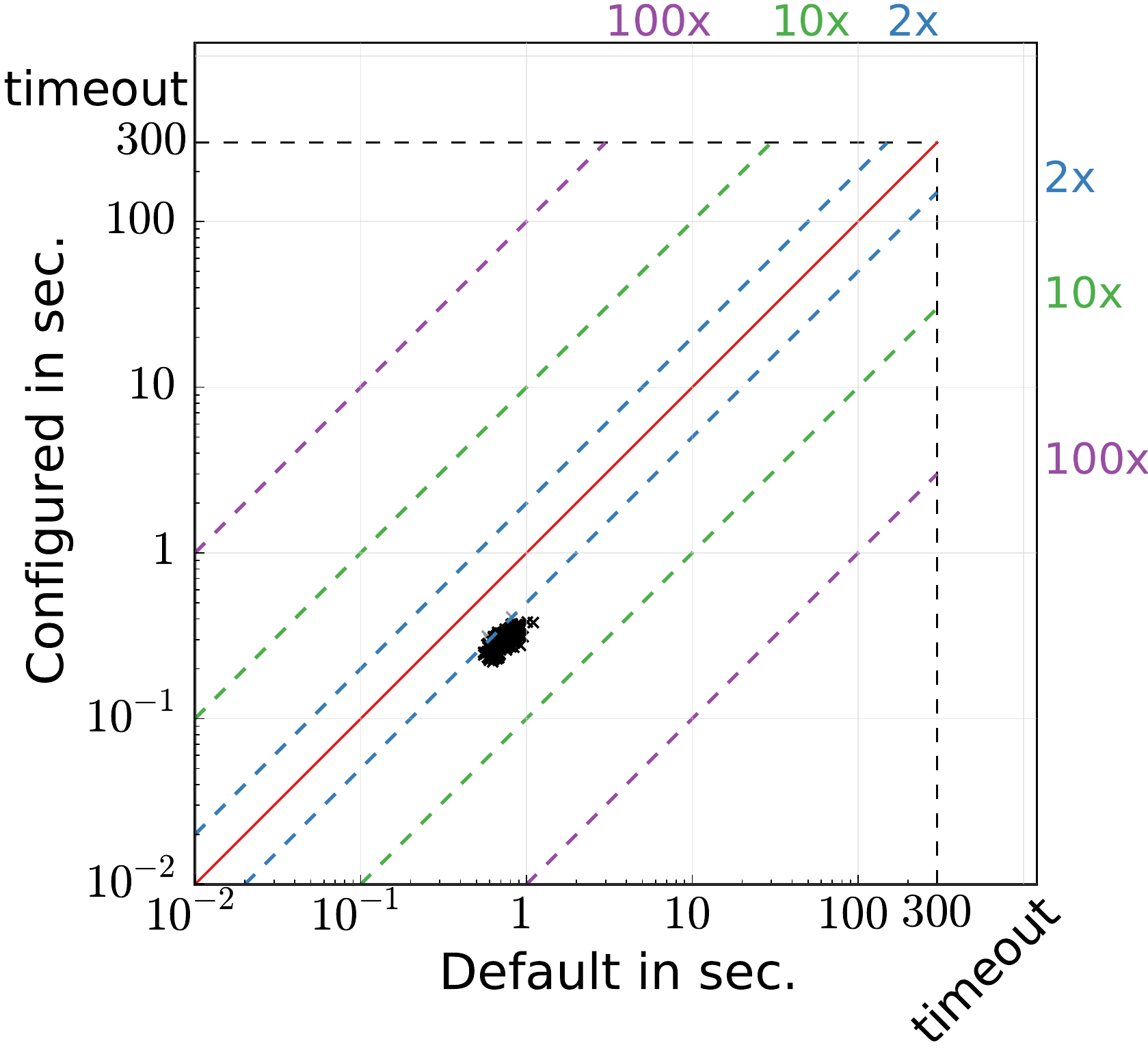}}
		\captionsetup{justification=justified}
		\caption{Scatter plots of default vs.\ configured \clasp{}, the gold medal winner of the \random{} track of CSSC 2014.\label{fig:clasp-random-2014}}
		\end{figure}
\vspace*{-0.5cm}
		\begin{table}[H]
		\setlength\tabcolsep{0.3em}
		\sffamily\footnotesize\centering
		\begin{tabular}{l | ccc | c | cc}
		\toprule[1.0pt]
		 & \multicolumn{4}{c|}{$\#$timeouts default $\to$ $\#$ timeouts configured (on test set)} & \multicolumn{2}{c}{Rank} \\
		 & \threecnf & \kthree & \unif & \textit{Overall} & def & {\scriptsize{CSSC}}\\
		\midrule
		\clasp & $18 \to \mathbf{0}$ & $0 \to 0$ & $0 \to 0$ & $18 \to \mathbf{0}$ & 2 & 1\\
		\myrowcolour{}\dccsat & $1 \to \mathbf{0}$ & $0 \to 0$ & $1 \to \mathbf{0}$ & $2 \to \mathbf{0}$ & 1 & 2\\
		\minisathacksmall & $166 \to \mathbf{99}$ & $5 \to \mathbf{1}$ & $0 \to 0$ & $171 \to \mathbf{100}$ & 5 & 3\\
		\myrowcolour{}\riss & $160 \to \mathbf{113}$ & $2 \to 2$ & $1 \to \mathbf{0}$ & $163 \to \mathbf{115}$ & 4 & 4\\
		\sparrow & $126 \to 126$ & $8 \to \mathbf{1}$ & $0 \to 0$ & $134 \to \mathbf{127}$ & 3 & 5\\
		\bottomrule[1.0pt]
		\end{tabular}
		\caption{Results for CSSC 2014 competition track \random. For each solver and
		benchmark, we show the number of test set timeouts achieved with the default
		and the configured parameter settings, bold-facing the better one; we broke
		ties by the solver's average runtime (not shown for brevity, but the average
		runtimes important for tie breaking were 13 seconds for configured \clasp{}
		and 21 seconds for configured \dccsat{}). We aggregated results across all benchmarks to compute the final ranking.\label{tab:random2014}}
		\end{table}
	\end{minipage}
}
\end{figure}

The \random{} track consisted of three random benchmarks detailed in \ref{sec:benchmarks-random}: \threecnf{}, \kthree{}, and \unif{}.
The instances in \unif{} are all unsatisfiable, while the other two sets contain both satisfiable and unsatisfiable instances.
Figure~\ref{fig:clasp-random-2014} visualizes the improvements achieved by configuration on these benchmarks for the best-performing solver \clasp{}.
\clasp{} benefited most from configuration on benchmark \threecnf{}, where it reduced the number of timeouts from 18 to 0. 
For the other benchmarks, it could already solve all instances in its default
parameter configuration, but configuration helped reduce its average runtime by factors of 3~(\kthree) and 2~(\unif), respectively.
Table~\ref{tab:random2014} summarizes the results of all solvers for these benchmarks. 
We note that solver \dccsat{} showed the best default performance, and that after configuration, it also solved all instances from the three benchmark sets, only ranking behind \clasp{} because the latter solved these instances faster.

\subsubsection{Results of the \randomsat{} Track} \label{sec:cssc2014-results-randomsat}

\begin{figure}[t]
\fbox{
	\noindent\begin{minipage}[t!]{0.96\linewidth}
		\begin{center}\bf{}\underline{Results for CSSC 2014 \randomsat{} track}\end{center}\vspace*{-0.7cm}
		\begin{figure}[H]
		\centering
		\captionsetup{justification=centering,captionskip=0cm}
		\subfloat[\threesatonek{}, PAR-10: $132 \to 53$][\threesatonek{}\\ PAR-10:
		$132 \to 53$]{\includegraphics[width=0.333\textwidth]{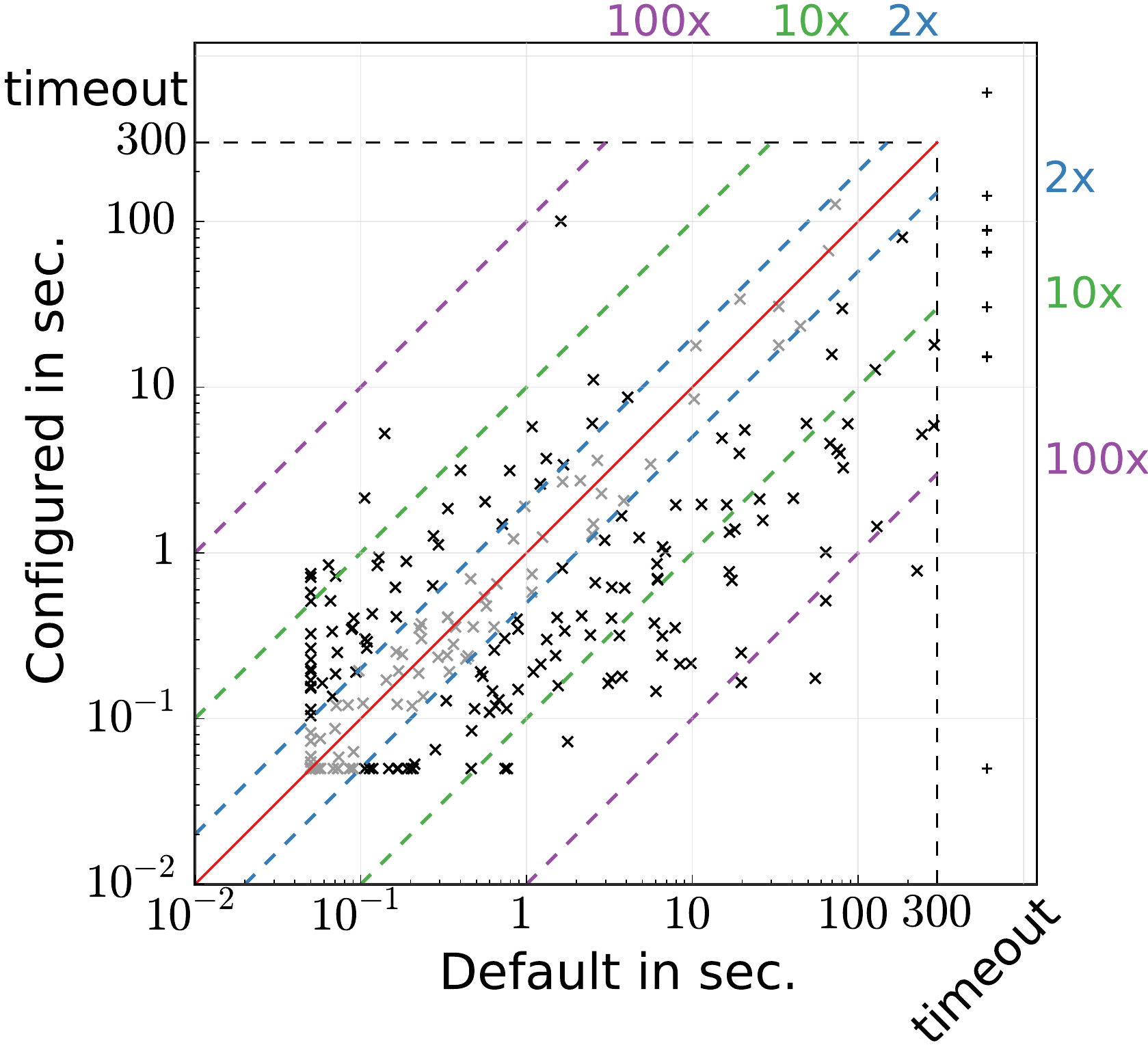}}
		\subfloat[\fivesatfiveh{}, PAR-10: $3000 \to 2$][\fivesatfiveh{}\\ PAR-10:
		$3000 \to 2$]{\includegraphics[width=0.333\textwidth]{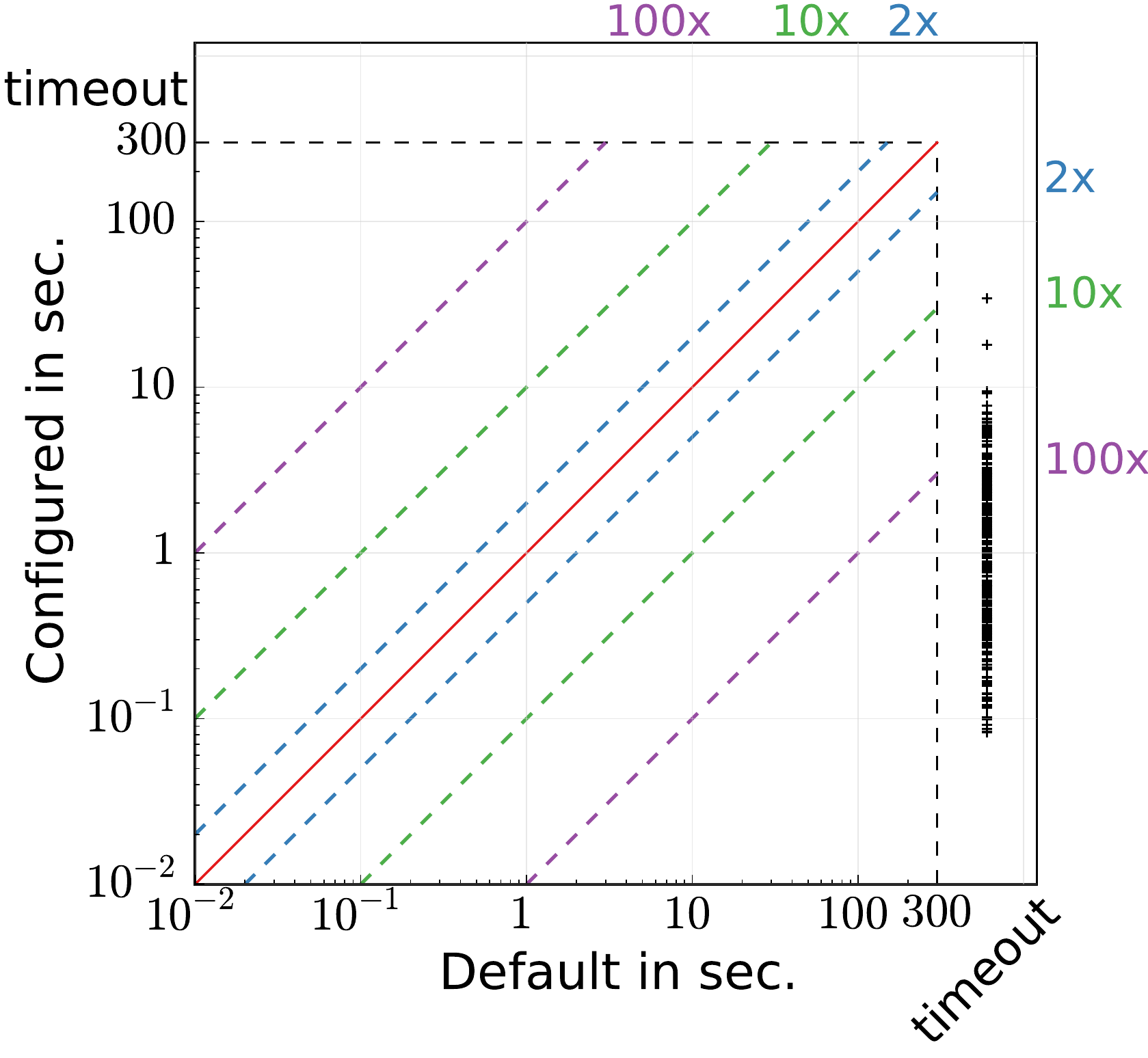}}
		\subfloat[\sevensatninety{}, PAR-10: $337 \to 15$][\sevensatninety{}\\ PAR-10:
		$337 \to 15$]{\includegraphics[width=0.333\textwidth]{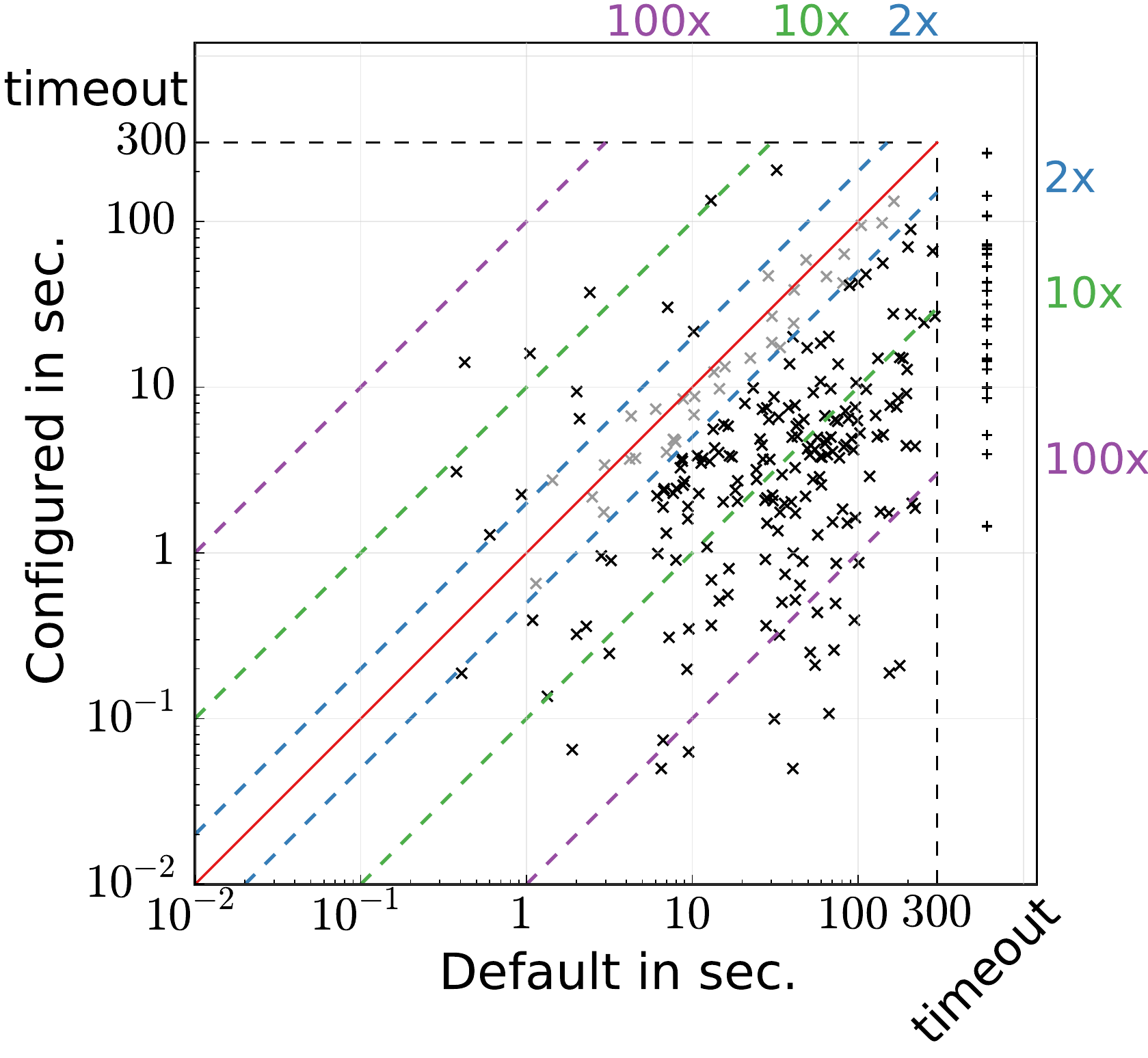}}
		\captionsetup{justification=justified}
		\caption{Scatter plots of default vs.\ configured \probsat{}, the gold medal winner of the \randomsat{} track of CSSC 2014.\label{fig:probsat-randomsat-2014}}
		\end{figure}
\vspace*{-0.5cm}
		\begin{table}[H]
		\setlength\tabcolsep{0.3em}
		\sffamily\footnotesize\centering
		\begin{tabular}{l | ccc | c | cc}
		\toprule[1.0pt]
		 & \multicolumn{4}{c|}{$\#$timeouts default $\to$ $\#$ timeouts configured (on test set)} & \multicolumn{2}{c}{Rank} \\
		 & \threesatonek & \fivesatfiveh & \sevensatninety & \textit{Overall} & def & {\scriptsize{CSSC}}\\
		\midrule
		\probsat & $10 \to \mathbf{4}$ & $250 \to \mathbf{0}$ & $24 \to \mathbf{0}$ & $284 \to \mathbf{4}$ & 4 & 1\\
		\myrowcolour{}\sparrow & $9 \to \mathbf{5}$ & $250 \to \mathbf{0}$ & $3 \to 3$ & $262 \to \mathbf{8}$ & 3 & 2\\
		\csccsat & $2 \to 2$ & $0 \to 0$ & $\mathbf{3} \to 6$ & $\mathbf{5} \to 8$ & 1 & 3\\
		\myrowcolour{}\yalsat & $\mathbf{6} \to 7$ & $0 \to 0$ & $5 \to 5$ & $\mathbf{11} \to 12$ & 2 & 4\\
		\clasp & $250 \to 250$ & $250 \to 250$ & $250 \to 250$ & $750 \to 750$ & 5 & 5\\
		\myrowcolour{}\minisathacksmall & $250 \to 250$ & $250 \to 250$ & $250 \to 250$ & $750 \to 750$ & 6 & 6\\
		\bottomrule[1.0pt]
		\end{tabular}
		
		\caption{Results for CSSC 2014 competition track \randomsat. For each solver and benchmark, we show the number of test set timeouts achieved with the default and the configured parameter settings, bold-facing the better one; we broke ties by the solver's average runtime (not shown for brevity). We aggregated results across all benchmarks to compute the final ranking.\label{tab:randomsat2014}}
		\end{table}
	\end{minipage}
}
\end{figure}

The \randomsat{} track consisted of the three benchmarks detailed in \ref{sec:benchmarks-random}: 
\emph{3sat1k}, \emph{5sat500} and \emph{7sat90}.
Figure \ref{fig:probsat-randomsat-2014} visualizes the improvements configuration achieved on these benchmarks for the best-performing solver \probsat{}.
\probsat{} benefited most from configuration on benchmark \fivesatfiveh{}: its
default did not solve a single instance in the maximum runtime of 300 seconds,
while its configured version solved all instances in an average runtime below 2
seconds! Since timeouts at 300s yield a PAR-10 score of 3000, the PAR-10
speedup factor on this benchmark was $1\,500$, the largest we observed in
the CSSC. On the other two scenarios, configuration was also very beneficial,
reducing \probsat{}'s number of timeouts from 24 to 0 (7sat90) and from 10 to 4 (3sat1k), respectively.
Table \ref{tab:randomsat2014} summarizes the results of all solvers for these
benchmarks, showing that next to \probsat{}, only \sparrow{} benefited from
configuration. Neither of the CDCL solvers (\clasp{} and \minisathack{}) solved
a single instance in any of the three benchmarks (in either default or
configured variants). For the other two SLS solvers, \yalsat{} and \csccsat{}, the defaults were already well tuned for these benchmark sets. Indeed, we observed overtuning to the training sets in one case each: \yalsat{} for \threesatonek{} and \csccsat{} for \sevensatninety.
Overall, the configurability of \probsat{} and \sparrow{} allowed them to place first and second, respectively, despite their poor default performance (especially on \fivesatfiveh, where neither of them solved a single instance with default settings).

\section{Post-Competition Analyses} \label{sec:discussion-of-results}

While the previous sections focussed on the results of the respective
competitions, we now discuss several analyses we performed afterwards
to study overarching phenomena and general patterns.

\changed{
\subsection{Why Does Configuration Work So Well and How Can It Fail?} \label{sec:why-does-it-work}

Several practitioners have asked us why automated configuration can yield the large speedups over the default configuration we observed.
We believe there are two key reasons for this:
\begin{itemize}
	\item No single algorithmic approach performs best on all types of benchmark instances; this is precisely the same reason that algorithm selection approaches (such as SATzilla~\cite{xu-jair08a} or 3S~\cite{kadioglu-cp11a}) work so well. 
	
	\item Solver defaults are typically chosen to be robust across benchmark families. For any given benchmark family $F$, highly parameterized solvers can, however, typically be instantiated to exploit the idiosyncrasies of $F$ substantially better. (These improvements only need to generalize to other instances from $F$, not to other benchmark families.)
\end{itemize}

However, algorithm configuration does not necessarily work in all cases.
For example, in the \crafted{} track of the CSSC 2014, we encountered a case in which the configured solver performed somewhat \emph{worse} than the default solver: \claspshort{} configured on benchmark family \labsshort{} timed out on 93 test instances, whereas its default only timed out on 87 test instances (see also Figure \ref{fig:cssc14-clasp-labs}).
Two obvious causes suggest themselves in the case of such a failure:
\begin{itemize}
\denselist
	\item insufficiently long configuration runs (which can result in worse-than-default
performance on the \emph{training} set\footnote{Worse-than-default performance on the training set is possible since configurators only base their decisions on a \emph{subset} of the training instances; that subset increases over time and only reaches the full training set when the configuration process is given enough time.}); and/or 
	\item overtuning on the training set that does not generalize to the \emph{test} set.
\end{itemize}
We investigated the configuration of \claspshort{} on \labsshort{} further after the competition, and found that in this case \emph{both} of these effects applied:
In the CSSC, training performance slightly deteriorated ($86\rightarrow{} 87$ timeouts);
and the improved training performance we found with a larger configuration 
budget\footnote{Specifically, we ran 32 \smac{} runs for 10 days each.} afterwards
($86\rightarrow{} 83$ timeouts) also did not generalize to the test set ($87\rightarrow{} 88$ timeouts).

\begin{figure}[tb]
\captionsetup{justification=centering}
\subfloat[\queens{}]{\includegraphics[width=0.48\textwidth]{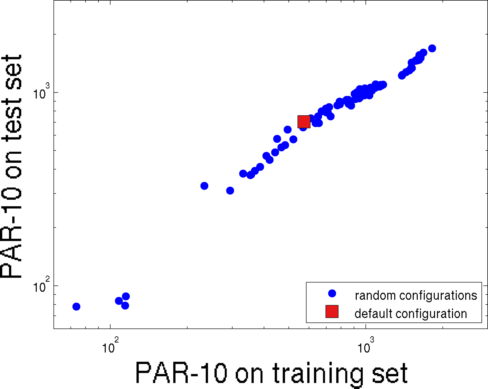}}~
\subfloat[\labsshort{}]{\includegraphics[width=0.48\textwidth]{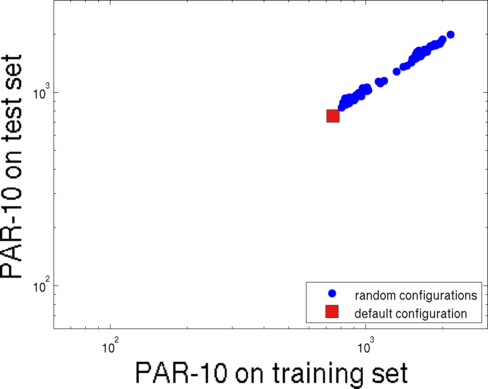}}
\captionsetup{justification=justified}
\caption{Training \vs{} test PAR-10 scores of 100 random \claspshort{} configurations and the \clasp{} default.
\label{fig:score_correlation}}
\end{figure}

To contrast the conditions under which configuration can fail and under which it works well, we compared the 
configuration of \claspshort{} on benchmarks \labsshort{} ($87\rightarrow 93$ test timeouts in the CSSC)
and \queens{} ($81\rightarrow 0$ test timeouts in the CSSC).
For this analysis, we sampled 100 \claspshort{} configurations uniformly at random and evaluated their PAR-10 
training and test scores on both of the benchmarks; Figure \ref{fig:score_correlation} shows the result.
The first observation we make directly based on the figure is that for \queens{}, about 20\% of the random configurations outperform 
the default, whereas for \labsshort{} none of them do: the default is simply very good to start with for \labsshort{} and thus much harder to beat. 
Second, since several configurations are very good (i.e., fast) for \queens{}, configurators can make progress much faster (and also take full advantage of adaptive capping to limit the time spent with poor configurations); indeed, in
the CSSC the configurators managed to perform about $8$ times more \claspshort{} runs in the same time (2 days) 
for \queens{} than for \labsshort{} (averaging about $14\,400$ vs.\ $1\,800$ runs). This explains why 
configurators require more time to improve training performance on \labsshort{}. 
%
%
Third, to assess the potential for overtuning, we studied how training and test performance correlate.
Visually, Figure \ref{fig:score_correlation} shows a strong overall correlation of PAR-10 training and test scores for both of the benchmarks; Spearman correlation coefficients are indeed high: 0.99 (\queens{}) and 0.98 (\labsshort{}). However, for the top 20\% of sampled configurations, the correlation is much stronger for \queens{} (0.98) than for \labsshort{} (0.49).
This explains why 
improvements on the \labsshort{} training set do not necessarily translate to improvements on its test set.
%
%


}

\subsection{Overall Configurability of Solvers} \label{sec:configurability-discussion}

\begin{figure}[tb]
\captionsetup{justification=centering}
\subfloat[CSSC 2013]{\includegraphics[height=17em]{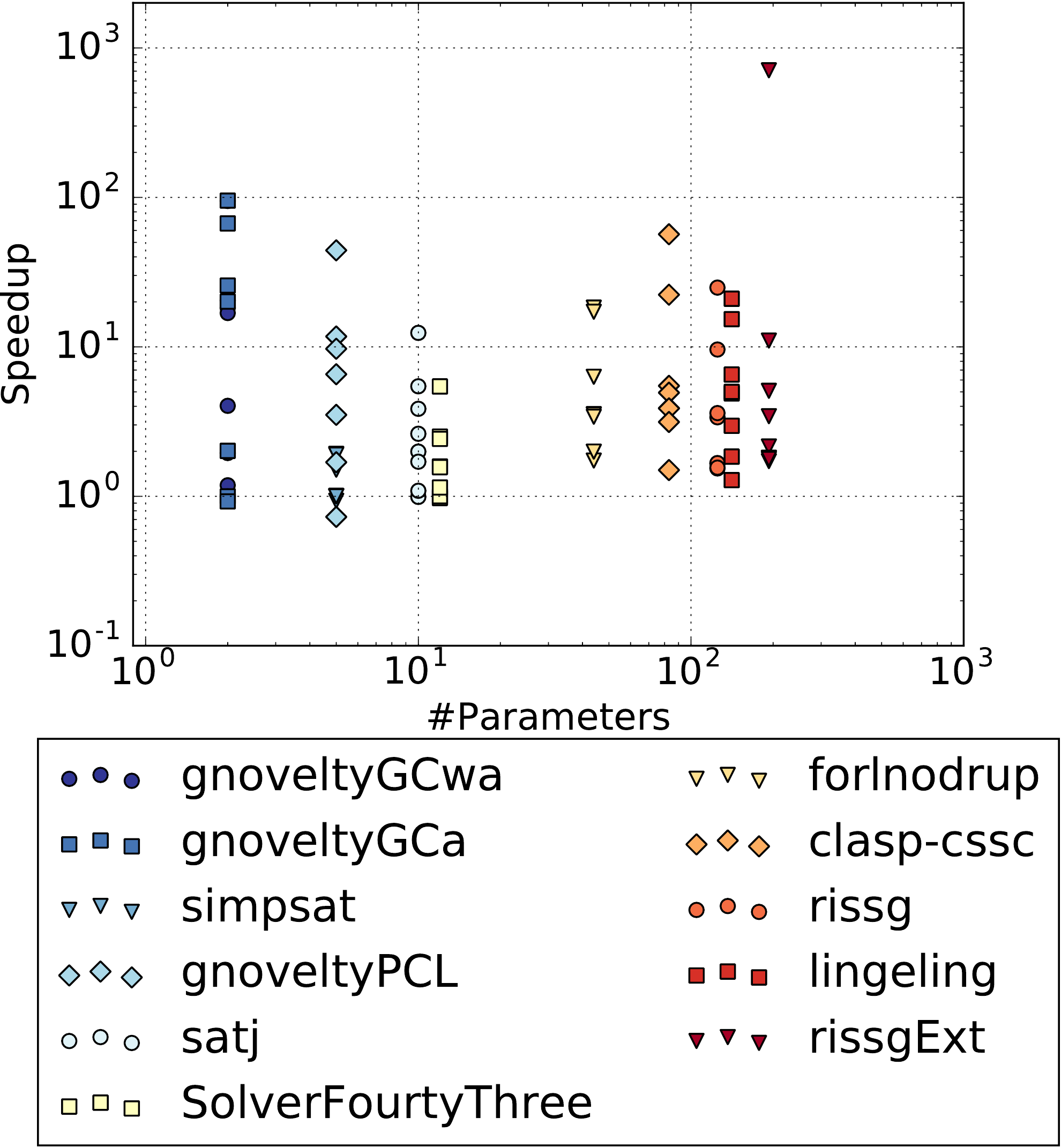}}~
\subfloat[CSSC 2014]{\includegraphics[height=17em]{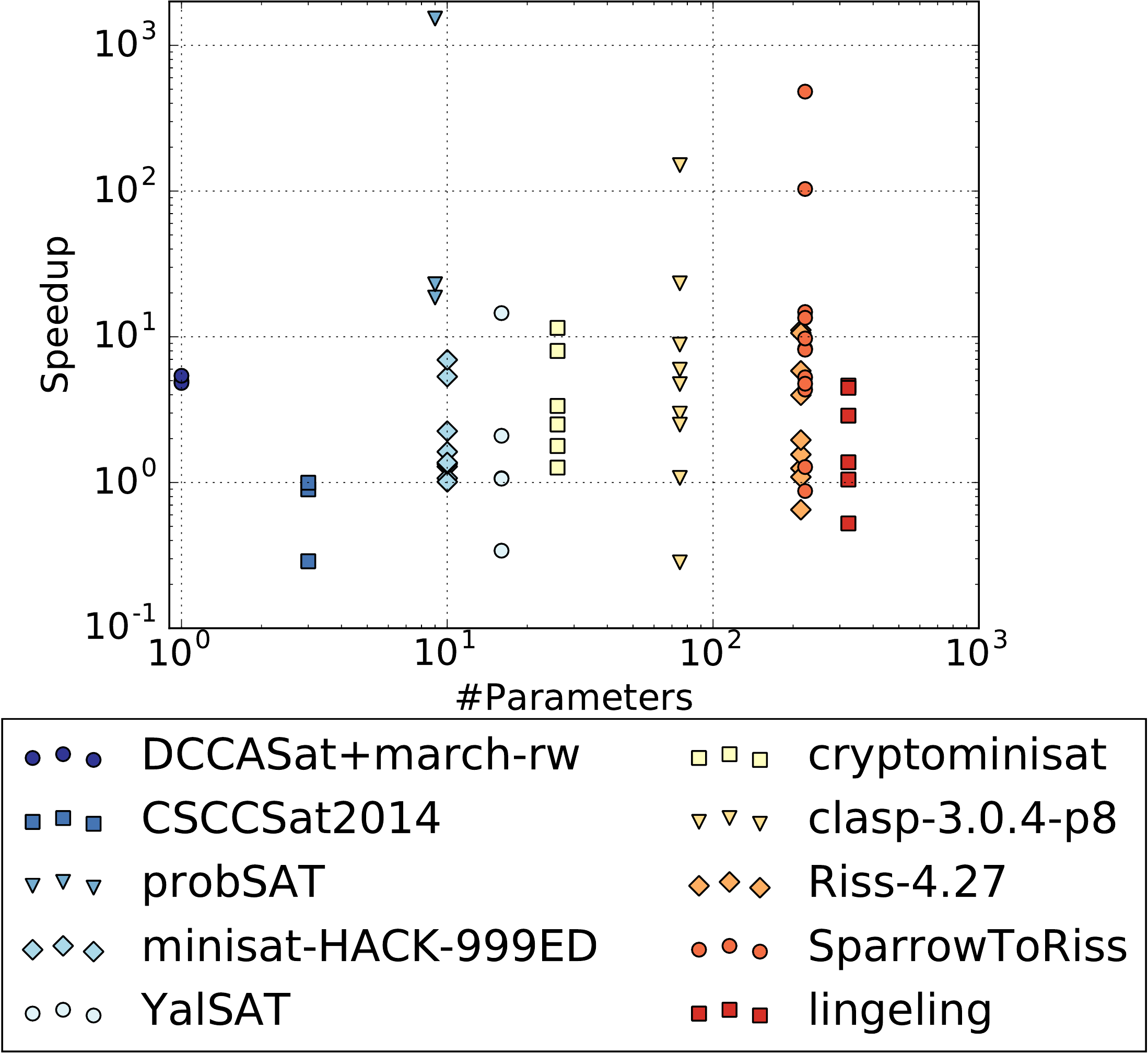}}
\captionsetup{justification=justified}
\caption{Number of solver parameters vs.\ PAR-10 speedup factor of configured
over default solver. The speedup factor for each solver considers only instances that were solved by at least one of the default solver and the configured solver. 
Each symbol denotes one benchmark the solver was run on. 
\changed{Figure \ref{fig:configurability-vs-parameters-par1} in the appendix shows the same figure based on PAR-1 for comparison.}
\label{fig:configurability-vs-parameters}}
\end{figure}

Some solvers consistently benefited more from configuration than others. Here,
we quantify the \emph{configurability} of a solver on a given benchmark by the
PAR-10  speedup factor its configured version achieved over its default version,
computed on the set of instances solved by at least one of the two.
We then examine the relationship between configurability and number of parameters 
to determine whether solvers with many parameters consistently benefited more or less from configuration than solvers with few parameters.\footnote{Of course, it is simple to construct examples where a solver with a single parameter is highly configurable (\eg{}, let the parameter have a poor default setting) or where a solver has many parameters but does not benefit from configuration at all (\eg{}, a solver could expose many parameters that are not actually used at all). The focus of our analysis is therefore on the relationship between configurability and the number of parameters that a solver author reasonably expected would be useful to expose.}

Figure \ref{fig:configurability-vs-parameters} shows that configurability was
indeed high for solvers with many parameters (e.g., the variants of
\lingeling{}, \emph{Riss}, and \claspbasic{}), but that it did not increase monotonically in the number of parameters: some solvers with very few parameters were surprisingly configurable.
For example, configuration sped up the single-parameter solver \dccsat{} by at
least a factor of four in all three benchmarks it was configured for, while
the 4-parameter solver \csccsat{} was not improved at all by configuration.
Furthermore, \probsat{}, which achieved the best single-benchmark performance
improvement (as previously discussed in Section
\ref{sec:cssc2014-results-randomsat}), has only 9 parameters.

We note that the notion of configurability used here is 
strongly dependent on the time budget available for configuration. 
In the next section, we investigate this issue in more detail.

\subsection{Impact of Configuration Budget} \label{sec:impact-configuration-budget}

The runtime budget we allow to configure each solver has an obvious impact on the results.
In one extreme case, if we let this budget go towards zero, the configuration
pipeline returns the solver defaults (and we are back in the setting of the
standard SAT competition). For small, non-zero budgets, we can expect solvers
with few parameters to benefit from configuration more, since their
configuration spaces are easier to search.
On the other hand, if we increase the time budget, solvers with larger
parameter spaces are likely to benefit more than those with smaller parameter
spaces (since larger parts of their configuration space can be searched given
additional time).

\begin{figure}[tb]
		\centering
		\captionsetup{justification=centering}
		\subfloat[\threecnf\label{fig:perf-over-time-different-spaces-threecnf}]{\includegraphics[width=0.333\textwidth]{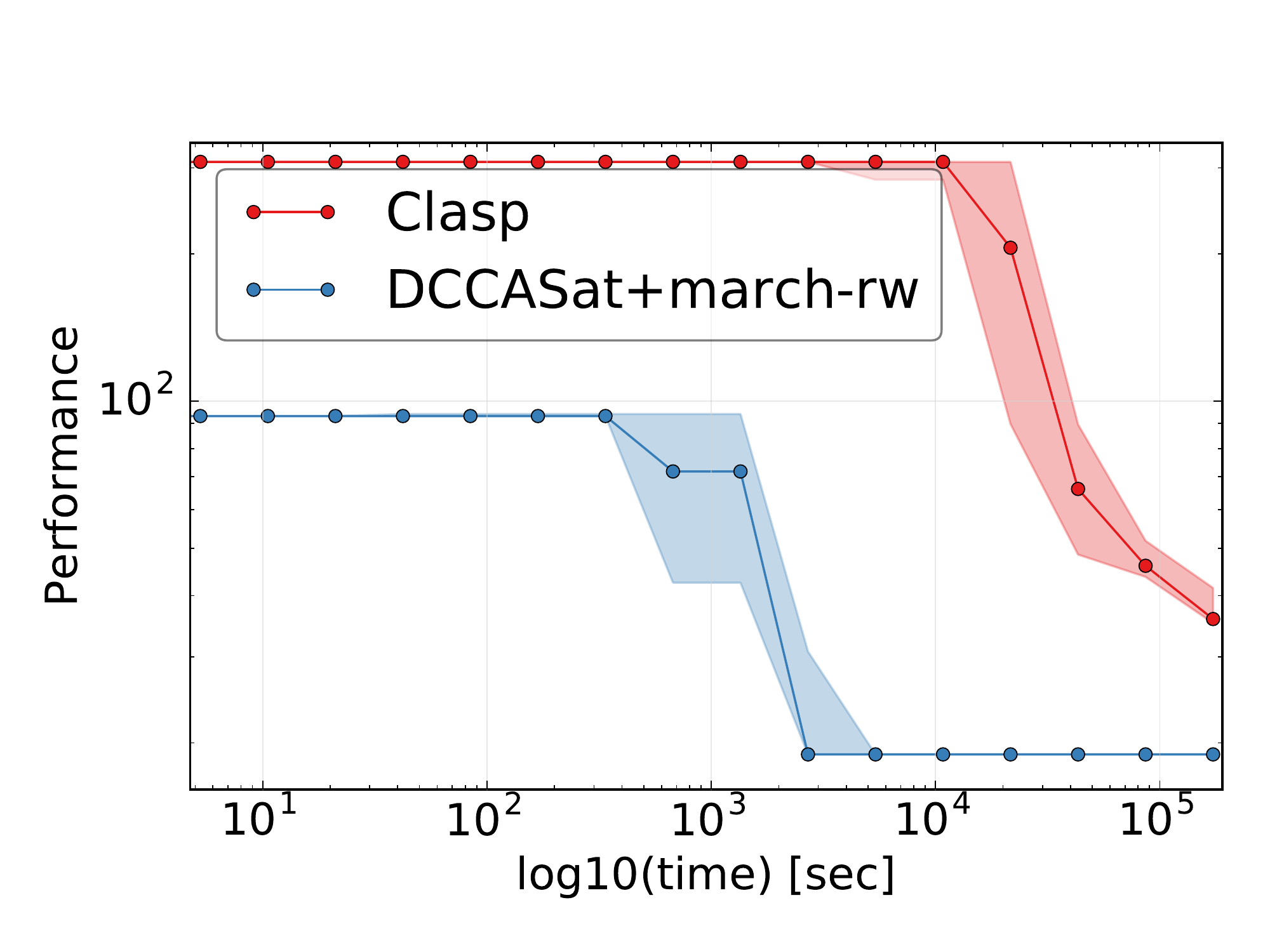}}
		\subfloat[\kthree]{\includegraphics[width=0.333\textwidth]{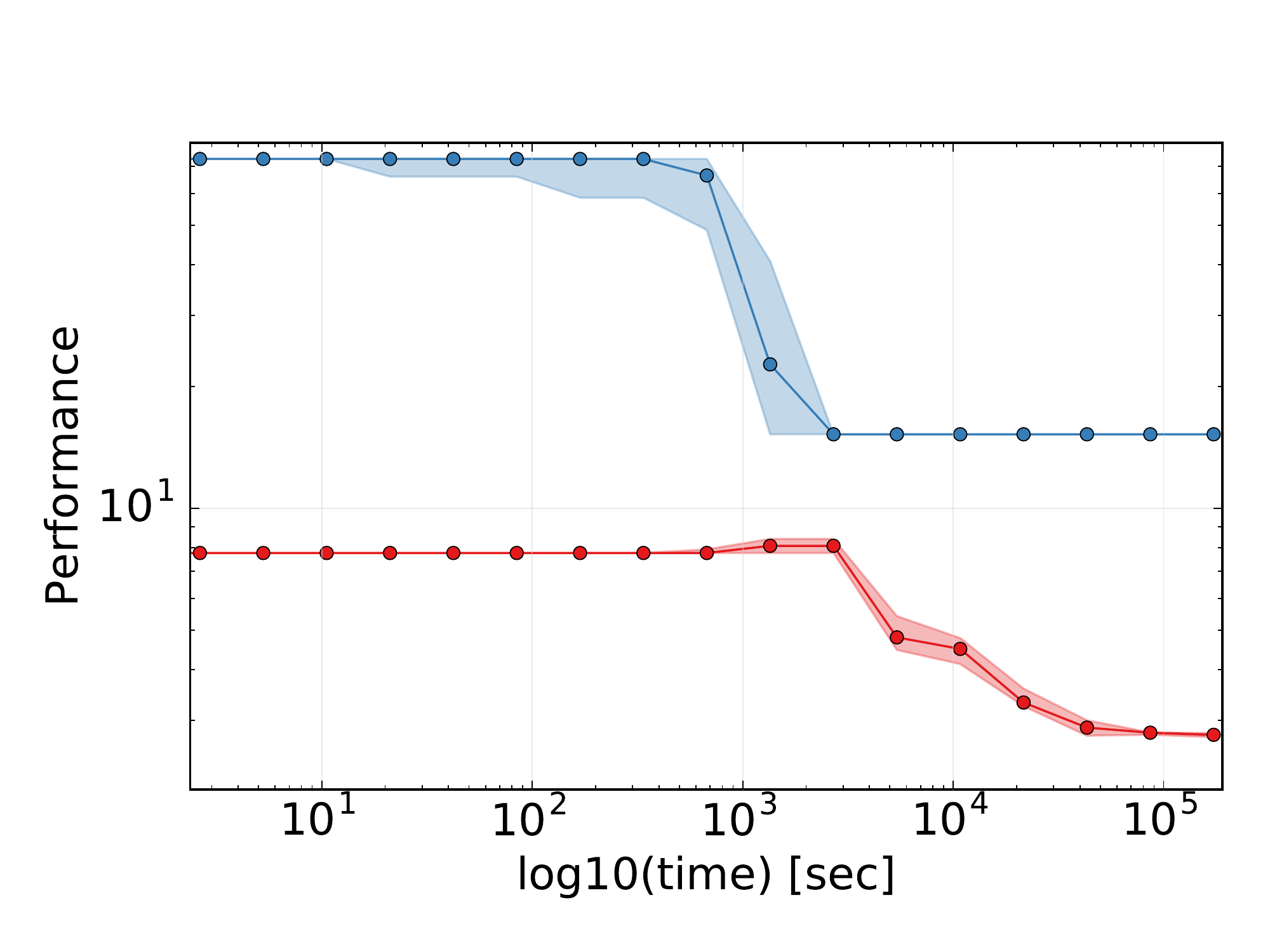}}
		\subfloat[\unif]{\includegraphics[width=0.333\textwidth]{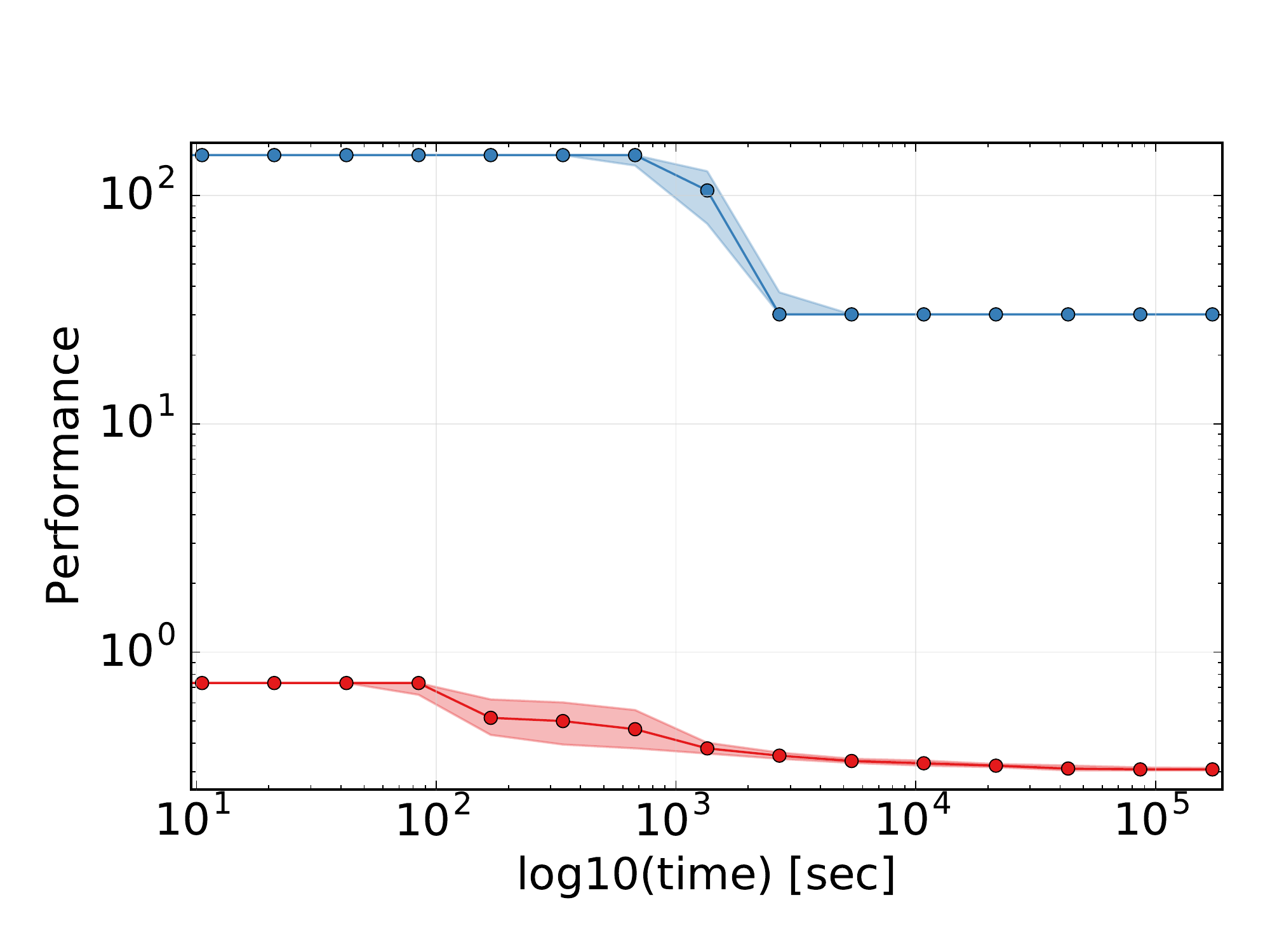}}
		\captionsetup{justification=justified}
		\caption{Performance on the CSSC 2014 track \random{} achieved by
		configuration of \dccsat{} (1 parameter) and \clasp{} (75 parameters) as a
		function of the configuration budget.
		For each of the solvers, we plot mean $\pm$ one standard
		deviation of the PAR$10$ performance of the incumbent configurations found by 4 runs of \smac{} over time. 
		\label{fig:perf-over-time-different-spaces}}
\end{figure}

Figure \ref{fig:perf-over-time-different-spaces} illustrates this phenomenon for the two top solvers in the \random{} track of CSSC 2014.
With the competition's configuration budget of two days across 4 cores, \clasp{} performed better than \dccsat{} (both solved all test instances, with average runtimes of 13 vs.\ 21 seconds).
In the extreme case of no time budget for configuration, \dccsat{} would have won against \clasp{}, since its default version performed much better (2 vs.\ 18 timeouts), and, in fact, Figure \ref{fig:perf-over-time-different-spaces-threecnf} shows that it required a configuration budget of at least $10^4$ seconds to find improving \clasp{} parameters for the \threecnf{} benchmark (where the default version of \clasp{} produced 18 timeouts).
While the configuration of \dccsat{}'s single parameter had long converged by
$10^4$ seconds, the configuration of \clasp{}'s 75 parameters continued to
improve performance until the end of the configuration budget, and, in particular for the \threecnf{} benchmark, performance would have likely continued to improve further if the budget had been larger.

We thus conclude that the solver's flexibility should be chosen in relation to the available budget for configuration: solvers with few parameters can often be improved more quickly than highly flexible solving frameworks, but, given enough computational resources and powerful configurators, the latter ones can typically offer a greater performance potential.

\changed{
\subsection{Results with an Increased Cutoff Time for Validation} \label{sec:impact-validation-time}

Next to the overall time budget allowed for configuration, another important time limit is the cutoff time allowed for each single solver run; due to our limited overall budget, we chose this to be quite low: 300 seconds both for solver runs during the configuration process and for the final evaluation of solvers on previously unseen test instances. 

Here, we study how using a larger cutoff time at evaluation time affects results, mimicking a situation where we care about performance with a large cutoff time but use a smaller cutoff time for the configuration process to make progress faster. In fact, several studies in the literature (e.g., \cite{hutter-fmcad07a,khudabukhsh-ijcai09a,tompkins-sat11a}) used a smaller cutoff time for configuration than for testing, and we found that improvements with a time budget around 300 seconds often lead to improvements with larger cutoff times.

Table \ref{tab:indu2014-5000s} shows the results we obtained when using a cutoff time of 5000 seconds for validation (the same as the SAT competition) for the \indu{} track of CSSC 2014. Qualitatively, these results are quite similar to those obtained with an evaluation cutoff time of 300s (compare Table \ref{tab:indu2014}), with only few differences. 
As expected, given the larger cutoff time, all solvers solved substantially more instances (especially for the \bmcshort{} and \circuit{} benchmarks). 
Nevertheless, with a cutoff time of 5000 seconds, for all solvers, the configured variant (configured to perform well with a cutoff time of 300 seconds) still performed better than the default version, making us more confident that configuration does not substantially overtune to achieve good performance on easy instances only.
}

\begin{table}[t]
\changed{
\setlength\tabcolsep{0.4em}
\sffamily\footnotesize\centering
\begin{tabular}{l | ccc | c | cc}
\toprule[1.0pt]
 & \multicolumn{4}{c|}{$\#$timeouts default $\to$ $\#$ timeouts configured (on test set)} & \multicolumn{2}{c}{Rank} \\
 & \bmcshort & \circuit & \ibm & \textit{Overall} & def & {\scriptsize{CSSC}}\\
\midrule
\lingeling & $10 \to 10$ & $6 \to \mathbf{3}$ & $66 \to \mathbf{65}$ & $82 \to \mathbf{78}$ & 1 & 1\\
\myrowcolour{}\minisathack & $\mathbf{11} \to 12$ & $6 \to \mathbf{4}$ & $67 \to 67$ & $84 \to \mathbf{83}$ & 2 & 2\\
\riss & $19 \to \mathbf{12}$ & $3 \to 3$ & $70 \to \mathbf{68}$ & $92 \to \mathbf{83}$ & 3 & 3\\
\myrowcolour{}\sparrow & $19 \to \mathbf{14}$ & $3 \to 3$ & $70 \to \mathbf{69}$ & $92 \to \mathbf{86}$ & 4 & 4\\
\cryptominisat & $20 \to \mathbf{18}$ & $9 \to \mathbf{4}$ & $66 \to \mathbf{64}$ & $95 \to \mathbf{86}$ & 5 & 5\\
\myrowcolour{}\clasp & $31 \to \mathbf{18}$ & $3 \to 3$ & $70 \to \mathbf{69}$ & $104 \to \mathbf{90}$ & 6 & 6\\
\bottomrule[1.0pt]
\end{tabular}
\caption{Results for CSSC 2014 competition track \indu{}, when final solvers are evaluated with a per-run cutoff time of 5000 seconds instead of 300 seconds (as in Table \ref{tab:indu2014}). 
The ranking is the same for the default and configured algorithms, but this can be attributed to chance as there are two ties in terms of timeouts, which are only broken by average runtimes: \minisathack{} (513.7s) \vs{} \riss{} (520.5s), and \sparrow{} (548.5s) \vs{} \cryptominisat{} (555.0s).
\label{tab:indu2014-5000s}}
}
\end{table}

\subsection{Results with a Single Configurator} \label{sec:impact-configurator-discussion}

\begin{table}[tb]
\sffamily\small\centering
\setlength\tabcolsep{0.15em}
\begin{tabular}{l | ccccc | ccccc}
\toprule[1pt]
 & \multicolumn{5}{c|}{Training} & \multicolumn{5}{c}{Test}\\
 & \minismacd{} & \minismacc{} & \minipils{} & \miniggad{} & \miniggac{} &
 \minismacd{} & \minismacc{} & \minipils{} & \miniggad{} & \miniggac{}\\
\midrule
\dccsat & $1.0$ & $1.0$ & $1.0$ & $1.0$ & $1.0$ & $1.0$ & $1.0$ & $1.0$ & $1.0$ & $1.0$\\
\csccsat & $1.0$ & $1.0$ & $1.0$ & $3.0$ & $1.6$ & $1.0$ & $1.0$ & $1.0$ & $1.6$ & $1.0$\\
\probsat & $1.0$ & $1.2$ & $1.1$ & -- & -- & $1.0$ & $1.5$ & $1.1$ & -- & --\\
\minisathack & $1.2$ & $1.1$ & $1.2$ & $1.5$ & $2.0$ & $0.9$ & $1.0$ & $1.0$ & $1.0$ & $1.8$\\
\yalsat & $1.2$ & $2.2$ & $2.0$ & $18.1$ & $18.5$ & $1.1$ & $0.6$ & $1.8$ & $6.6$ & $12.3$\\
\cryptominisat & $1.1$ & $1.2$ & $1.4$ & $3.4$ & $3.9$ & $1.0$ & $1.3$ & $1.4$ & $3.5$ & $4.1$\\
\clasp & $1.1$ & $1.5$ & $1.9$ & -- & -- & $1.2$ & $0.9$ & $1.5$ & -- & --\\
\riss & $1.1$ & $1.1$ & $1.4$ & $10.5$ & $10.1$ & $1.2$ & $0.9$ & $1.1$ & $6.0$ & $7.4$\\
\sparrow & $1.3$ & $1.1$ & $2.0$ & -- & -- & $1.1$ & $1.0$ & $1.8$ & -- & --\\
\lingeling & $1.5$ & $1.6$ & $1.4$ & -- & $77.2$ & $1.4$ & $1.0$ & $1.3$ & -- & $30.5$\\
\bottomrule[1pt]
\end{tabular}
\caption{Geometric mean of \emph{PAR-10 slowdown factors over the CSSC result} (i.e., over using the oracle best of our five configuration approaches), computed across the CSSC 2014 scenarios for which the respective solver was
submitted. Note that on the training set $1.0$ is a lower bound for this metric, while on the
test set values smaller than $1.0$ are possible if one configuration generalizes better than another.
Entry `--' denotes that \gga{} was incompatible with the configuration space (see \ref{app:gga} for details).\label{tab:geometric-slowdowns}}
\end{table}

While the CSSC addressed the performance of SAT solvers rather than the performance of configurators, we have been asked whether our complex configuration pipeline was necessary, or whether a single configurator would have produced similar or identical results. Indeed, counting the choice of discretized \vs{} non-discretized parameter space, our pipeline used five configuration approaches (\paramils{}-discretized, \gga{}, \gga{}-discretized, \smac{}-discretized, and \smac{}). Thus, if one of these approaches had yielded the same results all by itself, we could have reduced our overall configuration budget five-fold.

To determine whether this was the case, we evaluated the solver performance we would have observed if we had used each configuration approach in isolation. 
For each configuration scenario and each approach, we computed the \emph{PAR-10 slowdown factor over the CSSC result} as the PAR-10 achieved with the respective approach, divided by the PAR-10 of the approach with best training performance (which we selected in the CSSC). If a configuration approach achieves a PAR-10 slowdown factor close to one, this means that it gives rise to solver performance close to that achieved by our full CSSC configuration pipeline.
For each solver, we then computed the geometric mean of these factors across the scenarios it was configured for.
%

Table \ref{tab:geometric-slowdowns} shows that both \smac{} variants performed
close to best for all solvers, meaning that we would have achieved similar results had we only used \smac{} in the CSSC. \paramils{} yielded the next best performance, followed by \gga{}.
Full results can be found in the accompanying technical report~\cite{hutter-tech14b}.
Despite \smac{}'s strong performance, we believe it will still be useful to run several configuration approaches in future CSSCs, both to ensure robustness and to assess whether some configuration scenarios are better suited to other configuration approaches.

\section{Conclusion} \label{sec:conclusion}

In this article, we have described the design of the Configurable SAT Solver Challenge (CSSC) and the details of CSSC 2013 and CSSC 2014. We have highlighted two main insights that we gained from this competition:
\vspace*{0.2cm}
\begin{enumerate}
	\item Automated algorithm configuration often improved performance substantially, in several cases yielding average speedups of orders of magnitude. 
	\item Some solvers benefited more from automated configuration than others, leading to substantially different algorithm rankings after configuration than before (as, \eg{}, measured by the SAT competition).
\end{enumerate}
\vspace*{0.2cm}
Also, the configuration budget influenced which algorithm would perform best, and with the competition budget of 2 days on 4--5 cores, algorithms with larger parameter spaces exhibited more capacity for improvement. 

These conclusions have interesting implications for algorithm design: if an algorithm is likely to be 
applied across a range of specialized applications, then it should be made flexible by parameterization of its key mechanisms and components, and this flexibility should be exploited by automated algorithm configuration. Our findings thus challenge the traditional approach to solver design that tries to avoid having too many algorithm parameters (since these parameters complicate manual tuning and analysis). Rather, they promote the design paradigm of \emph{Programming by Optimization (PbO)}~\cite{hoos-cacm12}, which aims to avoid premature design choices and to rather actively develop promising alternatives for parts of the design that enable an automated customization to achieve peak performance on particular benchmarks of interest. 
Indeed, in the CSSC, we have already observed a trend towards PbO, as evidenced by the introduction of a host of new parameters into state-of-the-art solvers, such as \riss{} and \lingeling{}, between 2013 and 2014.

Finally, there is no reason why a configurable solver competition should be appropriate 
and insightful only for SAT. On the contrary, similar events would be interesting in the context of many other challenging computational problems, such as answer set programming, constraint programming or AI planning. Another interesting application domain is automatic machine learning, where algorithm configuration can adapt flexible machine learning frameworks to each new dataset at hand~\cite{thornton-kdd13a,feurer-nips2015a}.
We 
believe that for those and many other problems, similar findings to those we
reported here for CSSC would be obtained, leading to analogous conclusions regarding algorithm design.

\section*{Acknowledgements} \label{sec:ack}

Many thanks go to Kevin Tierney for his generous help with running \gga{}, including his addition of new features, his suggestion of parameter settings and 
his conversion script to read the pcs format. We also thank the solver developers for proofreading the description of their solvers and their parameters.
For computational resources to run the competition, we thank Compute Canada (CSSC 2013) and the
German Research Foundation (DFG; CSSC 2014).
F.\ Hutter and M.\ Lindauer thank the DFG for funding this research under Emmy Noether grant HU 1900/2-1. H.~Hoos acknowledges funding through an NSERC Discovery Grant.

\section*{References} 

{
\begin{small}
\bibliographystyle{newapa}
\bibliography{local,lib,proc}
\end{small}
}

\appendix

\section{Benchmark Sets Used} \label{app:benchmarks}

We mirrored the three main categories of instances from the SAT competition:
industrial, crafted, and random. In 2014, we also included
a category of satisfiable random instances from the SAT Races.
For each of these categories, we used various benchmark sets, each of them split into 
a training set to be used for algorithm configuration and a disjoint test set.

For each category, to weight all benchmarks equally, we used the same number of test instances from each benchmark;
these test sets were subsampled uniformly at random from the respective complete test sets.

All benchmarks are summarized in Tables \ref{tab:benchmarks-2013} and \ref{tab:benchmarks-2014} in the main text.

\subsection{Industrial Benchmark Sets} \label{sec:benchmarks-indu}

\paragraph{\swv} This set of SAT-encoded software verification instances consists of 604 instances generated with the CALYSTO static
checker~\cite{babic-hvc07}, used for the verification of five programs: the spam filter Dspam, the SAT solver HyperSAT, the Wine Windows
OS emulator, the gzip archiver, and a component of xinetd (a secure version of inetd). 
We used the same training/test split as \emcite{hutter-fmcad07a}, containing 302 training instances and 302 test instances. We used this benchmark set in the 2013 CSSC.
(In 2014, we only used it for preliminary tests since it is quite easy for modern solvers.)

\paragraph{\hw} This set of SAT-encoded bounded model checking instances consists of 765 instances generated by \emcite{zarpas-sat05a}
These instances were originally selected by \emcite{hutter-fmcad07a} as the instances in 40 randomly-selected folders from the IBM Formal Verification Benchmarks Library. 
We used their original training/test split, containing 382 training instances and 383 test instances.
We used this benchmark set in both the 2013 and 2014 CSSCs.

\paragraph{\circuit} These instances were produced by a circuit-based CNF fuzzing tool, \fuzzsat{}~\cite{brummayer-sat10a} (version 0.1). As \fuzzsat{} was originally designed to produce semi-realistic test cases for debugging SAT solvers, the majority of the instances it produces are trivial; however, occasionally, it produces more challenging instances. The CircuitFuzz instances were found by generating 10,000 \fuzzsat{} instances and removing all those that could be solved within one second by \lingeling{}. 
This instance generator was originally described in detail by \emcite{bayless-lion14a}; we used the 300 instances from that paper as the training set (except one quite easy instance, `fuzz\_100\_25433.cnf', which was dropped unintentionally by a script) and produced 585 additional instances using the same method, to form a testing set. We used this benchmark set in both the 2013 and 2014 CSSCs.
\changed{We used these instances as part of the industrial track since they are ``structured in ways that resemble (at least superficially) real-world, circuit-derived instances''~\cite{bayless-lion14a}; a case could, however, also be made for them to be part of the crafted or random track.}


\paragraph{\bmc} This set of SAT instances was derived by unrolling the 2008 Hardware Model Checking Competition circuits~\cite{biere-misc08a}. Each of these instances is a sequential circuit with safety properties. Each  circuit  was unrolled to 50, 100, and 200 iterations using the tool \textit{aigunroll} (version 1.9.4) from the AIGER tools~\cite{biere-misc07a}. We omitted trivial instances that were proven SAT or UNSAT during the unrolling process. 
While we used the entire set in 2013, in 2014 we removed the 60 instances provided by Intel in order to allow us to publicly share the instances.


\subsection{Crafted Benchmark Sets} \label{sec:benchmarks-crafted}

\paragraph{\gi} 
These instances were first used in the 2013 SAT Competition~\cite{mugrauer2013} and were generated by encoding the graph isomorphism problem to SAT according to the procedure described by~\emcite{toran2013}. Given two graphs $G_1$ and $G_2$ with $n$ vertices and $m$ edges (for whom the isomorphism problem is to be solved) the generator creates a SAT formula with $n^2$ variables and $O(n)+O(n^3)+O(n^4)$ clauses. Consequently, the generated instances can contain very many clauses. The 2\,064 SAT instances in this set were generated from different types of graphs, \changed{with the number of vertices $n$ ranging from 10 to 1296.\footnote{\changed{Note that the larger graphs have varying node degrees, and that each node can only match with other nodes of the same degree; this allows the encoding to generate much fewer clauses than in the worst case of equal node degrees.}}} 
We split the instances uniformly at random into 1\,032 training and 1\,032 test instances; \changed{in both the 2013 and 2014 CSSCs, we only used 351 of the test instances}.

\paragraph{\labs} 
This set contains 651 low-autocorrelation binary sequence (LABS) search problems that were encoded to SAT problems by first encoding them as pseudo-Boolean problems and then as SAT problems. Instances from this set were first used in the SAT Competition 2013 in the crafted category~\cite{mugrauer2013-2}.
We split this benchmark set uniformly at random into 350 training and 351 test instances, and used it in both the 2013 and 2014 CSSCs.

\paragraph{\queens} 
These 835 instances~\cite{manthey-sat14r} represent a parameterized unsatisfiable variation of the well-known $n$-queens problem, in which the task is to place $n$ queens on a chess board with $n \times n$ fields such that they do not attack each other.
In the variation considered here, the (unsatisfiable) problem is to either place $n+1$ \emph{rooks} or $n+1$ queens on a board of size $n \times n$. Additional constraints enforcing that there is a piece in each row/column/diagonal make it easier to prove unsatisfiability, and these constraints can be enabled or disabled by generator parameters. \changed{We used the generator \newdame{} provided by Norbert Manthey to generate instances with $n \in [10,50]$, using all rooks or all queens, using six different problem encodings, and using all combinations of enabling/disabling all types of constraints. We then removed trivial instances, ending up with 835 instances. For the CSSC 2014, we selected 484 training instances uniformly at random and used the remaining 351 as test instances.}

\subsection{Random Benchmark Sets} \label{sec:benchmarks-random}

\paragraph{\kthree} This is a set of 600 randomly-generated 3-SAT instances at the phase transition (clause to variable ratio of approximately 4.26). It includes both satisfiable and unsatisfiable instances.
The set includes 100 instances each with 200 variables (853 clauses), 225 variables (960 clauses), 250 variables (1066 clauses), 275 variables (1172 clauses), 300 variables (1279 clauses), and
325 variables (1385 clauses). These 600 instances were generated by Lin Xu using the random instance generator from the 2009 SAT competition, and were previously described by \emcite{bayless-lion14a}. We employed their uniform random split into 300 training and \changed{300 test instances, using all 300 test instances in the CSSC 2013 (random track) and only a subset of 250 test instances in the CSSC 2014 (random track).}

\paragraph{\threecnf} This is a set of 750 random 3-SAT instances (satisfiable and unsatisfiable) at the phase transition, with 350 variables and 1493 clauses. These instances were generated by the ToughSAT instance generator~\cite{bebel-sat13a} and split into 500 training and 250 test instances uniformly at random. We used this benchmark set in the 2014 CSSC (random track).

\paragraph{\unif} 
This set contains only unsatisfiable 5-SAT instances generated uniformly at random with 50 variables and 1\,056 clauses (a clause-to-variable ratio sharply on the phase transition). 
The instances were generated by the uniform random generator used in the SAT Challenge 2012 and SAT Competition 2013, with satisfiable instances being filtered out by running the SLS solver \probsat{}. 
%
%
We used this benchmark set in both the 2013 and 2014 CSSCs (random track).

\paragraph{\threesatonek} 
This is a set of 500 3-SAT instances at the phase transition, all satisfiable. Each instance has 1000 variables and 4260 clauses. These instances were previously described by \emcite{tompkins-sat11}. 
We used their uniform random split into 250 training and test instances in the 2013 CSSC (random track) and in the 2014 CSSC (random satisfiable track).

\paragraph{\fivesatfiveh} 
This set contains 500 5-SAT instances generated uniformly at random with a clause-to-variable ratio of 20. 
Each instance is satisfiable and has 500 variables and 10000 clauses. This set was first used for tuning the SAT solver Captain Jack and other SLS solvers~\cite{tompkins-sat11}. We used the original uniform random split into 250 training and test instances in the 2014 CSSC (random satisfiable track).

\paragraph{\sevensatninety} 
This set contains 500 7-SAT instances generated uniformly at random with a clause-to-variable ratio of 85.  
Each instance is satisfiable and has 90 variables and 7650 clauses. This set was also first used for tuning the SAT solver Captain Jack and other SLS solvers~\cite{tompkins-sat11}. We used the original uniform random split into 250 training and test instances in the 2014 CSSC (random satisfiable track).

\subsection{Instance Features Used for these Benchmark Sets} \label{sec:instance-features}

As described in \ref{sec:algoconf-smac}, \smac{} can use instance features to guide its search. 
Such instance features have predominantly been studied in the work on SATzilla for algorithm selection~\cite{nudelman2004understanding,xu-jair08a} and in machine learning models for predicting algorithm runtime~\cite{hutter-aij14a}. 
These features range from simple summary statistics, such as the number of variables or clauses in an instance, to the results of short, runtime-limited probes with local search solvers. 
In the context of algorithm configuration, we can afford somewhat more expensive features than for algorithm selection since we only require them on the \emph{training} instances (not the test instances) and can compute them once, offline. Nevertheless, we kept feature computation costs low to not add substantially to the time required for algorithm configuration. 

For the instance sets where we already had available instance features from previous work, we used those features. In particular, we used the 138 features described by \emcite{hutter-aij14a} for the datasets \swv{}, \ibm{}, \threesatonek{}, \fivesatfiveh{}, and \sevensatninety{}.
For the set \unif{}, we did not compute features since these instances were very easy to solve even with algorithm defaults (note that \smac{} also worked very well without features).
For the other datasets, we computed a subset of 119 features, including basic
features and feature groups based on survey propagation, clause learning, local
search probing, and search space size estimates.\footnote{%
The code for computing
these features is available at \url{http://www.cs.ubc.ca/labs/beta/Projects/EPMs/}.
}



\section{Configuration Procedures} \label{app:conf-proc-details}

This appendix describes the configuration procedures we used in more detail. Configurators typically iterate the following steps: (1)~execute the target algorithm on one or more instances with one or more configurations for a limited amount of time; (2)~measure the resulting performance metric and (3)~decide upon the next target algorithm execution. Beyond the key question of {which configuration} to try next, configurators also need to decide how many runs and which instances to use for each evaluation, and after which time to terminate unsuccessful runs. 
\paramils, \smac{}, and \gga{} differ in how they instantiate these components.

\subsection{\paramils{}: Local Search in Configuration Space} \label{sec:algoconf-pils}

\paramils~\cite{hutter-jair09a}, short for iterated local search in parameter configuration space, generalizes the simple (often manually performed) tuning approach of changing one parameter at a time and keeping changes if performance improves. While that simple tuning approach is a local search that terminates in the first local optimum, \paramils{} carries out an iterated local search~\cite{lourencco-handbook03a} that applies perturbation steps in each local optimum $o$ in order to escape $o$'s basin of attraction and carry out another local search that leads to another local optimum $o'$. Iterated local search then decides whether to continue from the new optimum $o'$ or to return to the previous optimum $o$, thereby performing a biased random walk over locally optimal solutions.
\paramils{} only supports categorical parameters, so numerical parameters need to be discretized before \paramils{} is run.

\paramils{} is an algorithm framework with two different instantiations that differ in their strategy of deciding how many runs to use to evaluate each configuration. The most straightforward instantiation, \emph{BasicILS}(N), resembles the approach most frequently used in manual parameter optimization: it evaluates each configuration according to a fixed number of $N$ runs on a fixed set of instances. While this approach is simple and intuitive, it gives rise to the problem of how to set the number $N$. Setting $N$ to a large value yields slow evaluations; using a small number yields fast evaluations, but the evaluations are often not representative for the instance set $\Pi$ (for example, if we choose $N$ runs we can cover at most $N$ instances, even if we only allow a single run per instance).
The second \paramils{} instantiation, \emph{FocusedILS}, solves this problem by allocating most of its runs to strong configurations: it starts with a single run per configuration and incrementally performs more runs for promising configurations. This means that it can often afford a large number of runs for the best configurations while rejecting most poor configurations based on a few runs. There is also a guarantee that configurations that were `unlucky' can be revisited in the search, allowing for a proof that FocusedILS---if run indefinitely---will eventually identify the configuration with the best performance on the entire training set.

Finally, \paramils{} also implements a mechanism for adaptively choosing the time after which to terminate unsuccessful target algorithm runs.
Intuitively, when comparing the performance of two configurations $\vtheta_{1}$ and $\vtheta_{2}$ on an instance, and we already know that $\vtheta_{1}$ solves the instance in time $t_1$, we do not need to run $\vtheta_{2}$ for longer than $t_1$: we do not need to know precisely \emph{how bad} $\vtheta_{2}$ is, as long as we know that $\vtheta_{1}$ is better.
More precisely, each comparison of configurations in \paramils{} is with respect to an instance set $\Pi_{sub} \subset \Pi$, and evaluations of $\vtheta_{2}$ can be terminated prematurely when $\vtheta_{2}$'s aggregated performance on $\Pi_{sub}$ is provably worse than that of $\vtheta_{1}$.
In practice, this so-called \emph{adaptive capping} mechanism
can speed up \paramils{}'s progress by orders of magnitude when the best configuration solves instances much faster than the overall maximal cutoff time \cite{hutter-jair09a}.

For all experiments in this paper, we used the FocusedILS variant of the most recent publicly available \paramils{} release 2.3.7\footnote{\url{http://www.cs.ubc.ca/labs/beta/Projects/ParamILS/}} with default parameters.

\subsection{\gga{}: Gender-based Genetic Algorithm} \label{app:gga}
The \emph{Gender-based Genetic Algorithm (\gga{})}~\cite{ansotegui-cp09a} is a configuration procedure that maintains a population of configurations and proceeds according to an evolutionary metaphor, evolving the population over a number of \emph{generations} in which pairs of configurations mate and produce offspring. 
\gga{} also uses the concept of \emph{gender}: each configuration is labeled with a gender chosen uniformly at random, and when configurations are selected to mate there are separate selection pressures for each gender: configurations 
from the first gender are selected based on their empirical performance, 
whereas configurations from the other gender are selected uniformly at random.
The second gender thus serves as a pool of diversity, countering premature convergence to a poor parameter configuration.

Unlike \paramils{}' local search mechanism, \gga{}'s recombination operator for combining the parameter values of two parent configurations can operate directly on numerical parameter domains, avoiding the need for discretization.

Like \paramils{}, \gga{} implements an adaptive capping mechanism, elegantly combining it with a parallelization mechanism that lets it effectively use multiple processing units. 
\gga{} only ever evaluates configurations in the selection step for the first gender, and its strategy is to evaluate several candidates in parallel until the first one succeeds. Here, the number of configurations to be evaluated in parallel is taken to be identical to the number of processing units available, $\#units$.\footnote{This coupling of adaptive capping and parallelization is the reason that \gga{} should not be run on a single core if the objective is to minimize runtime.}

Like the FocusedILS variant of \paramils{}, \gga{} also implements an ``intensification'' mechanism for increasing the number of runs $N$ it performs for each configuration over time.
Specifically, it keeps $N$ constant in each generation, starting with small $N_{start}$ in the first generation, and linearly increasing $N$ up to a larger $N_{target}$ in generation $G_{target}$ and thereafter; $N_{start}$, $N_{target}$, and $G_{target}$, are parameters of \gga{}. 

For all experiments in the CSSC, we used the most recent publicly available version of \gga{}, version $1.3.2$.\footnote{\url{https://wiwi.uni-paderborn.de/dep3/entscheidungsunterstuetzungssysteme-und-operations-research-jun-prof-dr-tierney/research/source-code/}} 
\gga{}'s author Kevin Tierney kindly provided a script to convert the parameter configuration space description for each solver from the competition's pcs format\footnote{\url{http://aclib.net/cssc2014/pcs-format.pdf}} to \gga{}'s native xml format. This script allowed us to run \gga{} for all solvers except those with complex conditionals.

Next to the parameters $\#units$, $N_{start}$, $N_{target}$, and $G_{target}$ mentioned above, free \gga{} parameters include the maximal number of generations, $G_{max}$ and the size of the population, $P_{size}$. 
The setting of these parameters considerably affects \gga{}'s behaviour and also determines its overall runtime (when run to completion).
If there is an external fixed time budget (as in the CSSC), these parameters can be modified to ensure that GGA does not finish far too early (thus not making effective use of the available configuration budget) while simultaneously ensuring that runs do not take far too long (in which case configuration would be cut off in one of the first generations, where the search is basically still random sampling). 
It is thus important to set GGA's parameters carefully. 
We set the following parameters to values hand-chosen by Kevin Tierney for the CSSC (leaving all other parameters at their default values):
$\#units = 4$, $P_{size} = 50$, $G_{target} = 75$, $G_{max} = 100$, $N_{start} = 4$, $N_{target} = \text{\#(training instances in the scenario)}$.\footnote{Actually, due to a miscommunication, we first ran experiments with $N_{target} = 2000$, obtaining somewhat worse results than reported here. After double-checking with Kevin Tierney we then re-ran everything with the correct value of $N_{target}$ that depended on the number of training instances in each configuration scenario. We only report these latter results here.} 

We performed a post hoc analysis, which suggests that these parameters may yet not be optimal: \gga{} often finished relatively few generations within its configuration budget. It might thus make sense to use a smaller value of $N_{target}$ in the future to reduce the number of instances considered per configuration. However, this means that \gga{} would never consider all instances and may overtune as a result. How to best set \gga{}'s parameters is therefore an open research question.

\subsection{\smac{}: Sequential Model-based Algorithm Configuration} \label{sec:algoconf-smac}

In contrast to the model-free configurators \paramils{} and \gga{}, \smac{}~\cite{hutter-lion11a} is a
sequential model-based algorithm configuration method, which means that it uses predictive models of algorithm performance \cite{hutter-aij14a} to guide its search for good
configurations. More specifically, it uses previously observed
$\langle{}$configuration, performance$\rangle{}$ pairs $\langle{}\vtheta,
f(\vtheta)\rangle{}$ to learn a random
forest of regression trees (see, \eg{}, \cite{breimann-mlj01a}) that express a function
$\hat{f}:\vTheta \rightarrow \mathds{R}$ predicting the
performance of arbitrary parameter configurations (including those not
yet evaluated) and then uses this function to guide its search. 
When instance characteristics $x_\pi \in \mathcal{F}$ are available for each problem instance $\pi$,
\smac{} uses observed $\langle{}$configuration, instance characteristic, performance$\rangle{}$ triplets $\langle{}\vtheta,
x_\pi, f(\vtheta,\pi)\rangle{}$ to learn a function $\hat{g}:\vTheta \times \mathcal{F} \rightarrow \mathds{R}$ that predicts the
performance of arbitrary parameter configurations on instances with arbitrary characteristics. These so-called \emph{empirical performance
models}~\cite{hutter-aij14a} are then marginalized over the instance characteristics of all training benchmark instances in order to derive
the function $\hat{f}$ that predicts average performance for each parameter configuration: 
$\hat{f}(\vtheta) = \mathds{E}_{\pi \sim \Pi_{train}} \left[ \hat{g}(\vtheta, \pi) \right].$

This performance model is used in a sequential optimization process as follows.
After an initialization phase, \smac{} iterates the following three steps: (1)~use the performance
measurements observed so far to fit a marginal random forest model $\hat{f}$;
(2)~use $\hat{f}$ to select a promising configuration $\vtheta \in \vTheta$ 
to evaluate next, trading off exploration in new parts of
the configuration space and exploitation in parts of the space known to
perform well; and (3)~run $\vtheta$ on one or more benchmark instances and
compare its performance to the best configuration observed so far.

\smac{} employs a similar criterion as FocusedILS to determine how many runs to perform
for each configuration, and for finite configuration spaces in the limit it also provably 
converges to the best configuration on the training set.
Unlike \paramils, \smac{} does not require that the parameter space be discretized.

When used to optimize target algorithm runtime, \smac{} implements an adaptive capping mechanism similar to the one used in \paramils{}.
When this capping mechanism prematurely terminates an algorithm run we only observe a lower bound
of the algorithm's runtime. In order to construct predictive models of algorithm runtime in the presence of such 
so-called \emph{right-censored data points}, \smac{} applies model-building techniques derived from the survival 
analysis literature~\cite{hutter-bayesopt11}. 

\section{Hors-Concours Solver \rissgExt{}} \label{sec:riss3ext-discussion}

So far, we have limited our analysis to the ten open-source solvers that competed for medals.
Recall that one additional solver, \rissgExt{}, only participated \emph{hors concours}. It was not eligible for a medal, because it had been submitted as closed source, being based on a highly experimental code branch of \rissg{} that had not been exhaustively tested and was therefore likely to contain bugs. 

As discussed in Section \ref{sec:controlled_execution}, our experimental protocol included various safeguards against such bugs: we measured runtime and memory externally, compared reported solubility status against true solubility status where this was known, and checked returned models when an instance was reported satisfiable. 
Our configuration pipeline detected and penalized these crashes automatically, enabling the configuration procedures to continue their search and find \rissgExt{} configurations with no or few crashes. In fact, the final best configurations identified by our configuration pipeline performed very well and would have handily won both the industrial and the crafted track of the CSSC 2013 had \rissgExt{} been submitted as open source: in the industrial track, it only left 82 problem instances unsolved (compared to 115 for \lingeling{}); and in the crafted track only 44 (compared to 96 for \clasp{}). Even though most of the instances \rissgExt{} did not solve were due to it crashing, all of these were `legal' crashes that simply did not output a solution (such as segmentation faults). In particular, we never observed \rissgExt{} to produce an incorrect output for a CSSC test instance with known satisfiability status.

However, empirical tests with benchmark instances are of course no substitute for formal correctness guarantees, and even seasoned solvers can have bugs.
Indeed, after the competition, \rissgExt{}'s developer found a bug in it (in on-the-fly clause improvement~\cite{on-the-fly-clause-improvement}) that caused some satisfiable instances to be incorrectly labeled as unsatisfiable.\footnote{Personal communication with \rissgExt{}'s developer Norbert Manthey.}
This being the case, it was fortunate that \rissgExt{} was ineligible for medals. 

While empirical testing on benchmark instances, as done in a competition, can never guarantee the correctness of a solver, in future CSSCs, we consider tightening solubility checks on the benchmark instances used, by either limiting the benchmark sets to contain only instances with known satisfiability status or to require (and check) proofs of unsatisfiability, as in the certified UNSAT track of the SAT competition.

\changed{
\section{Additional results with PAR-1 score} \label{sec:more-results-par1}\vspace*{-0.1cm}

\begin{figure}[t]
\captionsetup{justification=centering}
\subfloat[CSSC 2013]{\includegraphics[height=17em]{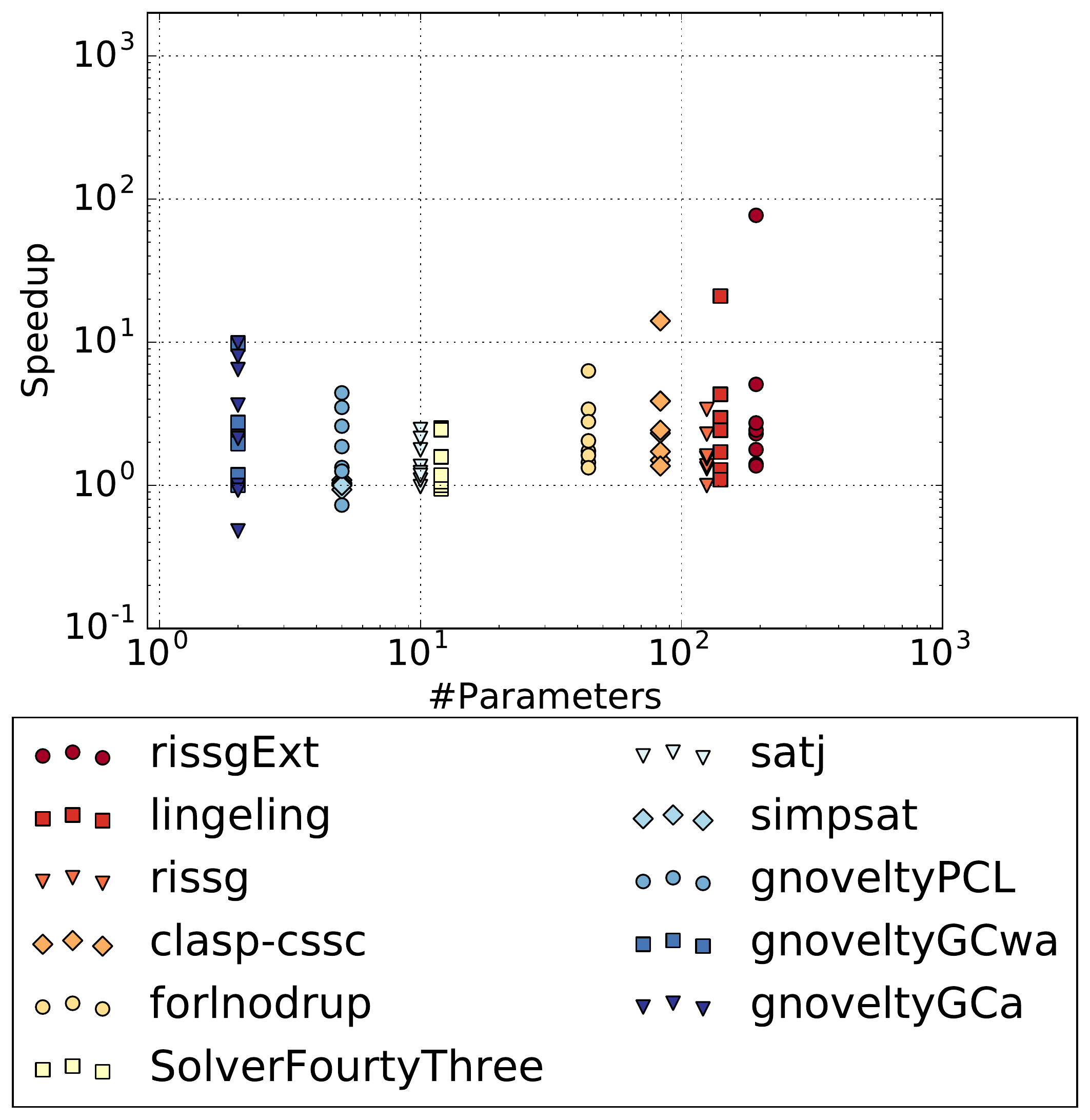}}~
\subfloat[CSSC 2014]{\includegraphics[height=17em]{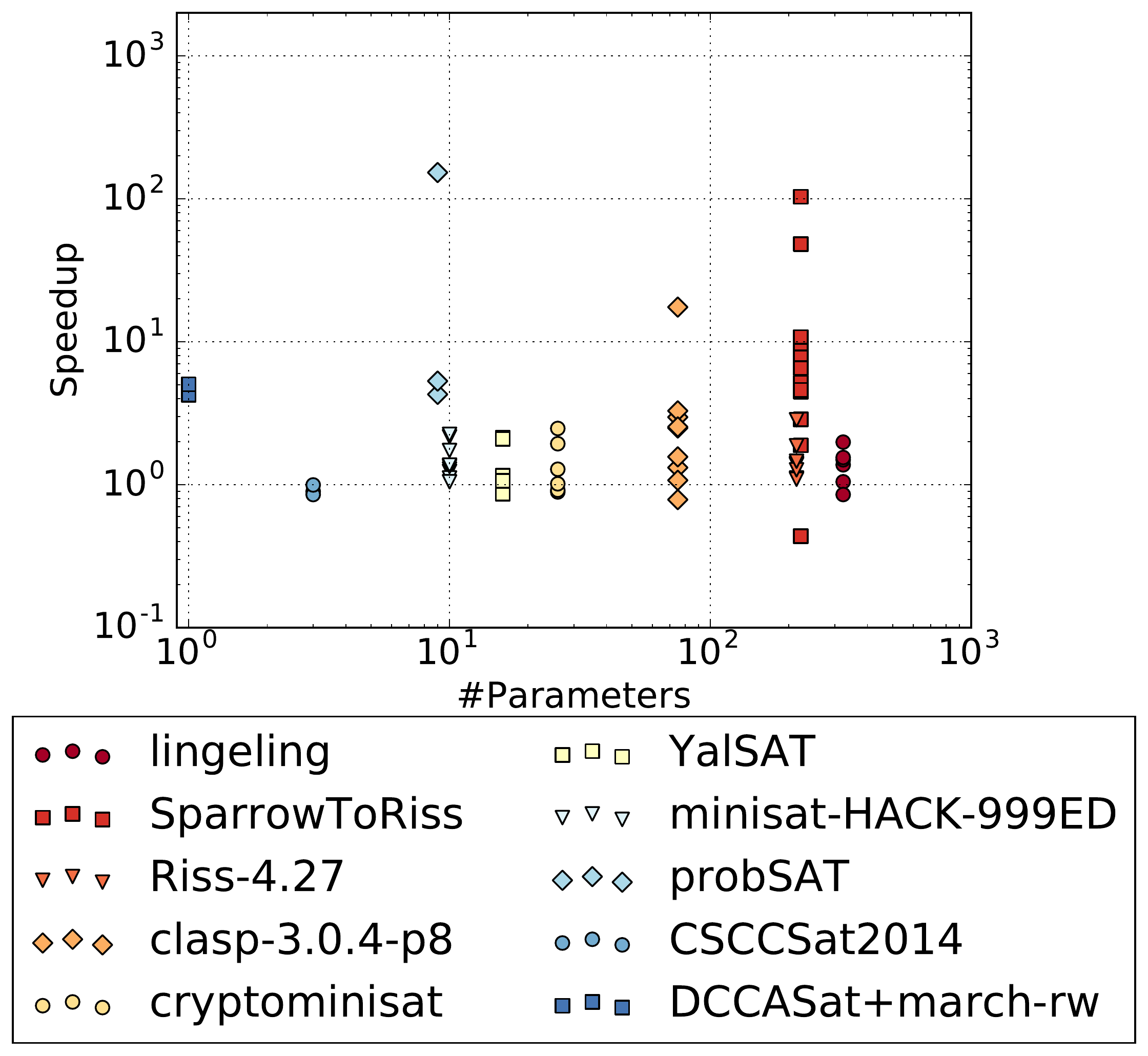}}
\captionsetup{justification=justified}
\vspace*{-0.1cm}
\caption{Same as Figure \ref{fig:configurability-vs-parameters}, but using PAR-1 instead of PAR-10 score.\label{fig:configurability-vs-parameters-par1}\vspace*{-0.3cm}
}
\end{figure}
}

Figure \ref{fig:configurability-vs-parameters-par1} visualizes runtime speedups obtained for each solver, counting timeouts at the cutoff time as the cutoff time itself (PAR-1). Compared to the PAR-10 results in Figure \ref{fig:configurability-vs-parameters}, speedups with PAR-1 are up to a factor of ten smaller for benchmark/solver combinations with many timeouts for the default, but otherwise results are qualitatively similar.

\end{document}